\documentclass[journal]{IEEEtran}

\usepackage{graphicx} 
\usepackage{amsmath} 
\usepackage{amssymb}  
\usepackage{amsfonts}
\usepackage{subfigure}
\usepackage{ctable}
\usepackage{makecell}

\makeatletter
\let\NAT@parse\undefined
\makeatother
\usepackage[numbers]{natbib}

\usepackage{amsthm}
\theoremstyle{definition}
\newtheorem{definition}{Requirement}[]

\usepackage{irap_acronyms}
\usepackage{irap_math}
\usepackage{irap_SIunits}
\usepackage{irap_misc}

\usepackage{soul,color}
\usepackage{verbatim} 
\usepackage{multirow}

\usepackage{lipsum}
\usepackage{xcolor}
\newcommand{\B}[1]{{\textbf{#1}}}
\newcommand{\bl}[1]{{\textcolor{black}{#1}}}

\DeclareMathOperator*{\argmin}{argmin}

\begin{document}

\title{
  Scan Context\texttt{++}: \bl{Structural Place Recognition Robust to Rotation and Lateral Variations \\ in Urban Environments}
}

\author{Giseop Kim,~\IEEEmembership{Student Member,~IEEE,}
        Sunwook Choi,
        and Ayoung Kim~\IEEEmembership{Member,~IEEE,}
\thanks{Giseop Kim is with the Department of Civil and Environmental Engineering, KAIST, Daejeon, S. Korea \texttt{paulgkim@kaist.ac.kr}
        }%
\thanks{Sunwook Choi is with Autonomous Driving Group, NAVER LABS. \texttt{sunwook.choi@naverlabs.com}
        }%
\thanks{Ayoung Kim is with the Department of Mechanical Engineering, SNU, Seoul, S. Korea \texttt{ayoungk@snu.ac.kr}
        }%

\thanks{Portions of this work were presented in part at the 2018 IEEE/RSJ International Conference on Intelligent Robots and Systems \cite{kim2018scan}.}
\thanks{Code will be available https://github.com/gisbi-kim/scancontext}
\thanks{This work was supported by [Localization in changing city] project funded by NAVER LABS Corporation and by the NRF (NRF-2019K2A9A1A06070173).}
}

%
%

\markboth{IEEE Transactions on Robotics,~Vol.~X, No.~X, Month~Year}
{Kim \MakeLowercase{\textit{et al.}}: Scan Context TODO}
%



\maketitle

\begin{abstract}

Place recognition is a key module in robotic navigation. The existing line of studies mostly focuses on visual place recognition to recognize previously visited places solely based on their appearance. In this paper, we address structural place recognition by recognizing a place based on structural appearance, namely from range sensors. Extending our previous work on a rotation invariant spatial descriptor, the proposed descriptor completes a generic descriptor \bl{robust to} both rotation \bl{(heading)} and translation \bl{when roll-pitch motions are not severe}. We introduce two sub-descriptors and enable topological place retrieval followed by the 1-DOF semi-metric localization thereby bridging the gap between topological place retrieval and metric localization. The proposed method has been evaluated thoroughly in terms of environmental complexity and scale. The source code is available and can easily be integrated into existing LiDAR \ac{SLAM}.

\end{abstract}


%
\IEEEpeerreviewmaketitle

\acresetall
\section{Introduction}
\label{sec:intro}

Recognizing a previously visited place is important for various robot missions (e.g., loop detection in \ac{SLAM} \cite{kim2018scan}, global localization for a kidnapped robot \cite{kim2019}, or multi-robot mapping \cite{saeedi2016multiple}). Describing a place with a set of compact representations has been tackled in depth within the computer vision and robotics community, yielding many state-of-the-art visual place recognition methods \cite{cummins2011appearance, milford2012seqslam,lowry2015visual,galvez2012bags}. In contrary to the flourishing studies on visual place recognition, studies on range sensors are still missing a solid solution to this global localization problem.

Recent studies have reported \cite{kim2018scan, kim2019, cieslewski2016point, kim2018stereo, mo2020place, oertel2020augmenting} that structural information could be more effective than appearance particularly within outdoor environments. \bl{These studies had attempted to overcome the major bottlenecks resulting from unstructured, unordered, and sparse range sensor data, which make consuming input data harder than pixelated image data. Existing methods have focused on compactly summarizing a place, but they have rarely achieved invariances in structural place recognition.}


Our preliminary version of this paper presented in \cite{kim2018scan} tried to establish this compact representation by capturing the highest structural points \bl{when the level of roll-pitch disturbances is not severe (e.g., under \unit{10}{\degree}) such as a wheeled robot or a slow walking hand-held system}. This strategy allowed us to achieve robustness for underlying structural variance (e.g., dynamic objects and seasonal changes) for incoming \ac{LiDAR} measurements. Although our previous Scan Context showed meaningful performance, the algorithm failed to achieve invariance in the lateral direction and was inefficient using a brute-force search. Overcoming these limitations in \cite{kim2018scan}, we complete the algorithm to include both rotational and lateral \bl{robustness} thereby introducing a generic \textit{structural place recognition} for a range sensor. Secondly, the modified algorithm improved previously brute-force search to use sub-descriptors and expedited the process by the order of magnitude. \bl{In summary, our new contributions are:}


\begin{itemize}

\item \bl{\textbf{Robustness to Lateral/Rotational Changes: }} Missing lateral invariance may be a critical issue in an urban environment where lane-level change is inevitable. To resolve this limitation, we generalized the previous descriptor \bl{to include both lateral and rotational robustness simultaneously}. \bl{This is achieved via \textit{Scan Context augmentation} based on urban road assumption.}


\item \textbf{Semi-metric Localization: } Combining place retrieval and metric localization, our global place recognition method bridges the gap between topological and metric localization. The proposed method provides not only the retrieved map place index but also 1-DOF (yaw or lateral) initial guess for metric refinement such as \ac{ICP}.

\item \textbf{Lightweight and Modules Independence: } As a global localizer, the proposed method does not require prior knowledge or any geometric constraints (e.g., odometry). The implementation is lightweight provided in a single C++ and header pair and readily integrable to existing \ac{SLAM} framework.

\item \textbf{Real-time Performance on CPU: } By introducing compact summarizing sub-descriptors, \textit{keys}, we achieved substantial cost reduction. The \textit{retrieval key} based tree search eliminates naive pixel-wise comparison followed by \textit{aligning key} based pre-alignment. Our method runs in real-time supporting up to \unit{100}{Hz} (e.g., average \unit{7.4}{ms} on \texttt{KITTI 00} \cite{geiger2012we}) without requiring GPU.

\item \textbf{Extensive Validation: } We evaluate the proposed method \bl{across diverse and challenging test scenarios} to validate both in-session and multi-session scenarios. We note the existing precision-recall curve may not fully capture the loop-closure performance for \ac{SLAM} research missing evaluation on the match distribution. We propose to use DR (distribution-recall) curves to measure not only the recalls but also their diversity for the meaningful loop-closure.

\end{itemize}


\section{Related Works}
\label{sec:related}

In this section, we provide a literature review on place recognition in both visual and structural aspects. We briefly review recent place recognition works, focusing on the sensor modality as well as global and local descriptions.


\subsection{Place Recognition for Visual Sensing}
\label{sec:litvis}

For visual recognition, both the local and global aspects of the place summarization were examined. The local description-based methods relied on detecting and describing handcrafted local keypoints (i.e., a small patch) \cite{bay2008speeded, rublee2011orb}. Using these local descriptors, Bayesian inference \cite{cummins2011appearance} or bag-of-words vocabulary tree \cite{galvez2012bags} was applied for place recognition. \citeauthor{cadena2012robust} proposed fusing the bag-of-words and a \ac{CRF} matching of 3D geometry for a stereo camera system.

Compared to local descriptors, global descriptors are more compact in representation and robust to local noises. The entire image is encapsulated by a single condensed representation (e.g., a fixed-size vector \cite{sunderhauf2015performance, arandjelovic2016netvlad} or a downsized image \cite{milford2012seqslam}) without maintaining a set of local keypoint descriptors. Similarly, as in local descriptors, recent studies in global descriptors enhanced the performance by exploiting structural information. \citeauthor{oertel2020augmenting} \cite{oertel2020augmenting} reported that the use of structural cues when making a global descriptor yields higher performance than appearance-only methods. \citeauthor{mo2020place} \cite{mo2020place} fed reconstructed 3D sparse points into a \ac{LiDAR} descriptor pipeline, which outperformed appearance-only-based global descriptors.

\subsection{Place Recognition for Range Sensing}
\label{sec:litlidar}


\subsubsection{LiDAR}

The early phase of LiDAR-based place recognition focused on 2D range data \cite{tipaldi2010flirt, tipaldi2013geometrical}. \bl{\citeauthor{olson2009real} proposed correlative scan matching-based loop closure detection for 2D LiDAR \cite{olson2009real, olson2009recognizing}}. As 3D LiDAR appeared, 3D point cloud summarization drew attention. For the initial 3D LiDAR place recognition methods \cite{steder2010robust, rusu2010fast, bosse2013place}, local keypoint-based approaches were used, similar to following the early history described above in the visual domain.

A point cloud from a 3D LiDAR poses challenges in a different aspect. First, the data is unstructured without having a constant and consistent grid density. Second, the data sparsity grows as the range increases, varying the target object density depending on the sensing range. These sensor characteristics make the local descriptions unstable; thus, a courser summarization unit that is robust to the local noise and inconsistent point density is preferred. M2DP \cite{he2016m2dp} compressed a single LiDAR scan into a global descriptor (i.e., 192D vector) that is robust to noisy input. PointNetVLAD \cite{angelina2018pointnetvlad} leveraged a learning-based approach to summarize a place into a single vector representation.

However, despite the performance and robustness of global descriptors, one drawback is that they do not secure invariance compared with local-based methods. As reported in \cite{kim2019}, these global descriptors were less invariant to the transformation (e.g., heading changes) because transformed local point coordinates may produce different embedding and cause failure in place recognition (\figref{fig:exampletax}). \bl{Recently, similar to our approach, a semi-handcrafted heading-invariant feature learning approach named LocNet \cite{yin20193d} was proposed. However, compared to LocNet}, achieving \bl{not only} rotational \bl{but also} translational invariance is required while maintaining the performance of the current state-of-the-art global point cloud descriptors.


In this line of study, a local characteristic such as segment or height was examined. For example, \citeauthor{dube2020segmap} proposed a segment-based global localization method using a handcraft segment descriptor \cite{dube2017segmatch} and learned segment embeddings \cite{dube2020segmap}. They recovered a relative transformation between two matched frames through geometric consistency checks, even under severe viewpoint changes such as reverse revisits. Our preliminary work, \textit{Scan Context} \cite{kim2018scan}, proposed to make a 2D descriptor based on the height of the surrounding structures. This descriptor obtained rotational invariance and yielded relative yaw as a by-product. Stemming from this work, some authors \cite{oreos19iros, Chen2019OverlapNetLC} tried to simultaneously estimate the relative yaw between two scans and their similarity. Learning-based approaches included semi-learned \cite{yin2018locnet,kim2019}, and full learning-based \cite{oreos19iros, Chen2019OverlapNetLC} methods.

\subsubsection{Radar}
\label{sec:litradar}

More recently, a long-range perceptible \ac{FMCW} radar has been highlighted in robotics applications \cite{RadarRobotCarDatasetICRA2020, kim2020mulran}. Radars provide far longer range and robustness compared to cameras and LiDAR; however, radar place recognition methods are still not mature. Exploiting the image-like format of radar data, some studies leveraged computer vision techniques to describe a radar image at local \cite{barnes2020under} and global description \cite{suaftescu2020kidnapped, gadd2020look} levels. However, the projection model of the radar image inevitably eliminates height information generating a top-down view. To handle this elevation loss, \citeauthor{hong2020radarslam} \cite{hong2020radarslam} used a LiDAR descriptor M2DP \cite{he2016m2dp} but using the intensity of a pixel in lieu of a point's height. Similarly, \cite{kim2020mulran} showed the feasibility of Scan Context by replacing the height with the intensity.


\section{Requirements for Structural Place Recognition}
\label{sec:probdefmain}

\subsection{Terminology and Problem Definition}
\label{sec:probdef}

\bl {
We first define our \textit{place recognition} problem. As a robot traverses an environment, a set of range sensor measurements is streamed with increasing timestamps. We consider every single sensor measurement $z_t$ acquired at a certain spatial location $l_t$ at time $t$ as a \textit{place}. A \textit{map} is a database, a set of all streamed measurements after the time a robot has started a mission. Then, our place recognition can be defined as finding a revisited place within a map for a query place. It is also important to trustfully decide whether there is no revisited place in a map. \textit{Revisitedness} is satisfied for two places, $a$ and $b$, temporally apart from a certain window size (i.e., $|{t_b} - {t_a}| > \delta_t $), if the Euclidean distance between two places' spatial locations is less than a certain threshold (i.e., $|l_{b} - l_{a}| < \delta_l $).
}

\bl {
To construct such a place recognition system, two submodules are required. The first is description function $f(\cdot)$. To ease handling noisy or heavy raw measurements, a raw measurement $z_t$ is encoded into a more compact form called descriptor $\textbf{f}_t = f(z_t)$. The second is retrieval that defines a similarity function ${sim}(\cdot, \cdot)$ or distance function $D(\cdot, \cdot)$. It takes two descriptors and returns a scalar-value similarity or distance in the descriptor space. Then, the place recognition is reduced to the nearest search problem using the description and similarity functions when a query measurement $z$ and a map are given. One can conclude that two places $a$ and $b$ are the same if the descriptor distance $D(\textbf{f}_a, \textbf{f}_b)$ is lower than a threshold $\tau$.
}

\begin{table}[!b]
    \vspace{-3mm}
    \caption{Required invariance for visual and structural place recognition.}
  \centering
    \vspace{-1mm}
      \begin{tabular}{l|ccc|cc|cc}
                & \multicolumn{3}{c|}{Internal Factor} & \multicolumn{2}{c|}{External Factor} & \multicolumn{2}{c}{Sensor Specification} \\\hline
      Visual    & R          & T             & S      & I              & W                  & FOV              & \multicolumn{1}{l}{}  \\
      Structral & R          & T             & SP     & D              & SV                 & FOV              & NR                    \\
    \end{tabular}\\
    \vspace{5pt}
    \begin{tabular}{|lll|}
      \hline
      R: rotation    & I: illumination         & FOV: field of view \\
      T: translation & W: weather              & NR: number of rays \\
      S: scale       & D: dynamic objs         &   \\
      SP: sparsity   & SV: structural variance &\\\hline
    \end{tabular}
    \vspace{5pt}
    \label{tab:taxonomy}
\end{table}

\subsection{Invariance}
\label{seq:reqinv}

Most LiDAR place recognition methods \cite{he2016m2dp, dube2020segmap, yin2018locnet, Chen2019OverlapNetLC} have been tested over less complex environments \cite{geiger2012we, fordcampus} with few dynamic objects or viewpoint changes. The existing research has mostly focused on increasing the discriminability of a descriptor, rather than on defining and overcoming structural diversity.
We provide a taxonomical analysis of the potential \bl{nuisances} for structural place recognition, as shown in \tabref{tab:taxonomy}. We categorized each invariance in the comparison to the rather widely-studied visual place recognition problem for each corresponding invariance type.

\subsubsection{Internal Factors} The measurement's variation could be derived from a robot itself, we named \textit{internal factors}. This includes rotation, translation, and scale changes of the sensor coordinate mostly induced by ego-motion (R, T, and SP in \tabref{tab:taxonomy}). \figref{fig:exampletax} illustrates the sample measurement discrepancy under rotational and translational variance. In terms of scale, the same object looks very different due to the variation in point cloud density caused by the sensing distance.

\subsubsection{External Factors} Similar to illumination changes (short-term variance) and weather changes (long-term variance) in the visual domain, structures may undergo similar variance in the short-term through occlusions by dynamic objects and in the long-term through permanent structural changes from construction or demolition. This external factor becomes critical as we deploy robots for long-term navigation.

\subsubsection{Sensor Characteristics} The last factor, sensor characteristics, maybe more range-sensor specific. Unlike the highly structured sensor data obtained by cameras, LiDAR point clouds are unstructured, and sensing changes dramatically depending on the sensor's specifications (e.g., range, number of rays, and point cloud resolution depending on \ac{FOV}). Thus, a generic place recognition system should be invariant to sensor specifications.

\begin{figure}[!t]


  \centering
  \subfigure[Rotational Displacement]{%
    \includegraphics[width=0.5\columnwidth]{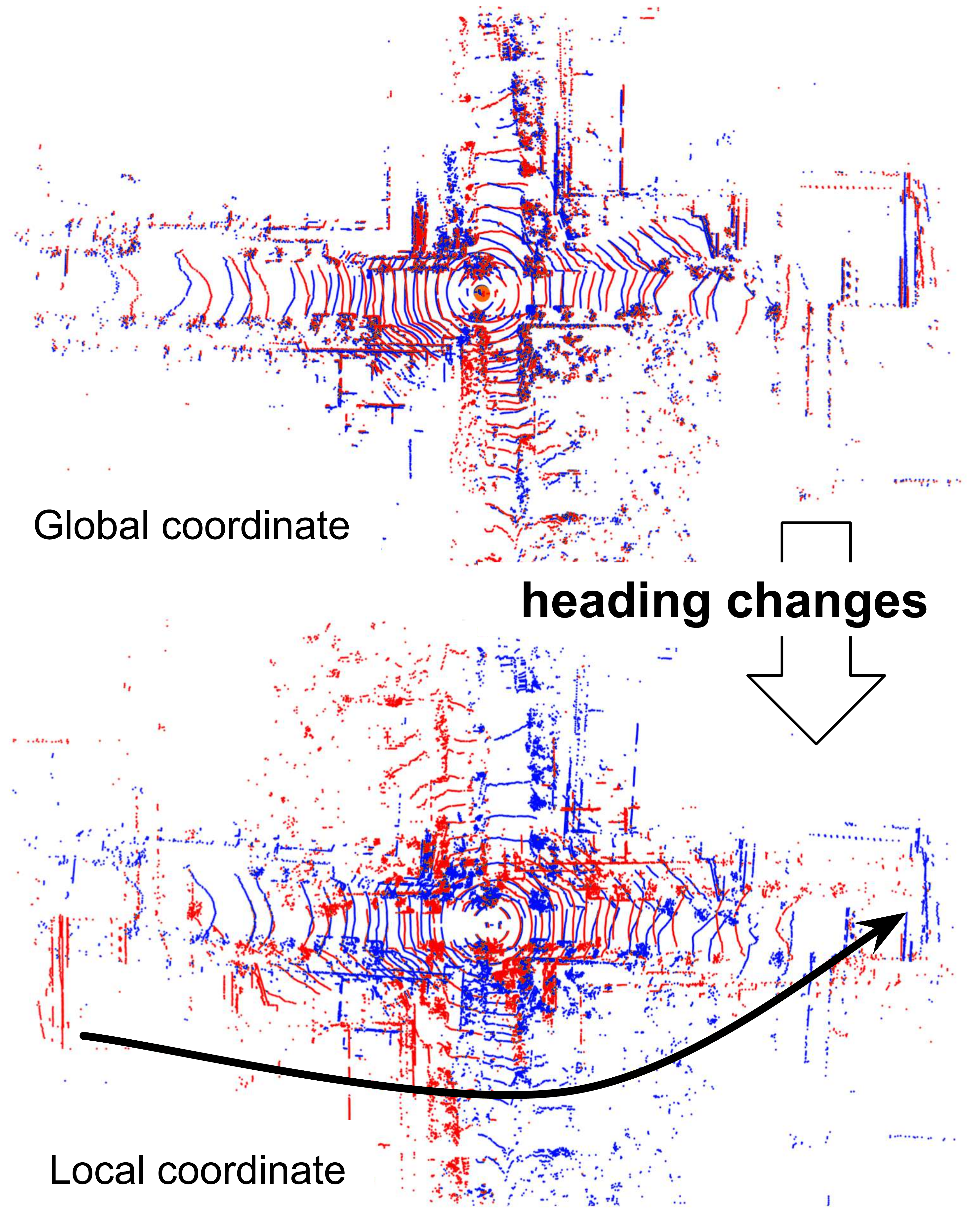}
    \label{fig:examplerot}
  }%
  \subfigure[Lateral Displacement]{%
    \includegraphics[width=0.47\columnwidth]{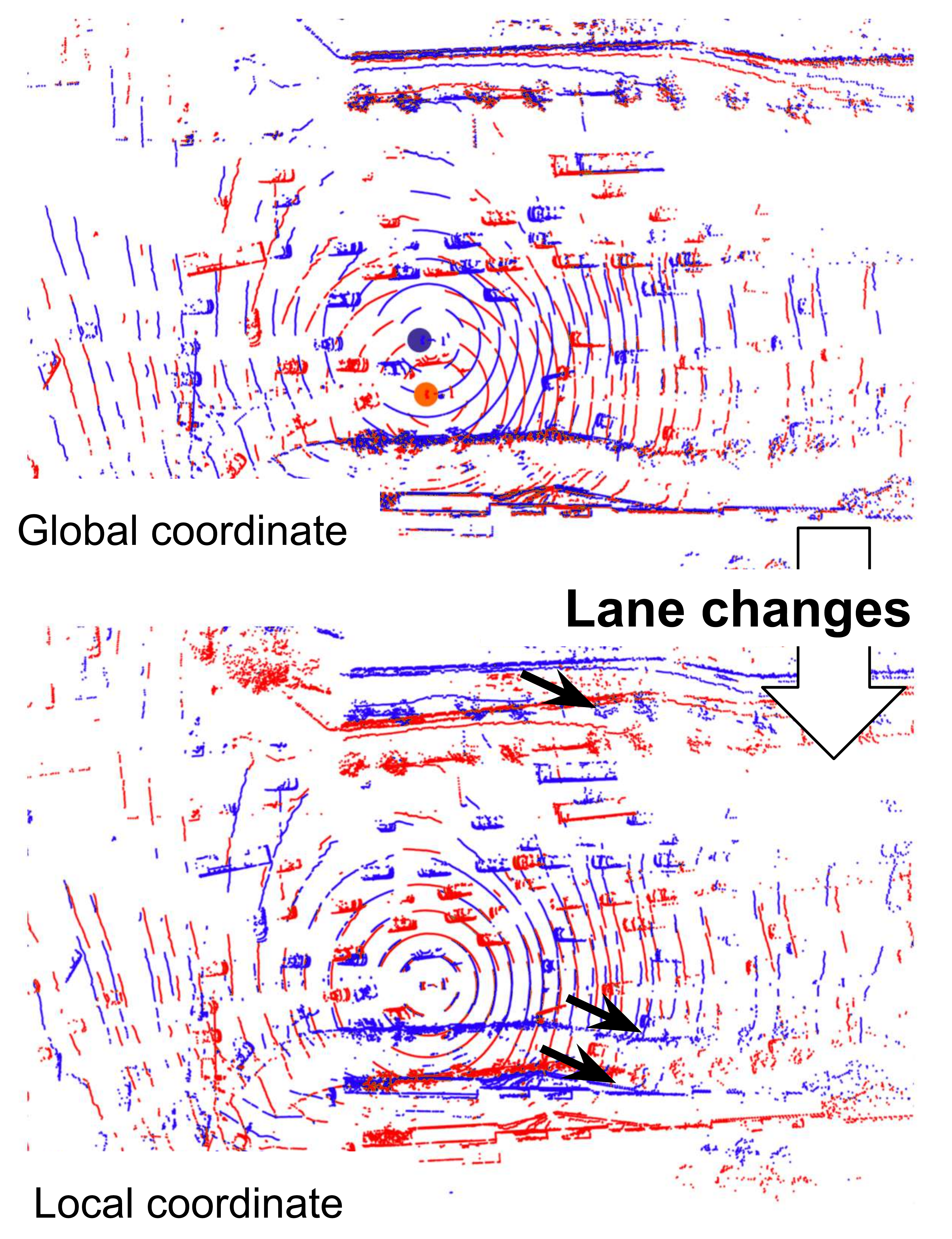}
    \label{fig:examplelane}
  }
  \caption{Sample point cloud undergoing rotational (e.g., reversed revisits) and translational (e.g., lateral lane changes) motion. The red indicates the query scans, and the blue depicts experience in the database. Unlike the view in global coordinates (i.e., world coordinate, top row), the measurement looks different in the local coordinates (i.e., sensor coordinate, bottom row). Also, note that many dynamic objects exist in both scans.
  }
  \label{fig:exampletax}
\end{figure}

\begin{figure*}[t!]
  \centering
  \includegraphics[width=0.68\textwidth,  trim = 20 10 -50 0, clip]{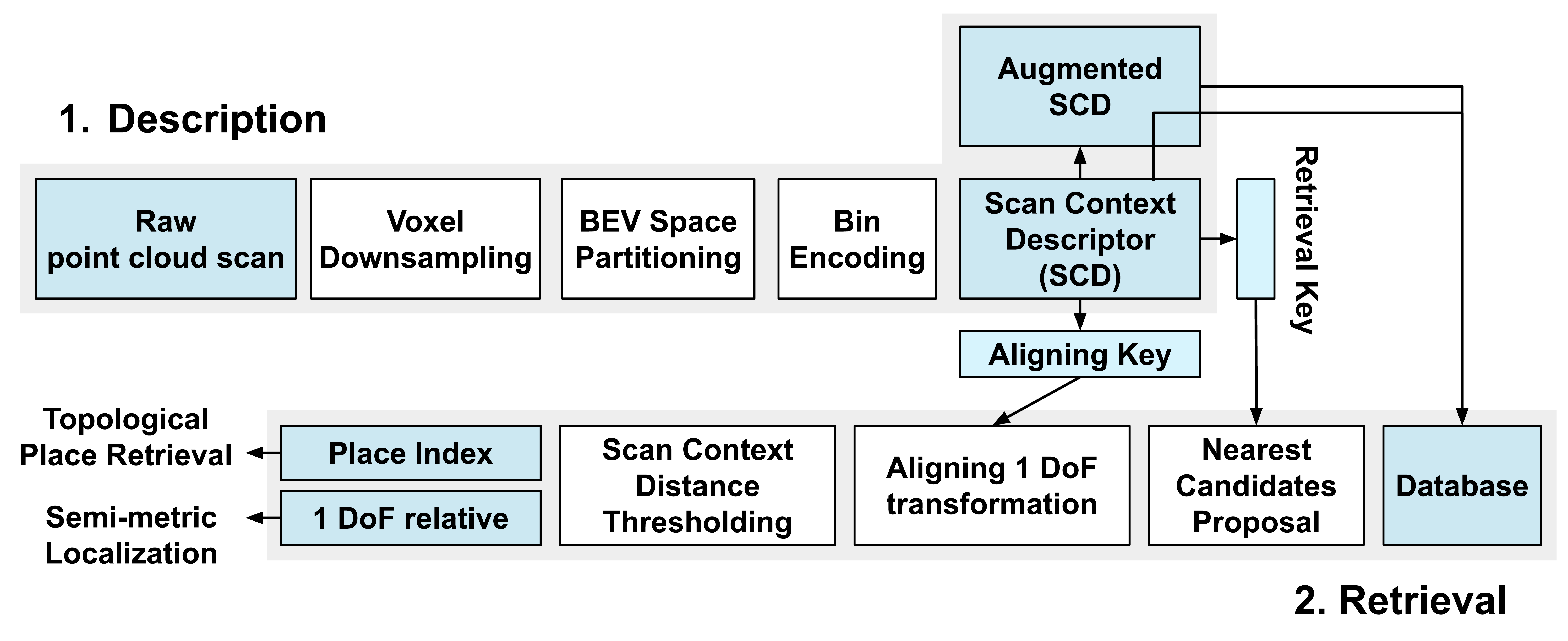}
  \caption{
    The overall framework. Given a raw range measurement, the proposed method seeks the corresponding place index from a set of places in the map.
  }
  \label{fig:pipeline}
\end{figure*}


\subsection{Overview}
\label{sec:overview}

The proposed method consists of two parts: (\textit{i}) place description and (\textit{ii}) place recognition. The overall pipeline is illustrated in \figref{fig:pipeline}. The place recognition module consists of place retrieval, semi-metric localization, and verification. In the next two sections, we will introduce each module in detail.


\section{Scan Context Descriptor (SCD)}
\label{sec:sc2}

In this section, we describe a novel spatial descriptor named \textit{\ac{SCD}}. The pipeline begins with partitioning the raw measurement and projecting them into discretized bins using \ac{BEV}. When dividing into the \ac{BEV} bins, two types of perpendicular bases (polar and Cartesian) are considered. After partition and coordinate selection, the subset of the measurement is encoded to its associated discretized bin using the bin encoding function. \bl{As we present, the invariance of the proposal place recognition module arises from the bin encoding function and the distance function.}

\subsection{Motivation}
\label{sec:scdmotiv}

Our descriptor and search engine were strongly motivated by the revisit patterns in urban environments. We found typical patterns due to the nonholonomic vehicle motion following traffic rules (e.g., lane-keeping). The dominant motion is locally two-dimensional, and the motion occurs with at most two directions that are likely to be disjointed. These typical patterns motivated the choice of two coordinate frames, polar and Cartesian, and the associated matching algorithm.

\subsection{Descriptor Axes and Resolution}
\label{sec:roipart}

\bl{We assume that the input is a single scan of 3D LiDAR.} The first phase for generating the descriptor is to partition a \bl{downsampled point cloud} within a \ac{ROI}. The upper bound of the \ac{ROI} and the partitioning resolution decide the shape of an \ac{SCD}. Given the partitioned raw measurements, we project them on 2D descriptor space; \bl{namely, the approach first (i) projects each 3D point to a 2D point, (ii) parametrizes the 2D point in polar or Cartesian coordinates, and (iii) obtains a scalar value (details in \secref{sec:bef}) for each bin by discretizing the 2D space.}

As shown in \figref{fig:pipeline}, we name the horizontal axis the \textit{\ac{A-axis}} and the vertical axis the \textit{\ac{R-axis}}. The change along the \ac{A-axis} corresponds to the column-wise shift; thus, pre-alignment along the \ac{A-axis} will allow us to infer a rough metric-level relative pose, overcoming changes in the associated direction. The choice of aligning/retrieval axes determines the type of \ac{SCD}, as either \ac{PC} and \ac{CC}.

\subsubsection{Polar Coordinates} As introduced in our earlier work \cite{kim2018scan}, the \ac{PC} adopts polar coordinates using the azimuth $\theta$ as the \ac{A-axis} and the radius $r$ as the \ac{R-axis}. Because the azimuth is on the \ac{A-axis}, the \ac{PC} is robust to rotational variance.

\subsubsection{Cartesian Coordinates} The \ac{CC} leverages Cartesian coordinates and uses the lateral direction ($y$) as the \ac{A-axis}. The longitudinal direction (or travel direction, $x$) becomes the \ac{R-axis}. Naturally, the descriptor is invariant to lateral direction translation.

\subsubsection{Descriptor Resolution} 
\begin{figure*}[!t]
  \centering
  \begin{minipage}{0.85\textwidth}
    \centering
    \subfigure[Partitioning and coordinates selection]{%
    \includegraphics[width=0.4\textwidth, trim = -30 0 0 0, clip]{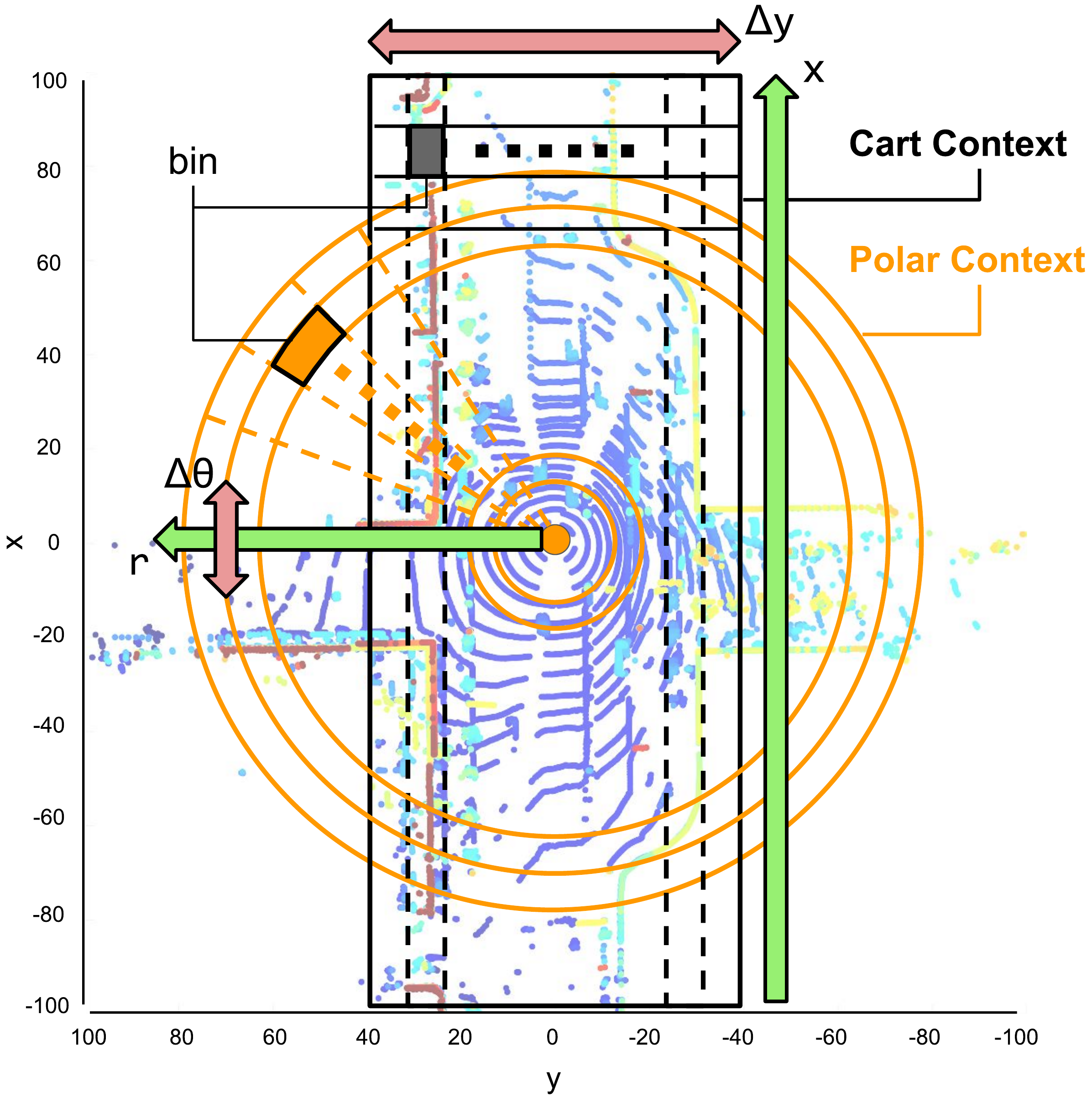}
    \label{fig:scdpartition}
    } \ \ \ \ \ \ \ \
    \centering
    \subfigure[\ac{SCD}s of \subref{fig:scdpartition}]{%
      \includegraphics[width=0.5\textwidth, trim = 0 -50 0 0, clip]{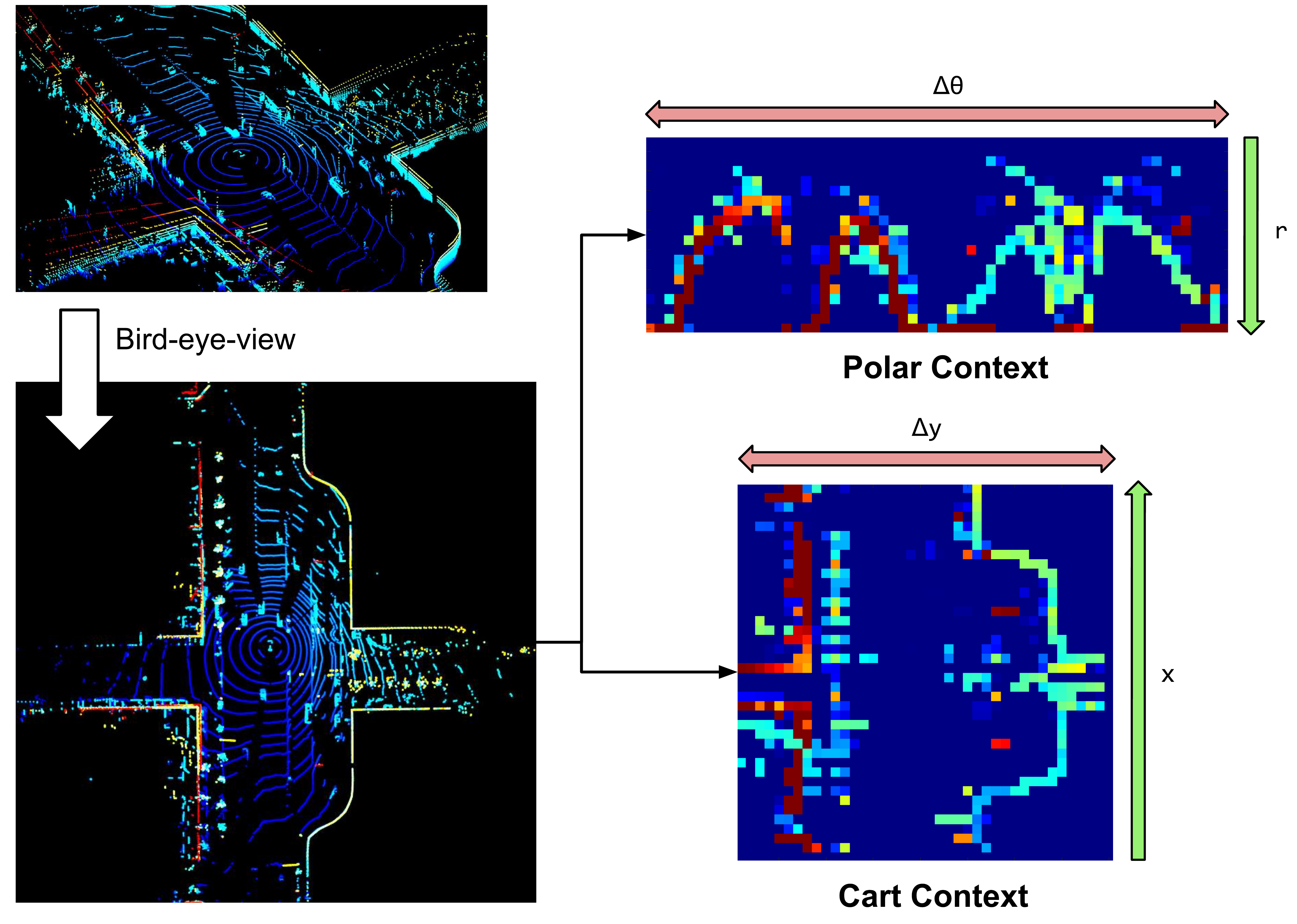}
      \label{fig:scdexample}
    }
  \end{minipage}

  \caption{\subref{fig:scdpartition} Sample point cloud and \subref{fig:scdexample} the associated \ac{SCD}s. In \subref{fig:scdpartition}, the yellow and gray color-filled entities depict bins for PC and CC, respectively. The red arrow represents the aligning axis. The green arrow represents the retrieval axis. In \subref{fig:scdexample}, each bin color indicates the maximum height in a bin; red is high (e.g., \unit{10}{m}), blue is low (e.g., \unit{0}{m}). In \subref{fig:scdexample}, the PC (top) has the parameters (20, 60, \unit{[0, 80]}{m}, \unit{[0, 360]}{\degree}), and the CC (bottom) has the parameters (40, 40, \unit{[-100, 100]}{m}, \unit{[-40, 40]}{m}).}
  \label{fig:scd}
\end{figure*}

The resolution of the axes determines the resolution of the descriptor, which is the user parameter of the proposed method. The user parameters are denoted as
\begin{equation}
  \label{eq:param}
  \small
  (\Delta_{R}, \Delta_{A}, [R_{min},R_{max}], [A_{min},A_{max}]),
\end{equation}
where each component indicates an ordered set consisting of the resolution of the \ac{R-axis}, the resolution of the \ac{A-axis}, the range of the \ac{R-axis}, and the range of the \ac{A-axis}, respectively. Sample parameter sets and their \ac{SCD} are given in \figref{fig:scdexample}. As will be discussed in \secref{sec:res}, coarse discretization implicitly reduces the influence of dynamic objects, noisy local structures, and computational cost.

\subsubsection{Independence of the Input Modality}
\label{sec:inputdata}

The partitioning is independent of the measurement's distribution or data type (e.g., a \ac{BEV} image, voxels, or 3D points). Therefore, our descriptor is generic with regard to any range measurements. The descriptor representation covers not only 3D point clouds but also other range sensors such as radar \cite{kim2020mulran} by selecting a proper bin encoding function in \secref{sec:bef}.


\subsection{Bin Encoding Function}
\label{sec:bef}

We denote a single disjoint section partitioned by the aligning and retrieval axes as a \textit{bin}. A single bin includes a subset of a robot sensor measurement ($Z_{ij} \in Z$), where the $i$ and $j$ indicate the \ac{A-axis} and \ac{R-axis} indexes, respectively. The bin may be empty, $Z_{ij} = \emptyset$, when no range data falls into the bin, in which case we assign a value of $0$ to that bin.

For each subset of measurement $Z_{ij}$ for bin $(i,j)$, we assign a representative value using a \textit{bin encoding function} $\psi(\cdot)$. The bin encoding function should be able to encapsulate the subset of the raw data in order to make the descriptor discernable and robust to the \bl{nuisances} (\tabref{tab:taxonomy}).

\begin{definition}{ }
  A \textit{bin encoding function}, $\psi : Z_{ij} \rightarrow \mathbb{R}$, is invariant to the internal factors and independent of sensor specifications.
  \label{req:bef}
\end{definition}

%
%

Following our previous work \cite{kim2018scan}, we propose to assign the maximum height of 3D points within a bin. The intuition behind this selection stems from an urban planning concept called \textit{isovist} \cite{benedikt1979take, KIM201974}. In this concept, the maximally visible structure and its visible volume's polygon shape decide the use of a place and make a place discernable. Focusing on the maximum height instead of structural shape eliminates the sparsity variation caused by the sensing resolution, range, and object size. Notably, any other function that meets the above requirement could be used as the encoding function. For the example of an \ac{FMCW} radar \cite{kim2020mulran}, the raw radar intensity value was adopted. Some follow-up studies of our previous work \cite{kim2018scan} leveraged LiDAR intensity \cite{wang2020intensity}, interpolated intensity \cite{kim2020mulran}, and difference in the height of 3D points \cite{mo2020place}. \bl{We note that heterogeneous LiDAR place recognition in situations where a mapper and a localizer are different (e.g., LiDAR's mounted height varies) is beyond this paper's scope because it does not obey the above requirement.}


\subsection{Scan Context Descriptor}
\label{sec:scd}

After the \ac{ROI} partitioning (\secref{sec:roipart}) and bin encoding (\secref{sec:bef}), each bin contains a representative feature summarizing the data within the bin (i.e., the maximum height for \ac{SCD}). We accumulate these bin values into a matrix form to complete a 2D descriptor for a place; the rows and columns of the matrix correspond to the retrieval axis and the aligning axis. The resulting descriptor can be understood as the contour of the skyline of the surrounding structures. Depending on the coordinate selection, we name the resulting 2D descriptor as \textit{Polar Context (PC)} or \textit{Cart Context (CC)}.

\subsubsection{Polar Context (PC)}
\label{sec:pc}

When the polar-coordinate \ac{ROI} is used, we name the resulting \ac{SCD} as a \textit{\ac{PC}}. The \ac{PC} is designed for rotation-invariant place recognition (e.g., revisit in the reversed direction) because the rotational variation corresponds to column-wise shifts.

\subsubsection{Cart Context (CC)}
\label{sec:cc}

Similarly, using Cartesian \ac{ROI} partitioning yields a \ac{SCD} called a \textit{\ac{CC}}. In a \ac{CC}, lateral translation is reflected as column-wise shifts; thus the \ac{CC} can handle lateral variation, including a revisit with lane changes.

Each \ac{SCD} has its own invariance for tackling the internal factors. \ac{PC} and \ac{CC} allow one dimension for the \ac{A-axis} and may be limited when rotation and translation occur simultaneously. To cope with this, we propose hallucinating the \ac{R-axis} to achieve robustness in both directions (\secref{sec:scdaug}).

\subsection{Distance between \ac{SCD}s}
\label{sec:dist}

Next, we define the proximity between two places by the similarity score of the associated \ac{SCD}.

\subsubsection{Alignment Score}
\label{sec:alignscore}

As illustrated in \figref{fig:shiftcc}, if two \ac{SCD}s are acquired from the same place, then two descriptors should contain consistent contents within a matrix but may reveal a column order difference. To measure similarity, therefore, we should examine the sum of the column-wise co-occurrences using the cosine similarity between two descriptors. This column-wise comparison is particularly effective for dynamic objects or partial noises. A cosine distance is used to compute a distance between two column vectors, $c_{Q}^{j}$ and $c_{M}^{j}$, at the same column index $j$. The distance between two descriptors is
\begin{equation}
  d(\textbf{f}_{Q}, \textbf{f}_{M}) = \frac{1}{N_{\text{A}}} \sum_{j=1}^{N_{\text{A}}} \left( 1 - \frac{c_{Q}^{j} \cdot c_{M}^{j}} {\Vert c^{j}_{Q} \Vert \Vert c^{j}_{M} \Vert } \right).
  \label{eq:imgDist}
\end{equation}
The subscripts $Q$ and $M$ indicate query and map places, where the descriptor's dimensions are $\textbf{f} \in \mathbb{R}^{N_R \times N_A}$. In addition, we divide the summation by the number of columns for normalization.

\subsubsection{Naive Column Alignment}
\label{sec:bestalign}

\begin{figure}[t!]
  \centering
  \includegraphics[width=0.95\columnwidth]{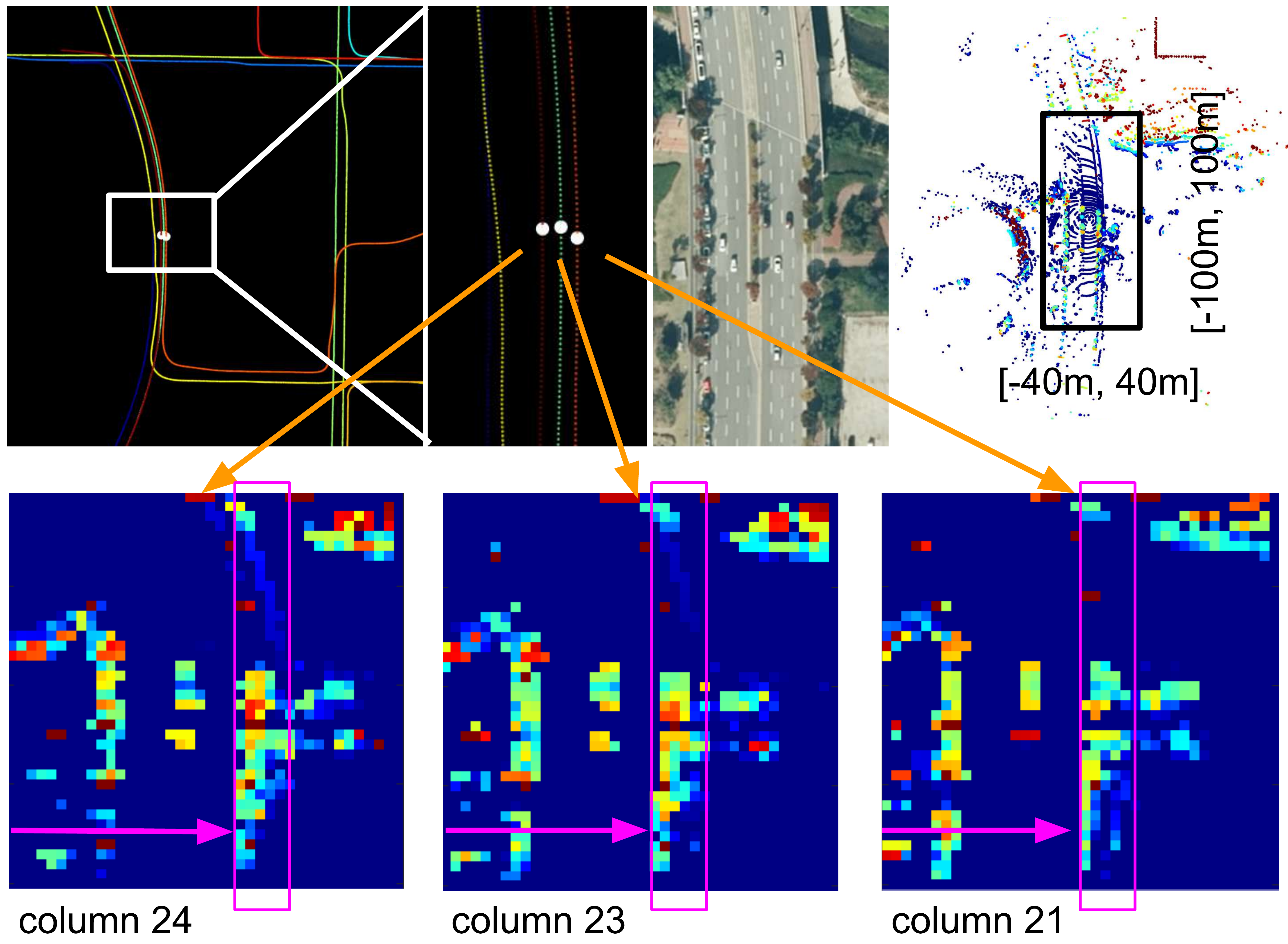}
  \caption{The three white dots in the top row indicate three sample nodes in the ground-truth trajectory. The vehicle visited the place three times while changing lanes. Below is the sample \ac{CC} corresponding to each node. Comparing the three sample \ac{CC}s, the contents are preserved within each column, while only the column orders are shifted among the nodes. The motion-induced change in the descriptor appears as a SCD column order shift in the descriptor space.}
  \label{fig:shiftcc}

\end{figure}

However, the column of the query \ac{SCD}, $\textbf{f}_{Q}$, may be shifted even in the same place (\figref{fig:shiftcc}). By simply shifting the order of the query descriptor while the $\textbf{f}_{M}$ is fixed, we can calculate distances with all possible column-shifted $\textbf{f}_{Q}$ and find the minimum distance. Then, the minimum distance of \equref{eq:imgDist} becomes our desired distance function $D(\cdot, \cdot)$ as
\begin{eqnarray}
  \label{eq:minImgDist}
  D(\textbf{f}_{Q}, \textbf{f}_{M}) &=& \min_{n \in [N_{\text{A}}]} d(\textbf{f}_{Q, n}, \textbf{f}_{M}) \label{eqn:minImgDist1} \ , \\ \nonumber
  n^{*} &=& \argmin_{n \in [N_{\text{A}}]} d(\textbf{f}_{Q, n}, \textbf{f}_{M}) \label{eqn:minImgDist2} \ ,
\end{eqnarray}
where $[N_{\text{A}}]$ indicates a set $\{1, 2, ..., N_{\text{A} \text{-} 1}, N_{\text{A}}\}$ and $\textbf{f}_{Q, n}$ is a \ac{SCD} whose columns are shifted from the original one by an amount $n$. The column-shift process aligns the rotational variance for \ac{PC} and lateral displacement for \ac{CC}.

\subsection{Sub-descriptors}
\label{sec:subdesc}

The abovementioned naive comparison over the full 2D descriptor is computationally expensive. To alleviate this cost we introduce two sub-descriptors. From the full 2D descriptor, \ac{SCD}, we extract 1D vectors by summarizing the descriptor in the row and the column direction. Each sub-descriptor plays a major role in place recognition and semi-metric localization.

\subsubsection{Retrieval key}
\label{sec:retrievalkey}

The first sub-descriptor introduced is the \textit{retrieval key}, $\textbf{v} \in \mathbb{R}^{N_R}$, a vector whose dimension is equal to the number of \ac{SCD} rows, $N_\text{R}$. Given any function that $f_{R}(\cdot)$ maps a column of a \ac{SCD} to a single real number, we \textit{squeeze} the column dimension of a \ac{SCD} by applying the $f_{R}(\cdot)$ for each row in a \ac{SCD}. Additionally, the following condition is required. For a given row $r$ of the \ac{SCD},
\begin{definition}{ }
  A \textit{retrieval key function} $f_{R} : r \rightarrow \mathbb{R}$ is permutation invariant.
\end{definition}
With this requirement, we can create a sub-descriptor that is unaffected by the column order; this means we can produce a consistent sub-descriptor independent of the internal factors of the nuisances (e.g., rotation or lane changes). Practically, we used the $L1$ norm for our experiments, but any other function can be used that maps a vector to a single real number by obeying the above requirement. The $L0$ norm was used in our previous work \cite{kim2018scan}.

\subsubsection{Aligning Key}
\label{sec:alignkey}

Similarly as with the retrieval key, we introduce the \textit{aligning key}  $\textbf{w} \in \mathbb{R}^{N_A}$ as another sub-descriptor of the \ac{SCD}, which is a vector whose dimension is equal to the number of \ac{SCD} columns $N_\text{A}$. Although no requirement is needed for the aligning key, we adopted the same $L1$ norm when summarizing a column.

\section{Three-stage Place Recognition}
\label{sec:alg}

Our place recognition algorithm consists of three parts: (\textit{i}) place retrieval using a \textit{retrieval key}, (\textit{ii}) semi-metric localization via pre-alignment using an \textit{aligning key}, and (\textit{iii}) full SCD comparison for potential refinement and localization-quality assessment.

\subsection{Place Retrieval using a Retrieval Key}
\label{sec:pr}

Existing widely adopted solutions leveraged past trajectory or motion uncertainties to reduce the search space \cite{dube2017segmatch, Chen2019OverlapNetLC}. Differing from them, we pursue global localization without prior knowledge. We solely rely on the descriptor itself while minimizing computational costs from global search by introducing sub-descriptors.



Using all extracted retrieval keys in the map, we construct a k-d tree for fast search and retrieve the closest place in terms of the retrieval key. Potentially, the top $k$ candidate indexes then may be retrieved to be verified at the full \ac{SCD} comparison phase. Interestingly, we empirically found that using only the best candidate ($k=1$) yields meaningful performance, outperforming the case using multiple candidates. A discussion on the candidate set size ($k$) will be presented in \secref{sec:topk}. As a result of the tree search, we topologically retrieve the corresponding map place for the query.

\begin{figure*}[!t]
  \centering
  \subfigure[\textit{Rot + Lat} revisit]{%
    \includegraphics[width=0.28\textwidth]{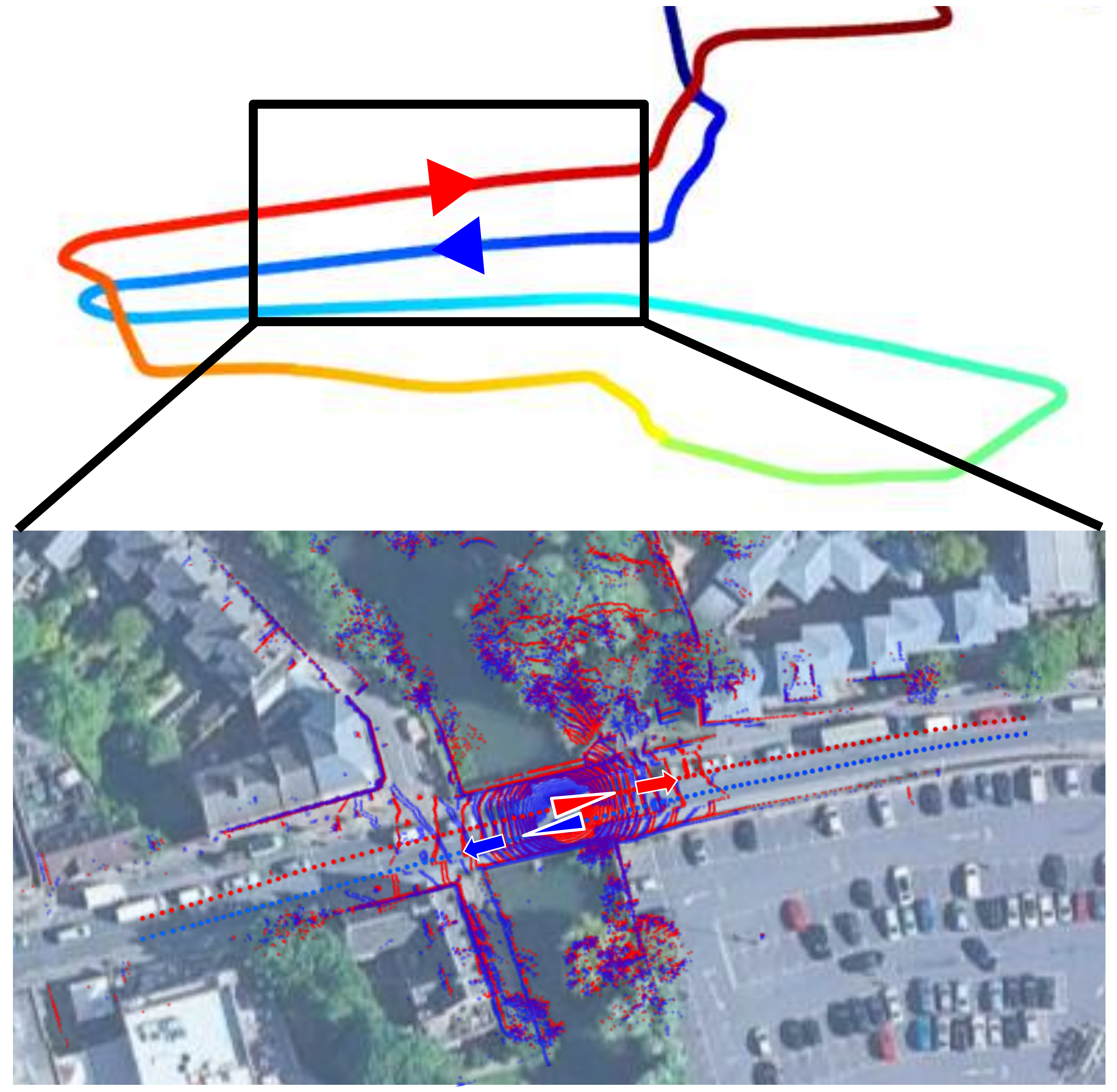}
    \label{fig:scdaug1}
  } \ \
  \subfigure[Polar Context Augmentation ]{%
    \includegraphics[width=0.42\textwidth, trim = 0 -40 0 20, clip]{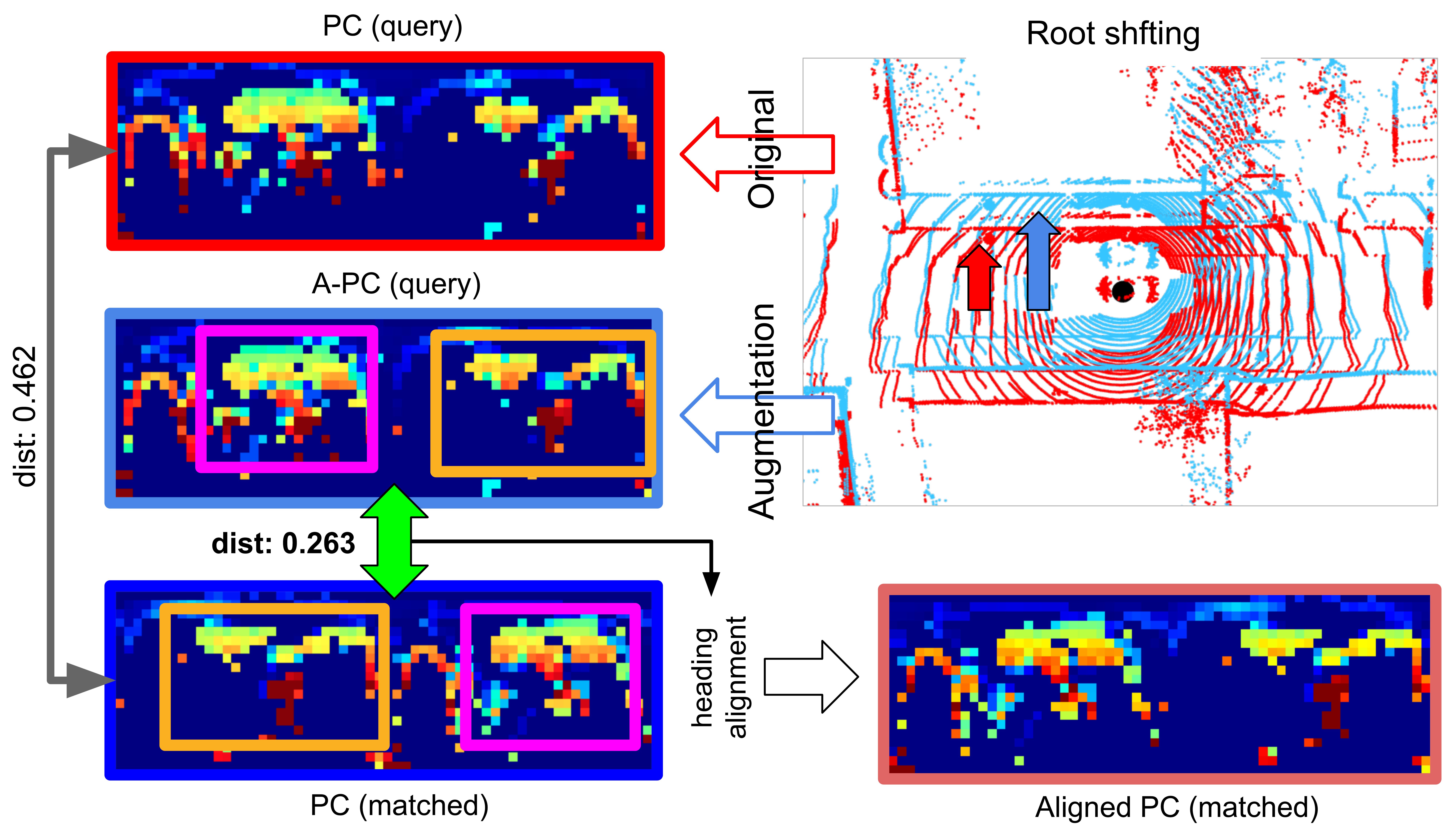}
    \label{fig:scdaug2}
  } \ \
  \subfigure[Cart Context Augmentation ]{%
    \includegraphics[width=0.24\textwidth, trim = 0 -70 0 0, clip]{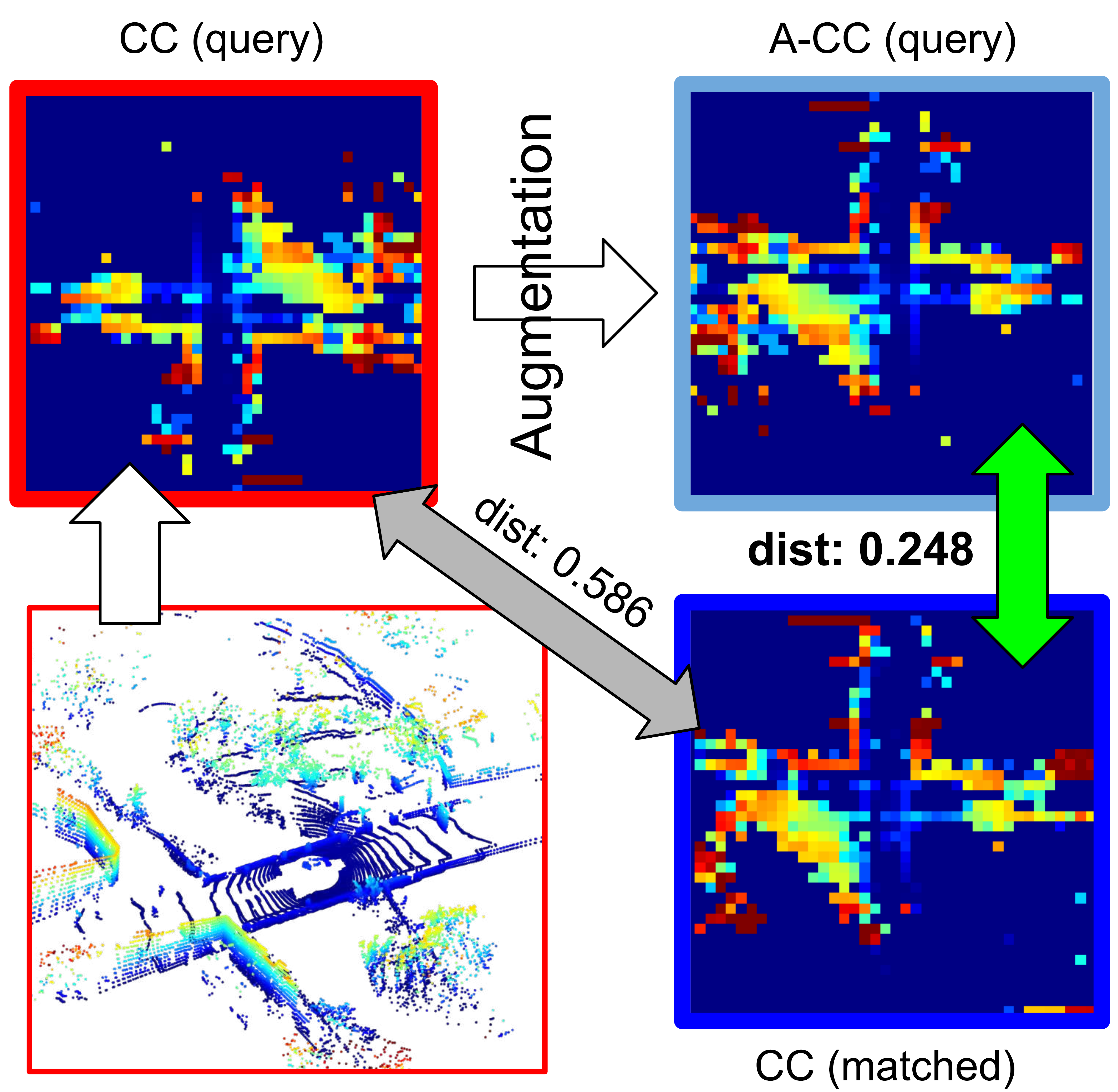}
    \label{fig:scdaug3}
  }%

  \caption{Illustration of the \ac{SCD} augmentation in \secref{sec:scdaug}. \subref{fig:scdaug1} The sample from \texttt{Oxford 2019-01-15-13-06-37} shows the revisit case with both rotational and translational change. \subref{fig:scdaug2} Polar Context augmentation includes explicit re-calculation of the descriptor by changing the vehicle's center pose. The original (red) pose based descriptor shows a larger distance than the shifted pose-based descriptor does. Only a sinlge virtual vehicle case is shown for visualization. Please be noted PC can recognize a place even under viewpoint changes (i.e., switched colored boxes in \figref{fig:scdaug2}). \subref{fig:scdaug3} Cart Context augmentation consists of simple sequential flips. Similarly, as in \ac{PC}, the augmented descriptor shows a closer distance to the map than the original descriptor.}
  \label{fig:scdaug}
\end{figure*}


\subsection{Semi-metric Localization using an Aligning Key}
\label{sec:prealign}

Given a retrieved candidate place, the typical \ac{SLAM} framework would proceed to metric-level localization by finding the relative pose between the query and candidate place recognized by the place retrieval module. Well-known approaches would include ICP and its variants, which compare two scans to find the optimal pose, minimizing an alignment cost. Despite their popularity, these metric localization methods may suffer local minima and required a good initial guess.

In the second phase of our place recognition algorithm, we exploit the aligning key and determine the partial relative pose through the pre-aligning phase. The naive brute-force version of the alignment \equref{eqn:minImgDist1} is computationally proportional to the number of columns $N_\text{A}$, which is heavier than the simple and frequently used $L2$ norm. We propose conducting brute-force aligning by using query and target aligning keys, instead of using the full \ac{SCD}s. The pre-alignment using the aligning key procedure is formalized as
\begin{equation}
  \hat{n}^{*} = \argmin_{n \in [N_{\text{inv}}]} d_{\textbf{w}}(\textbf{w}_{Q, n}, \textbf{w}_{M}) \ ,
  \label{eq:prealign1}
\end{equation}
where $\hat{n}^{*}$ is the estimated shift for the best alignment between the query and target \ac{SCD}s. We simply propose using $d_{\textbf{w}}$ as the $L2$ distance between two vectors. This computed column shift $\hat{n}^{*}$ can serve as a good initial value for further localization refinement such as \ac{ICP}. The evaluation of this initial guess is given in \secref{sec:icp}.

\subsection{Full Descriptor\bl{-based False Positive Rejection}}
\label{sec:fullscd}

The final step of place recognition is to compare the full \ac{SCD} \bl{to reject the potential false positive. As will be shown in \secref{sec:topk}, using a full descriptor may deteriorate the spatial discernibility.} Using the previously computed initial column shift $\hat{n}^{*}$, the original search space in \equref{eqn:minImgDist1} is shrinked to only the neighborhood $\mathcal{N}(\cdot)$ of the \textit{pre-aligned} shift $\hat{n}^{*}$.
\begin{equation}
  D(\textbf{f}_{Q}, \textbf{f}_{M}) = \min_{n \in \mathcal{N}(\hat{n}^{*})} d(\textbf{f}_{Q, n}, \textbf{f}_{M})  \ ,
  \label{eq:prealign2}
\end{equation}
This reduced search space may be insecure when the variation over columns is poor, when the upper vertical \ac{FOV} is low \cite{geiger2012we} when our maximum height bin encoding function hardly makes a diversified distribution. This can be overcome by developing a more discerning bin encoding function. However, as will be shown in experiments (see \secref{sec:exp}), we found that even an extremely tight choice of neighbor, $\mathcal{N}(\hat{n}^{*}) = \{\hat{n}^{*}\}$ (i.e., assuming the pre-aligning as the best alignment) is empirically enough and outperforms other methods.

Finally, we go over the $k$ candidates proposed by the k-d tree and search for candidates satisfying an acceptance threshold to select it as the revisited place.
\begin{equation}
  {c}^{*} = \argmin_{{c}_{k} \in \mathcal{C}} { D(\textbf{f}_{Q}, \textbf{f}_{M}^{c_k}) }, \ \text{s.t} \ D < \tau \ ,
  \label{equ:LoopFound}
\end{equation}
where $\mathcal{C}$ is the candidate index set extracted from the k-d tree, $\tau$ is the acceptance threshold, and $c^{*}$ is the index of the recognized place. Because we use $k=1$, this full descriptor similarity score performs as the validity check to confirm that $D < \tau$ before accepting the candidate as the correct match.

\subsection{Augmentation of the Scan Context Descriptor}
\label{sec:scdaug}

Because we construct a descriptor from the \ac{BEV}, the dominant motion complexity is reduced to 3-\ac{DOF} which is then summarized in a 2D descriptor. This indicate that both descriptors are deficient in certain \ac{DOF}. For example, \ac{PC} is written in the polar coordinates and loses the translational component; \ac{CC} is described in the Cartesian coordinates and lacks the rotational component. This deficiency is critical when revisit occurs in a combined motion. A typical example would be revisiting in a reversed route from the opposite lane. To overcome this limitation and impose robustness along the fixed axis, we created virtual \ac{SCD}s to augment a place, thereby achieving pseudo-invariance along the deficient direction.

\subsubsection{Augmented PC (A-PC)} We aimed to cover lane changes (\unit{2}{m} spaced lanes) and a reversed route (\unit{180}{\degree} heading change). \bl{During this augmentation process}, a \ac{PC} is synthetically duplicated by assuming virtual lateral displacement. Our particular interest is lane change, and we synthetically considered two virtual vehicle positions that are laterally \unit{2}{m} apart. Two additional \ac{A-PC}s are generated with respect to these virtual vehicle poses and root-shifted point clouds. This \textit{root shifting} is the same way as in our previous work \cite{kim2018scan}.

\subsubsection{Augmented CC (A-CC)} For \ac{CC}, the augmentation is as simple as a double flip on both axes. The lacking rotational component should encompass lane changes, and we flip the descriptor on both axes to create the \ac{A-CC}.

Both the \ac{A-PC} and the \ac{A-CC} are illustrated in \figref{fig:scdaug}. The augmented descriptors' place index is assigned as identical to its original one. For matching, empirically, we found maintaining a single k-d tree containing both original and augmented keys outperforms using multiple k-d trees.

\begin{table}[!b]
 \vspace{-3mm}
 \caption{Implementation details.}
  \centering
  \resizebox{\linewidth}{!}
  {
  \begin{tabular}{l|c|c}
  \hline
  Parameter     &  PC / A-PC  & CC / A-CC\\
  \toprule[1.2pt]
  Down sampling     & \multicolumn{2}{c}{$0.5 \times 0.5 \times 0.5 \meter^3$}\\\hline
  ROI               & $([\unit{0}{m},\unit{80}{m}],[\unit{0}{\degree},\unit{360}{\degree}])$
                    & $([\unit{-100}{m},\unit{100}{m}],[\unit{-40}{m},\unit{40}{m}])$\\\hline
  Resolution        & $20\times60$ ($\unit{4}{m},6^\circ$)&  $40\times40$ ($\unit{5}{m},\unit{2}{m}$)\\\hline
  Candidate \# ($k$)& \multicolumn{2}{c}{1}\\\hline
  \bl{Augmentations} \# & 2 & 1\\\hline
  Augmentation  & $\pm2\meter$ root shiftings in the lateral direction & double flip\\
  \bottomrule[1.2pt]
\end{tabular}
}
\vspace{5pt}
\label{tab:impl}
\end{table}


\begin{figure*}[!t]
  \centering

  \includegraphics[width=0.99\textwidth]{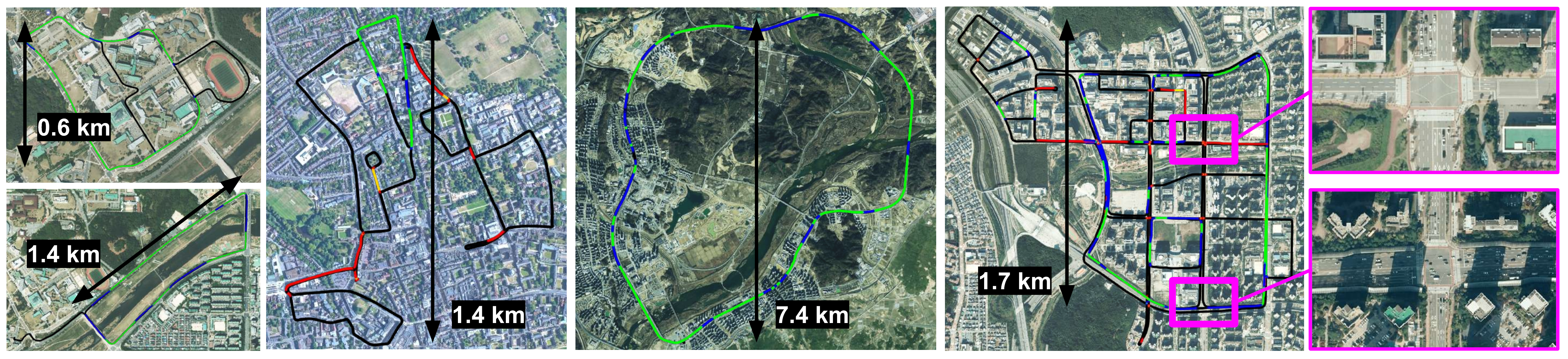}
  \caption
  {
    The dataset trajectories overlaid on each aerial map. The first column shows \texttt{KAIST 03 (MulRan)} and \texttt{Riverside 02 (MulRan)}, followed by \texttt{Oxford Radar RobotCar}, the \texttt{Sejong} sequence in \texttt{MulRan}, and \texttt{Pangyo (NAVER LABS)}. The magenta boxes on the right show the wide roads in the \texttt{Pangyo} sequence of the \texttt{NAVER LABS} dataset. The total length of each sequence and their characteristics are summarized in \tabref{tab:dataset}.
  }
  \label{fig:datasetviz}
\end{figure*}

\begin{table*}[!t]
  \caption
  {
    The datasets' details with respect to the invariance taxonomy in \tabref{tab:taxonomy}. 
  }
  \vspace{-1mm}
  \resizebox{\linewidth}{!}
  {
      \begin{tabular}{cc|cccccccc}

        \multicolumn{2}{c}{\textit{Induced variance}} &  &  & \multicolumn{1}{c}{\textit{T}}      & \multicolumn{1}{c}{\textit{R}}              & \multicolumn{1}{c}{\textit{D}}        & \textit{SV}        & \textit{NR}     & \textit{FOV} \\\hline

        \multicolumn{1}{c}{\textbf{Dataset}} & \multicolumn{1}{c|}{\textbf{Sequence}} &  \thead{\textbf{Path length (km)} \\ (\# revisits / \# total)} &  \thead{ \textbf{Avg/Max Speed} \\ (km/h)} & \thead{\textbf{Non-same direction Revisits} \\ (ratio)} & \textbf{Lane Changes} & \multicolumn{1}{c}{\textbf{Dyn Objs}} & \textbf{Inter-session} & \thead{\textbf{Sensor} \\ (\# rays)}  & \thead{\textbf{HFOV} \\ (\degree)} \\\hline

        \toprule[1.2pt] 

        \multirow{3}{*}{KITTI}  & \multicolumn{1}{c|}{00}   &   \thead{3.71 (852 / 4541) }         &   28.5 / 47.9    &   Y (3\%, 22 / 852)                  &   N                     & $\bigstar$              & N                  & 64           & Full         \\
                                & \multicolumn{1}{c|}{08}   &   \thead{3.21 (102 / 2377) }          &   27.5 / 46.7    &   Y (100\%, 102 / 102)    &   $\bigstar$            & $\bigstar$         & N                  & 64           & Full         \\

        \midrule[0.2pt] 

        \multirow{4}{*}{MulRan} & KAIST 03            &  \thead{6.25 (2055 / 4224) }            &  26.3 / 54.4   &   N (0\%, 0 / 2055)              &   N                                          &                  $\bigstar$                     & N                  & 64           & 290          \\
                                & Riverside 02        &  \thead{6.61 (2174 / 4870) }          &    35.6 / 66.6   &   N (0\%, 0 / 2174)              &   $\bigstar$$\bigstar$$\bigstar$                                        &       $\bigstar$$\bigstar$$\bigstar$                                & N                  & 64          & 290          \\
                                & Sejong 02 to Sejong 01     &  \thead{23.16 (17907 / 18090) }        &      40.0 / 67.4   &   N (0\%, 0 / 17907)              &   $\bigstar$$\bigstar$                                           &       $\bigstar$$\bigstar$                                & Y                  & 64          & 290 \\ 
      
        \midrule[0.2pt] 

        \multirow{4}{*}{ \thead{Oxford \\Radar RobotCar} } & 2019-01-11-13-24-51&  \thead{ 9.93 (2117 / 8192) }    &  24.4 / 42.2   &  Y (43\%, 901 / 2117)      &      $\bigstar$                                        &    $\bigstar$$\bigstar$                                 & N                 & 32           & Full         \\
                                & \thead{2019-01-15-13-06-37 \\ to 2019-01-11-13-24-51  }             &   \thead{ 8.89 (7391 / 7391) }    &  25.1 / 50.8   &  N (0\%, 0 / 7391)           &    $\bigstar$                                          &             $\bigstar$$\bigstar$                          & Y                  & 32           & Full         \\

        \midrule[0.2pt] 

        NAVER LABS              & Pangyo              &  \thead{31.37 (7025 / 21648) }          &   23.8 / 41.5   &  Y (29\%, 2021/7025)             &   $\bigstar$$\bigstar$$\bigstar$                                                & $\bigstar$$\bigstar$$\bigstar$             & N                  & 32               & Full         \\

        \bottomrule[1.2pt] 

    \end{tabular}
  }
  \vspace{1mm}
  \label{tab:dataset}


\end{table*}

\subsection{Computational Complexity}
\label{sec:complexity}

Among all of the introduced modules, the neighbor search is the most computationally demanding. Tree construction consumes periodic resources and the add-on augmentation step requires increased time computation proportional to the number of the augmentations. As will be shown in \secref{sec:timecost}, the number of augmentations and periodic tree maintenance are negligible. Even the main computational bottleneck of the retrieval module is extremely lightweight.

Naive descriptor comparison, as described in \equref{eq:imgDist} and \equref{eq:minImgDist}, requires the computation of $\mathcal{O}(N_A \cdot N_R \cdot N_A)$. This cost is substantially reduced by pre-alignment, as described in \equref{eq:prealign1} and \equref{eq:prealign2}, eliminating linear search through $N_A$ elements. The reduced computational cost becomes  $\mathcal{O}(N_A \cdot N_R \cdot 1)$. Approximating $N_A \sim N_R \sim N$, this reduction can be regarded as a reduction from $\mathcal{O}(N^3)$ to $\mathcal{O}(N^2)$ with the descriptor dimension $N$. For example, the \ac{CC} in \figref{fig:scdexample} is a square matrix with the format $N_A = N_R =N$.

\subsection{Implementation Details}
\label{sec:impl_detail}

The used parameters are listed as in \tabref{tab:impl}. Here, ROI and grid size determines the resolution. For example, $20\times60$ for PC indicates $80/20=\unit{4}{m}$ and $360/60 = 6^\circ$ resolution. Similarly, $40\times40$ for CC indicates $200/40=\unit{5}{m}$ and $80/40 =\unit{2}{m}$ resolution, for the \ac{R-axis} and the \ac{A-axis}, respectively. The discussion on parameter selection will be given in \secref{sec:discussion}.

\section{Dataset and Evaluation Criteria}
\label{sec:expsetup}


For the evaluation, we chose trajectories to cover broad revisit types including rotation and lateral changes. We describe the datasets and evaluation criteria below.

\subsection{Datasets}
\label{sec:dataset}

In total, eight sequences were selected from four publicly available datasets covering diverse environments: \texttt{KITTI Odometry} \cite{geiger2012we}, \texttt{MulRan} \cite{kim2020mulran}, \texttt{Oxford Radar RobotCar} \cite{RadarRobotCarDatasetICRA2020}, and \texttt{NAVER LABS}\footnote{https://hdmap.naverlabs.com/ and https://challenge.naverlabs.com/} datasets. The detailed characteristics of each sequence and the environment will be provided in the following subsections. The overlaid trajectories on the aerial map, as shown in \figref{fig:datasetviz}, illustrate the trajectory shape, scale, and surrounding environments (excluded well-known \texttt{KITTI} sequences). The details of the four datasets are summarized in \tabref{tab:dataset}.


\subsubsection{KITTI}

{KITTI Odometry}\footnote{http://www.cvlibs.net/datasets/kitti/eval\_odometry.php} \cite{geiger2012we} is the most widely used dataset for LiDAR place recognition \cite{he2016m2dp, dube2017segmatch, yin2018locnet, oreos19iros, Chen2019OverlapNetLC}. This dataset provides 64-ray LiDAR scans (Velodyne HDL-64E) We selected two sequences, \texttt{00} and \texttt{08}, with a sufficient number of loops. Note that sequence \texttt{08} is only composed of reverse loops.

\subsubsection{MulRan}

The Multimodal Range Dataset (MulRan) \cite{kim2020mulran} was specifically designed to support place recognition evaluation and contains a large number of loop events. This dataset provides 64-ray LiDAR scans (Ouster OS1-64) in 12 sequences covering a campus for a planned city. We chose three sequences: \texttt{KAIST}, \texttt{Riverside}, and \texttt{Sejong}.

\texttt{KAIST 03} is a campus environment with few dynamic objects and multiple well-distributed buildings. \texttt{Riverside 02} involves travel on roads along the riverside. This sequence includes few surrounding structures and many perceptually similar unstructured objects such as roadside trees, which are frequently repeated throughout the sequence. More critically, this sequence has multiple lane changes at the revisit phase (the blue parts in \figref{fig:riverside02_traj}), which enable us to quantitatively assess the methods' robustness under lateral changes. The third environment in MulRan, the \texttt{Sejong} sequence, encompasses the long circular route of a master-planned city called Sejong \cite{alicha-2014}. As a planned city, its environment reveals slowly varying structural changes even within a relatively short period of time. We chose \texttt{Sejong 01} and \texttt{Sejong 02} and examined the multi-session loop-closure capability and the robustness over a temporal gap (between June 2019 and August 2019).

\subsubsection{Oxford Radar RobotCar}

The \texttt{Oxford Radar Robot Car} \cite{RadarRobotCarDatasetICRA2020} dataset, which we simply call \texttt{Oxford}, is a radar extension of the Oxford RobotCar dataset \cite{RobotCarDatasetIJRR}. This extension provides range data from a \ac{FMCW} radar and two 32-ray 3D LiDARs (Velodyne HDL-32E) mounted at the left and right sides of the radar. For each place, we constructed a single point cloud by concatenating scans from the left and right LiDARs (their center is a new sensor coordinate) and used this newly generated scan for the evaluation. The sites of \texttt{Oxford} mostly have a maximum of two lanes and no expected heavy lateral displacement. Instead, the sequence contained reverse revisits occurring simultaneously with small lane changes (i.e., the red places in \figref{fig:oxfordjan11_traj}). This dataset enabled us to evaluate the robustness to concurrent rotation-and-lateral changes.

Among the repeatedly recorded sequences over the same site, we selected two sequences (\texttt{2019-01-11-13-24-51} and \texttt{2019-01-15-13-06-37}) whose INS and GPS signals were secured over the entire trajectory. The sequence \texttt{2019-01-11-13-24-51} was used for a intra-session place recognition validation as shown in \figref{fig:exp_oxford}. The selected sequences were also used to validate the inter-session place recognition performance, which is named \texttt{2019-01-15-13-06-37 to 2019-01-11-13-24-51} and is visualized in \figref{fig:oxfordmulti_traj}. We can see all global relocalizations (i.e., revisits) arose within the same direction.

\subsubsection{NAVER LABS}

The last evaluation sequence is a long single trajectory through highly urbanized environments, named \texttt{Pangyo}, from the \texttt{NAVER LABS} dataset\footnote{https://hdmap.naverlabs.com/ and https://challenge.naverlabs.com/} made by NAVER LABS. The long \unit{31}{km} sequence includes tall buildings, wide roads (the magenta boxes in \figref{fig:datasetviz}), and multiple revisits per place. More than half of the same-direction-revisits occurred in different lanes accompanied by rotation changes. We used \texttt{Pangyo} to validate a method's comprehensive performance and scalability.


\subsection{Correctness Criteria}
\label{sec:criteria}

The measure of the each place strongly depends on the applications and the target environment (e.g., indoors or outdoors). In this evaluation, we aimed to include changes of up to three lane (approximately \unit{8}{m}), which frequently occur in complex urban sites. By doing so, the robot recognizes a place even when revisiting occurs at a laterally separated location. Secondly, in \ac{SLAM} applications, coarse global loop detection typically followed by the pose regression module, generates a metric constraint between the query and the map. If the loop candidate is detected too broadly (e.g., \unit{25}{m} in \cite{suaftescu2020kidnapped}), then the accompanied fine localization module may fail. Considering these two aspects, we count the detected place as correct if a query place and a detected loop candidate place are less than \unit{8}{m} apart. We prepared 1$-$1.5m equidistant sampled measurements to avoid redundant frames during stop sections and to enable each place to contribute the same. The numbers of nodes for each sequence used for the evaluation are reported in \tabref{tab:dataset}.

\subsection{Evaluation Metrics}
\label{sec:evalmetric}

\begin{figure}[!t]
  \centering
  \subfigure[Trajectory and loop-closure event distribution]{%
    \includegraphics[width=.98\columnwidth, trim = 0 0 0 0, clip]{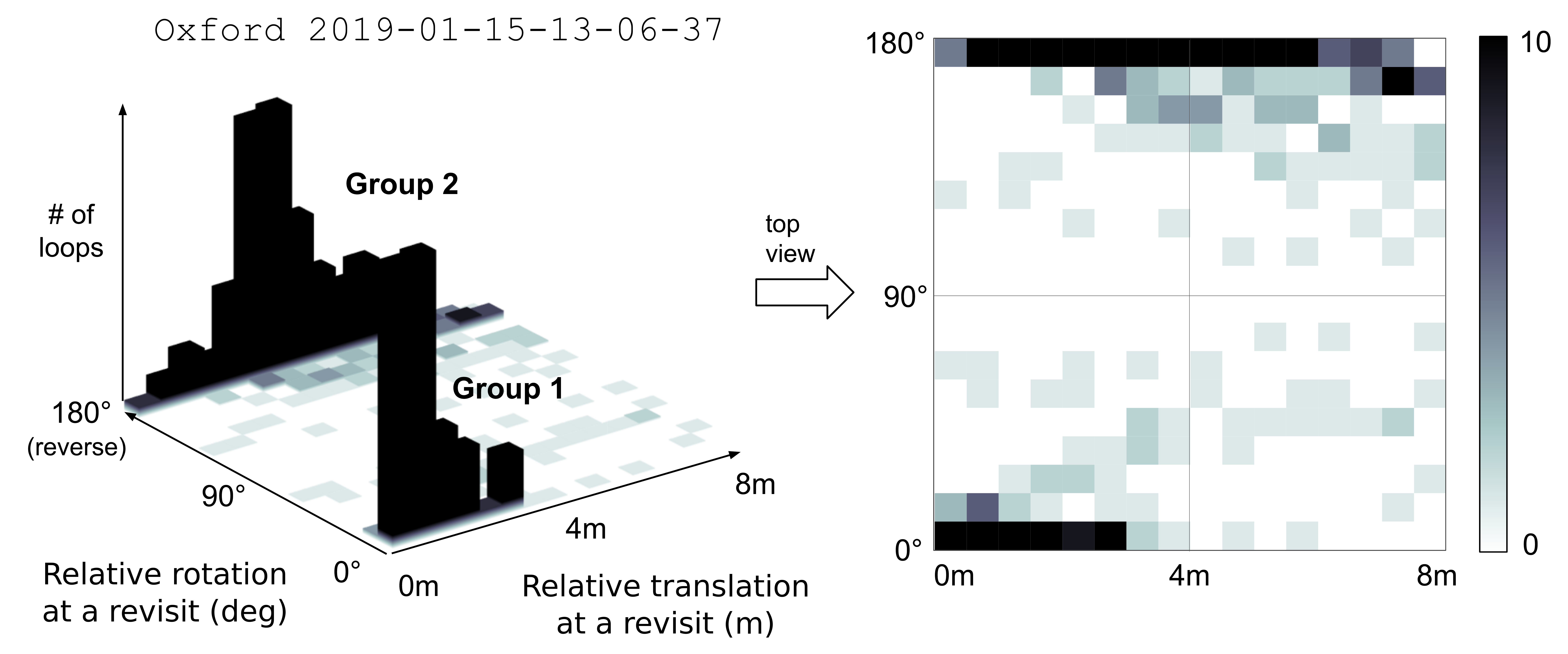}
    \label{fig:exp_toy_gt}
  }\\
  \subfigure[Simulation cases and their detected loops and KL-D]{%
    \includegraphics[width=.99\columnwidth, trim = 0 0 0 0, clip]{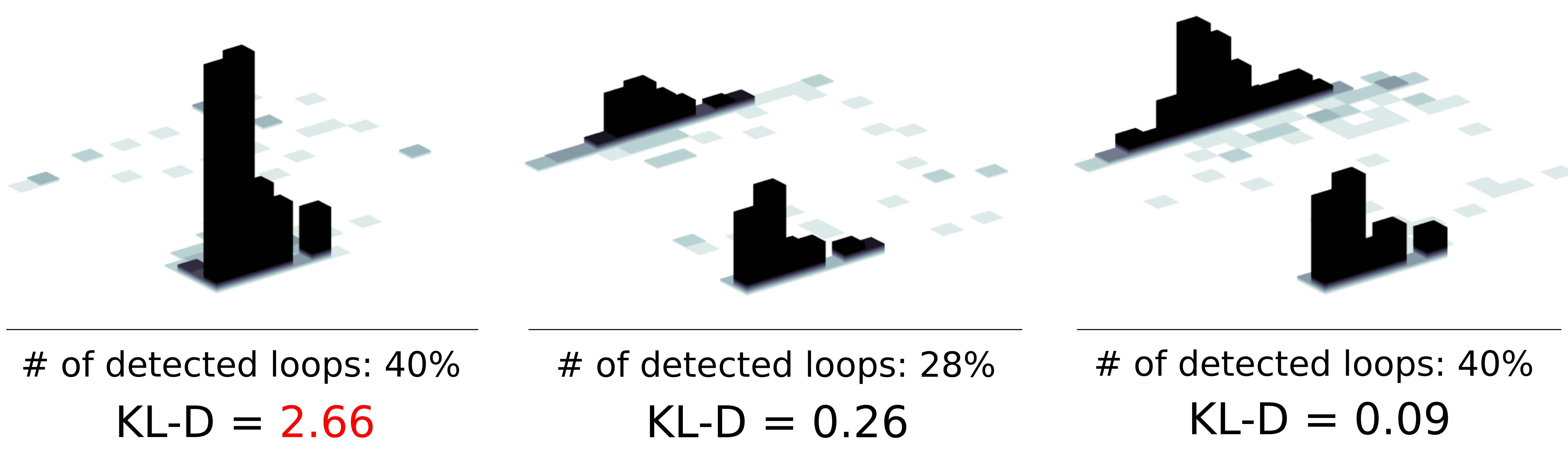}
    \label{fig:exp_toy_cases}
  }%
  \caption{\subref{fig:exp_toy_gt} True distribution of the loop-closures seen in a perspective view (left) and a top-down view (right). The grid size used in the visualization is (\unit{0.5}{m}, \unit{10}{}$^\circ$). The loop-closure events are majorly grouped into two. \subref{fig:exp_toy_cases} three sample algorithms showing different detected event distributions.}
  \label{fig:exp_toy}

\end{figure}


\subsubsection{Precision-Recall Curve} We used the precision-recall curve as a main evaluation metric \cite{lowry2015visual}. As argued in \cite{lowry2015visual}, for a place recognition system, increasing potential matches is important, even if a few false predictions occur \cite{sunderhauf2012switchable}. We also examined the maximum F1 score \cite{schutze2008introduction}, the harmonic mean of precision and recall, as our evaluation metric.

\subsubsection{Recall Distribution} We would like to note that the precision-recall curve may not fully reveal the performance toward loop-closure in a \ac{SLAM} framework. The spatial and temporal distributions of the loop-closure are essential for the \ac{SLAM}, while the precision-recall curve could be limited to measuring the distribution of place recognition. \textit{Not all recalls should be credited equally} from the point of view of \ac{SLAM} loop-closure. To value more distributed loop detections, we formulated the true revisits as the reference loop distribution and measuring \ac{KL} divergence against it.

As illustrated in \figref{fig:exp_toy_gt}, we constructed a histogram of the loop-closure event with respect to the translational and rotational variance between a nearest one in a map and a query pose. The sample revisit events collected from \texttt{Oxford 2019-01-15-13-06-37} contains two major groups. In this toy example, we simulated three algorithms showing different recall distributions and measured \ac{KL} divergence with respect to the ground-truth recall distribution. In \figref{fig:exp_toy_cases}, few loop-closures are found from the group 2 for the leftmost case. The other two showed better distributed loop-closure detections with respect to the internal factor variation, providing spatially unbiased localization performance. Even with smaller revisit detection, the middle case yielded better distribution showing lower KL-D value. During the evaluation, we show arrows to indicate that higher precision ($\uparrow$), higher F1 score ($\uparrow$), and lower KL-D ($\downarrow$) imply better performance.

Potentially, the Wasserstein distance (a.k.a. the earth mover's distance) or Jensen–Shannon Divergence could be the measure to use as also discussed in detail in \cite{wgan}. However, we chose to use \ac{KL}-D because we need to compare the relative distance between methods while having the GT distribution as the reference. Measuring relative information is favored over the symmetry. Here, we used the ground-truth loop-closure as the reference distribution and measure relative entropy against this reference.

\begin{figure}[!t]
  \centering
  \begin{minipage}{\columnwidth}
  \centering
  \subfigure[Revisit distribution (\texttt{KITTI 00})]{%
    \includegraphics[width=0.58\textwidth, trim = 0 -60 0 0, clip]{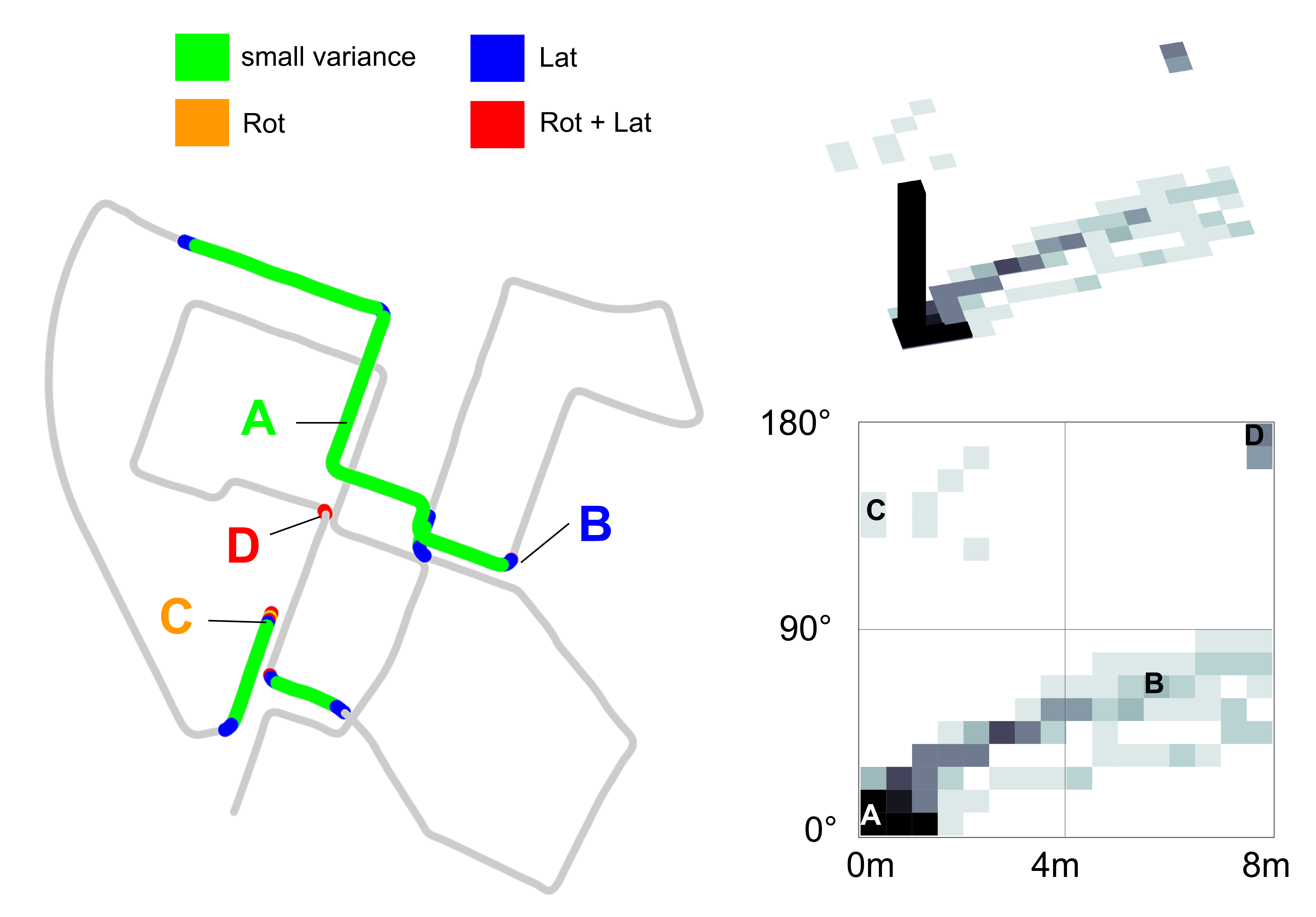}
    \label{fig:kitt00_traj}
  }%
  \subfigure[ PR curve ($\uparrow$) ]{%
    \includegraphics[width=0.35\textwidth, trim = 85 130 100 180, clip]{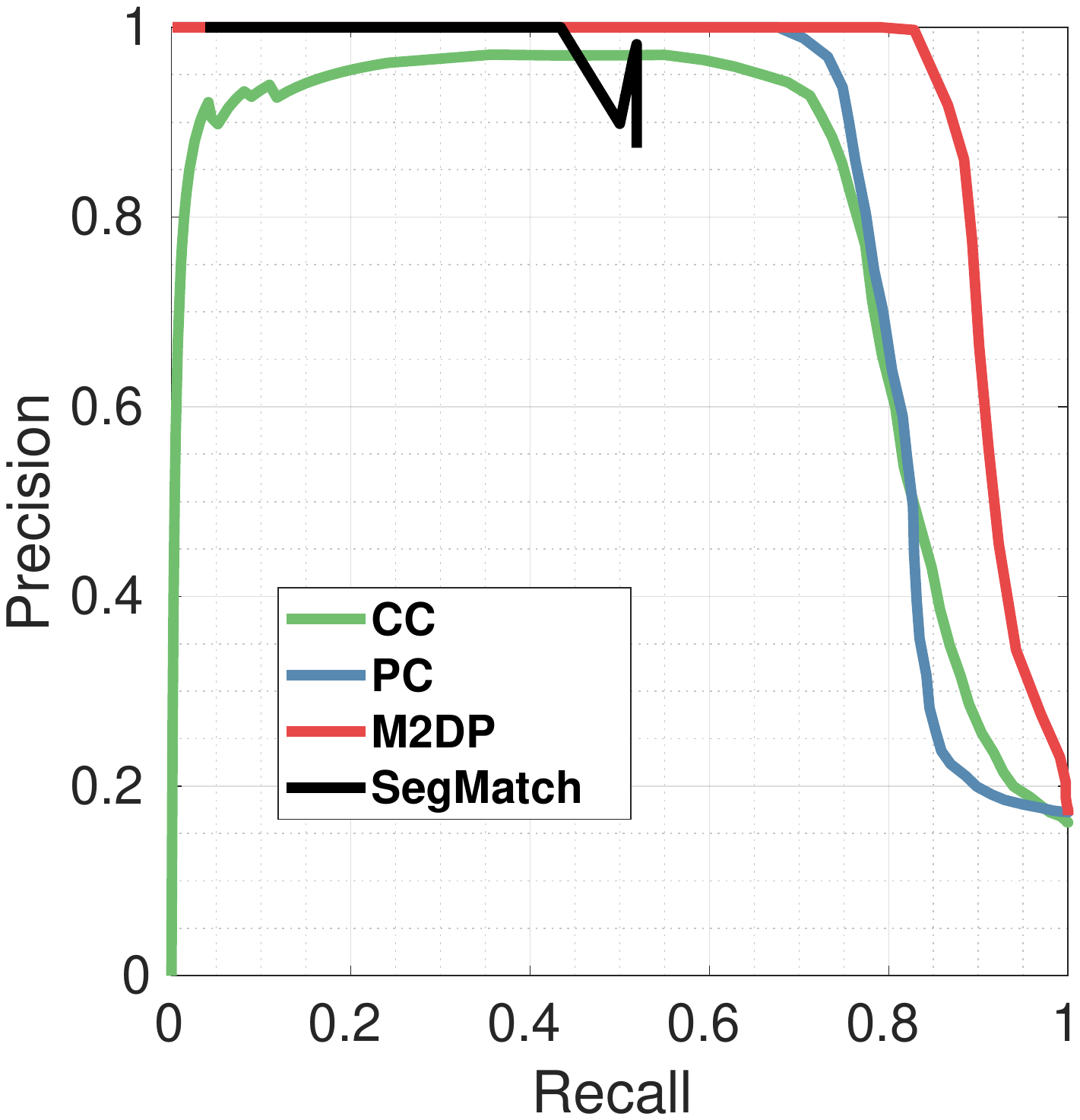}
    \label{fig:kitti00_pr}
  }\\
  \subfigure[Revisit distribution (\texttt{KAIST 03})]{%
    \includegraphics[width=.58\textwidth, trim = 0 -60 0 0, clip]{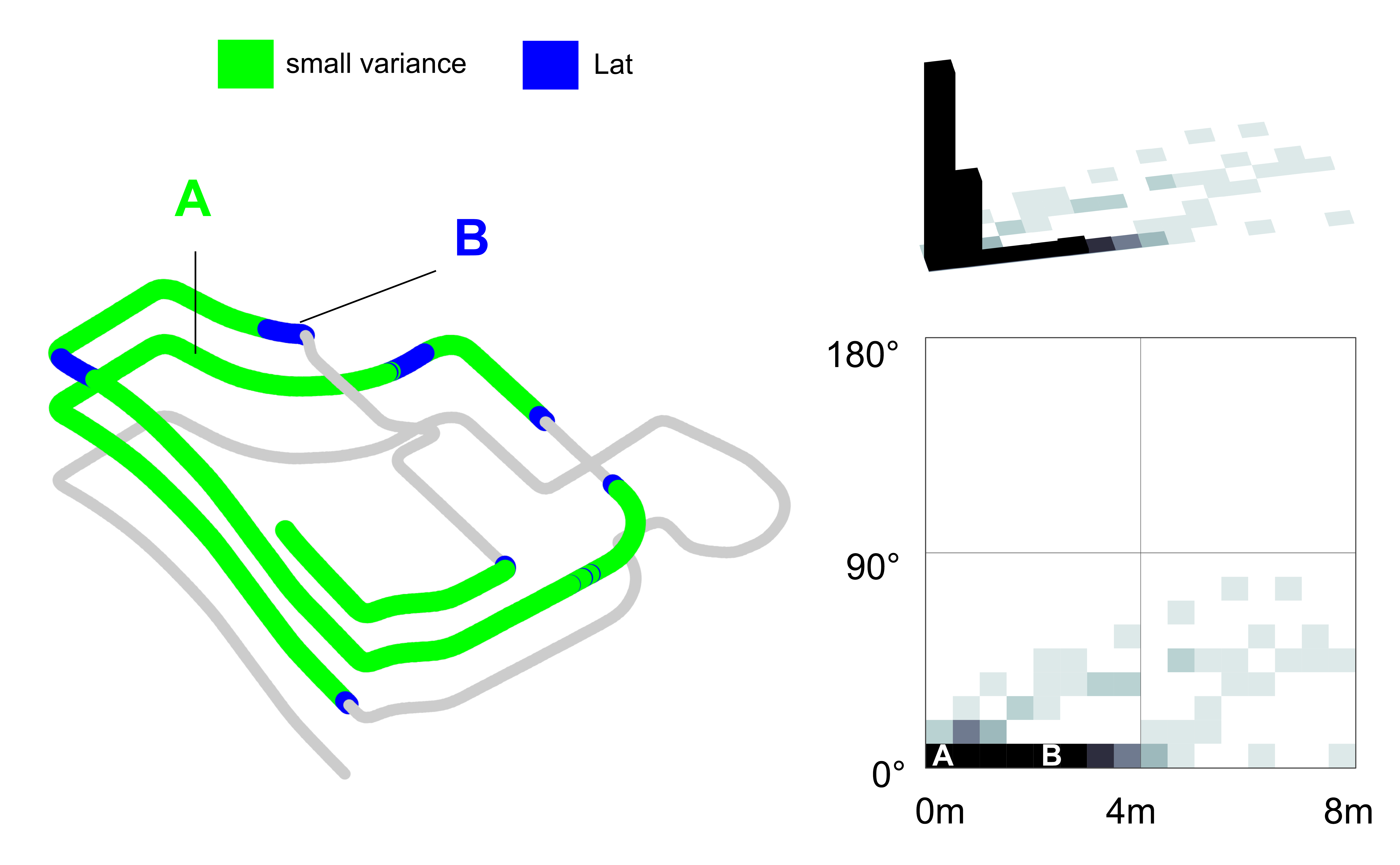}
    \label{fig:kaist03_traj}
  }%
  \subfigure[ PR curve ($\uparrow$)]{
    \includegraphics[width=0.35\textwidth, trim = 85 160 100 180, clip]{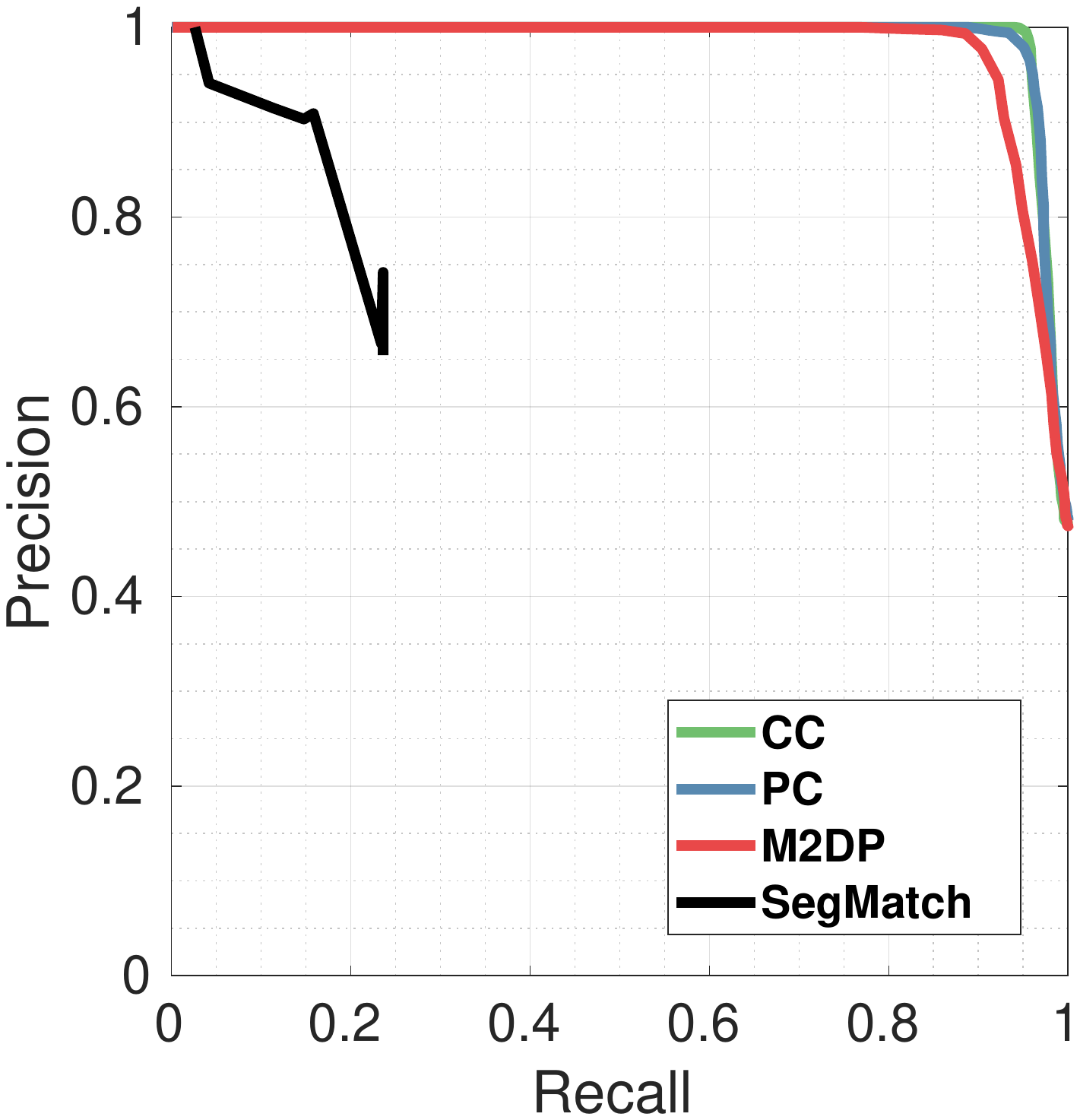}
    \label{fig:kaist03_pr}
  }%
  \end{minipage}
  \caption{
    (Left) The marked trajectories of the revisited places, (middle) the true revisit distributions, and (right) the PR curves. In \subref{fig:kitt00_traj} and \subref{fig:kaist03_traj}, the trajectories are color-coded by revisit types. Most of the loop-closure events are concentrated in the lower-left of the true-revisit distribution, revealing small rotational and translational variance. This concentrated distribution is depicted as a single peak in the perspective view.
  }
  \label{fig:exp_easy}


\end{figure}


\subsection{Comparison Targets}

We compared the proposed methods against two other methods: \textit{M2DP} and \textit{SegMatch}. All of the comparison targets are agnostic to sensor type (e.g., ray numbers) and run on CPU.

\subsubsection{SCD} We present the performance of \ac{PC} \cite{kim2018scan}, \ac{CC}, \ac{A-PC}, and \ac{A-CC}. For the proposed methods, we only retrieved a single candidate from the k-d tree ($k=1$). Downsample point cloud using a $\unit{0.5}{m}^3$ voxel is used to make a \ac{SCD} (\tabref{tab:impl}). The evaluation curves were acquired by changing the threshold of the \ac{SCD} distance.

\subsubsection{M2DP} Identical to our methods, M2DP \cite{he2016m2dp} only requires a point cloud from a single scan as an input. We followed the code and the parameters provided\footnote{https://github.com/LiHeUA/M2DP}, with one difference. We empirically found that applying \unit{0.1}{m} cubic voxel downsampling a priori boosts M2DP's performance, and we made this modification to secure better performance. The query descriptor is compared to all of the map descriptors in terms of Euclidean distance, which was used as a threshold.

\subsubsection{SegMatch} Among the three options in SegMatch \cite{dube2017segmatch}, we used the eigenvalue to describe a segment, which is the same as the author’s configuration designed for the \texttt{KITTI} dataset. We excluded the learning-based version, SegMap \cite{dube2020segmap}, because our method works on CPU and for a fair comparison. The evaluation curves for SegMatch are acquired by changing the segment feature distance threshold. Unlike the other global localization methods (ours and M2DP), SegMatch requires odometry information. During the evaluation, we leverage ground-truth to provide odometry. As will be seen, despite the exploitation of highly accurate odometry, SegMatch failed to overcome severe variance, while our method reliably localized without requiring any geometric prior. Not being a global descriptor as M2DP and SCD are, SegMatch only had a short range in its PR and DR curve. This is because the parameters in SegMatch are tuned to local segmentation and do not substantially affecting recall.

\begin{figure}[!t]
  \centering
  \begin{minipage}{0.37\columnwidth}
    \centering
    \subfigure[SegMatch]{%
      \includegraphics[width=\textwidth, trim = 160 120 110 90, clip]{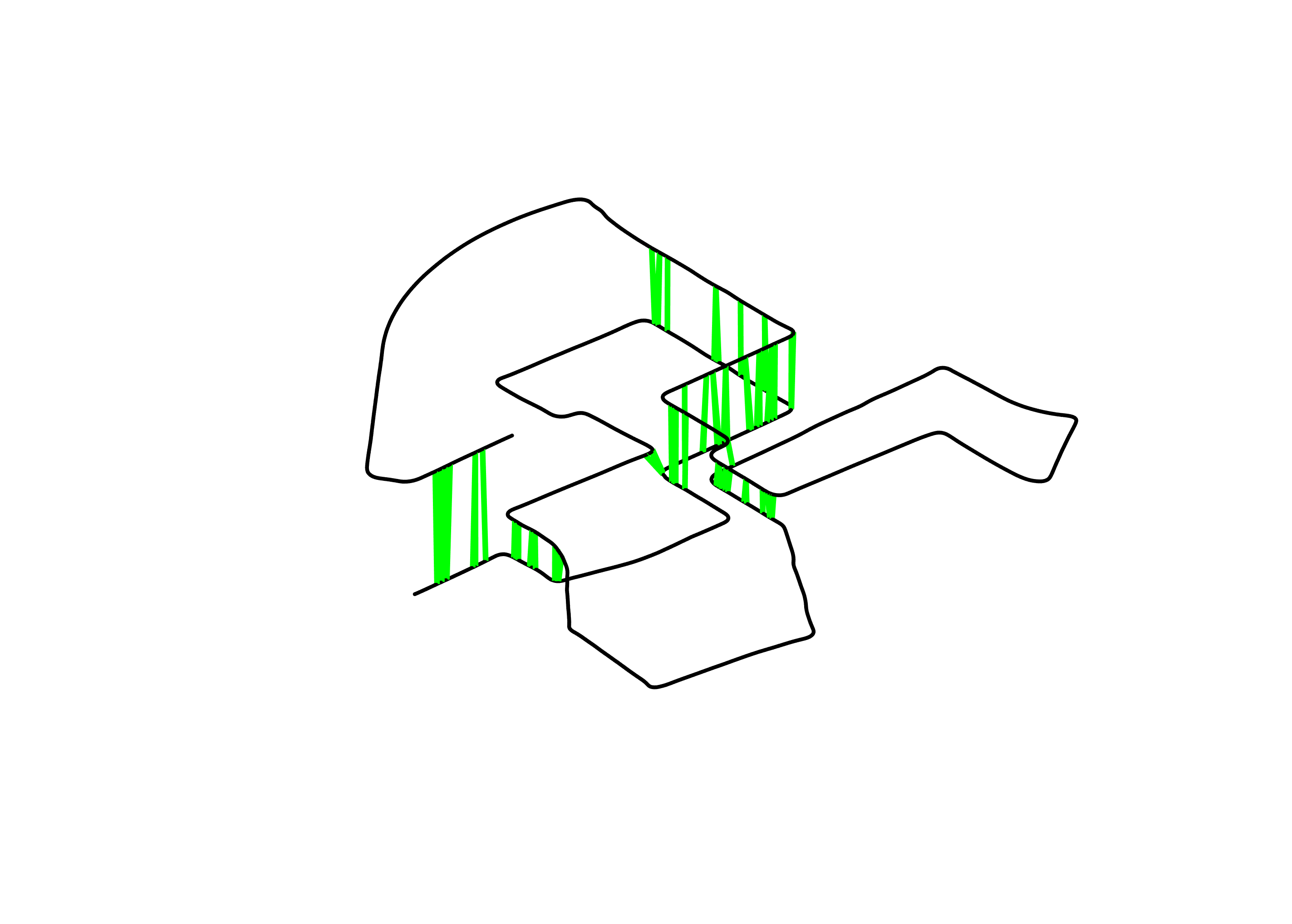}
      \label{fig:matchedkitti1}
    }\\
    \subfigure[PC]{%
      \includegraphics[width=\textwidth, trim = 200 130 130 120, clip]{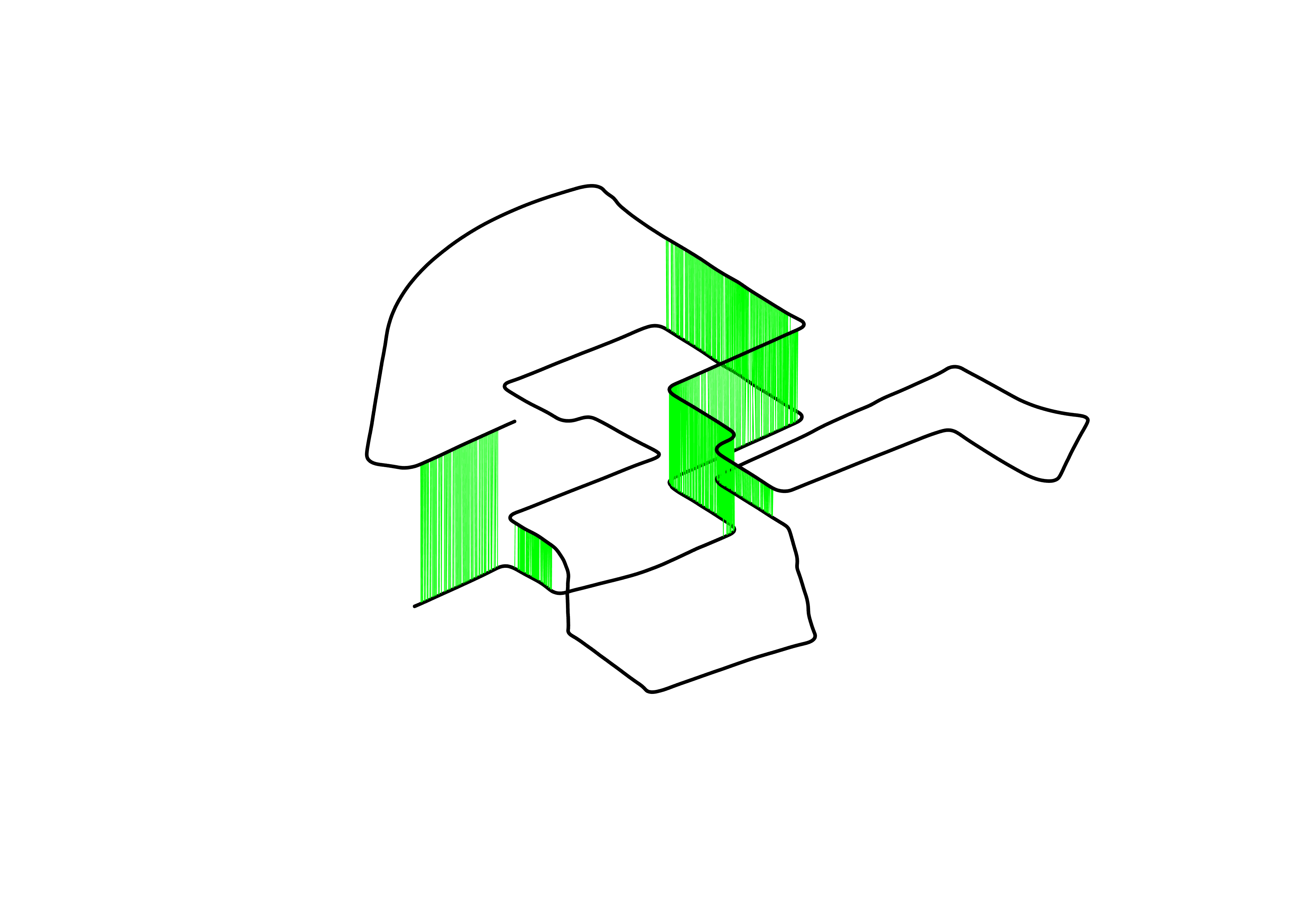}
      \label{fig:matchedkitti2}
    }%
  \end{minipage}
  \centering
  \begin{minipage}{0.5\columnwidth}
    \centering
    \subfigure[DR curve ($\downarrow$)]{%
      \includegraphics[width=\textwidth, trim = 0 0 0 0, clip]{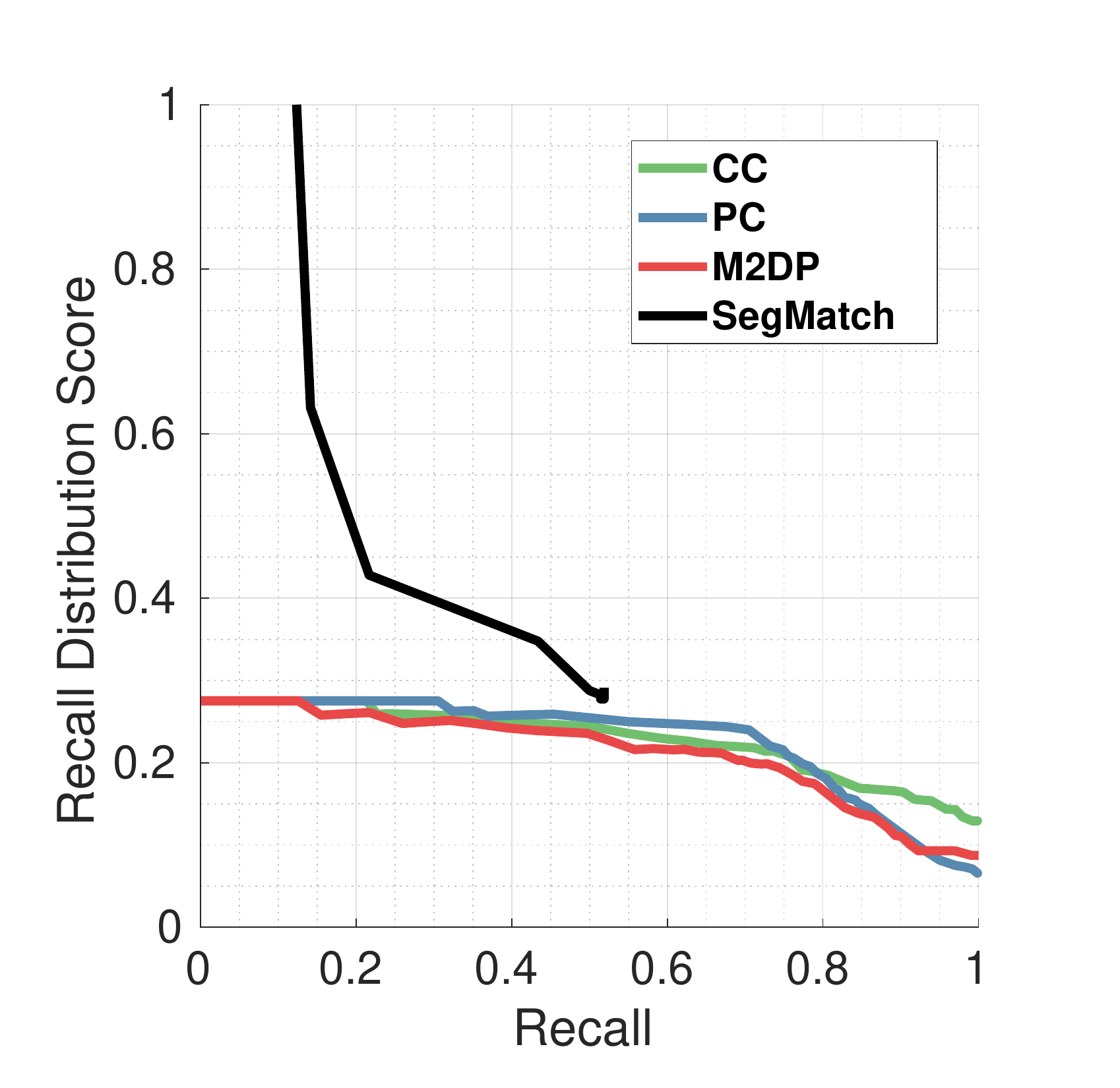}
      \label{fig:kitti00_dr}
    }
  \end{minipage}

  \caption{\subref{fig:matchedkitti1}-\subref{fig:matchedkitti2} Matched pairs (green) for \texttt{KITTI 00} at maximum precision in \figref{fig:kitti00_pr}. \subref{fig:kitti00_dr} The distribution-recall curve, named DR curve, in terms of \ac{KL} divergence with respect to the recall rate. Ideally, a flat curve with constant 0 \ac{KL} divergence for all recall rates would indicate perfectly distributed recalls. A lower score in the DR curve indicates better performance ($\downarrow$). In this DR curve, CC, PC, and M2DP all showed low KD divergence scores for all recall rates, which gradually decayed as recall increased. SegMatch detected sparse loop-closures, and the recall was lower than of the other methods. This appears as a larger KL divergence at low recall. However, the KD divergence decreased dramatically and reached a similar level of other methods. This indicates that SegMatch proposed more efficient loop-closures, achieving similar distribution score (i.e., KL divergence) with a smaller number of detections.}
  \label{fig:matchedriver}


\end{figure}




\begin{figure*}[!t]

  \centering
  \begin{minipage}{0.86\textwidth}\centering
  \subfigure[Trajectory and revisit distribution]{%
    \includegraphics[width=.57\textwidth, trim = 0 -50 0 0, clip]{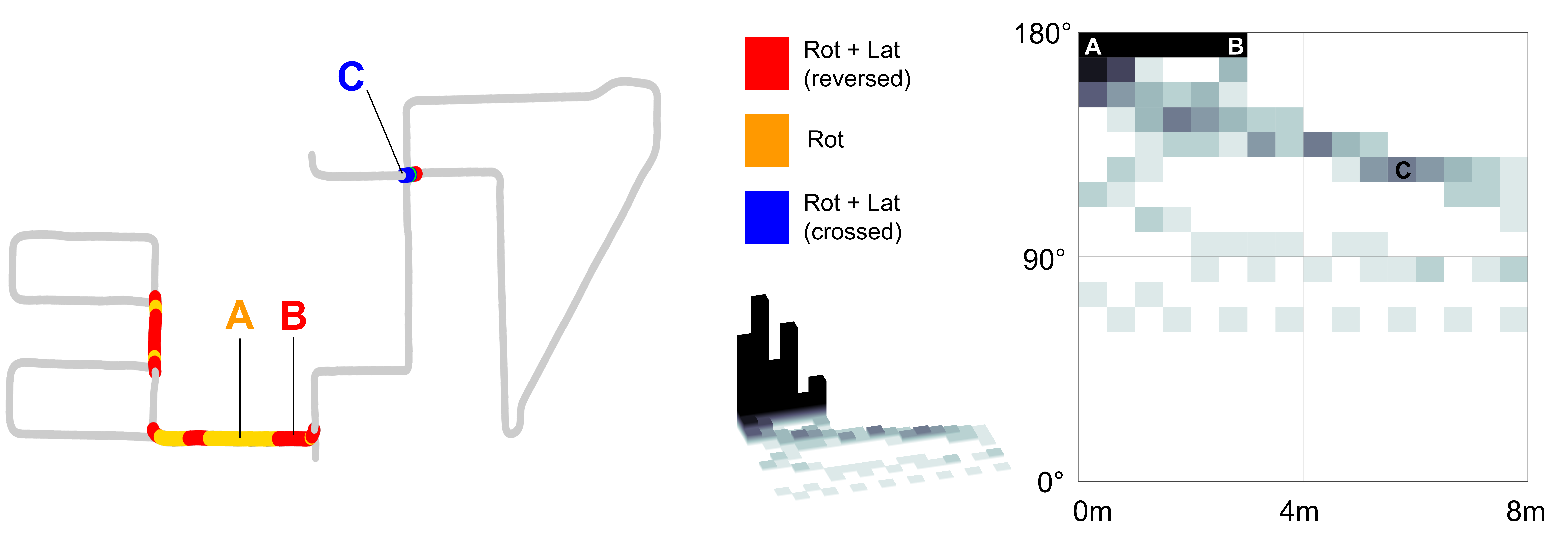}
    \label{fig:kitti08_traj}
  }%
  \subfigure[ PR curve ($\uparrow$)]{
    \includegraphics[width=0.2\textwidth, trim = 85 155 100 180, clip]{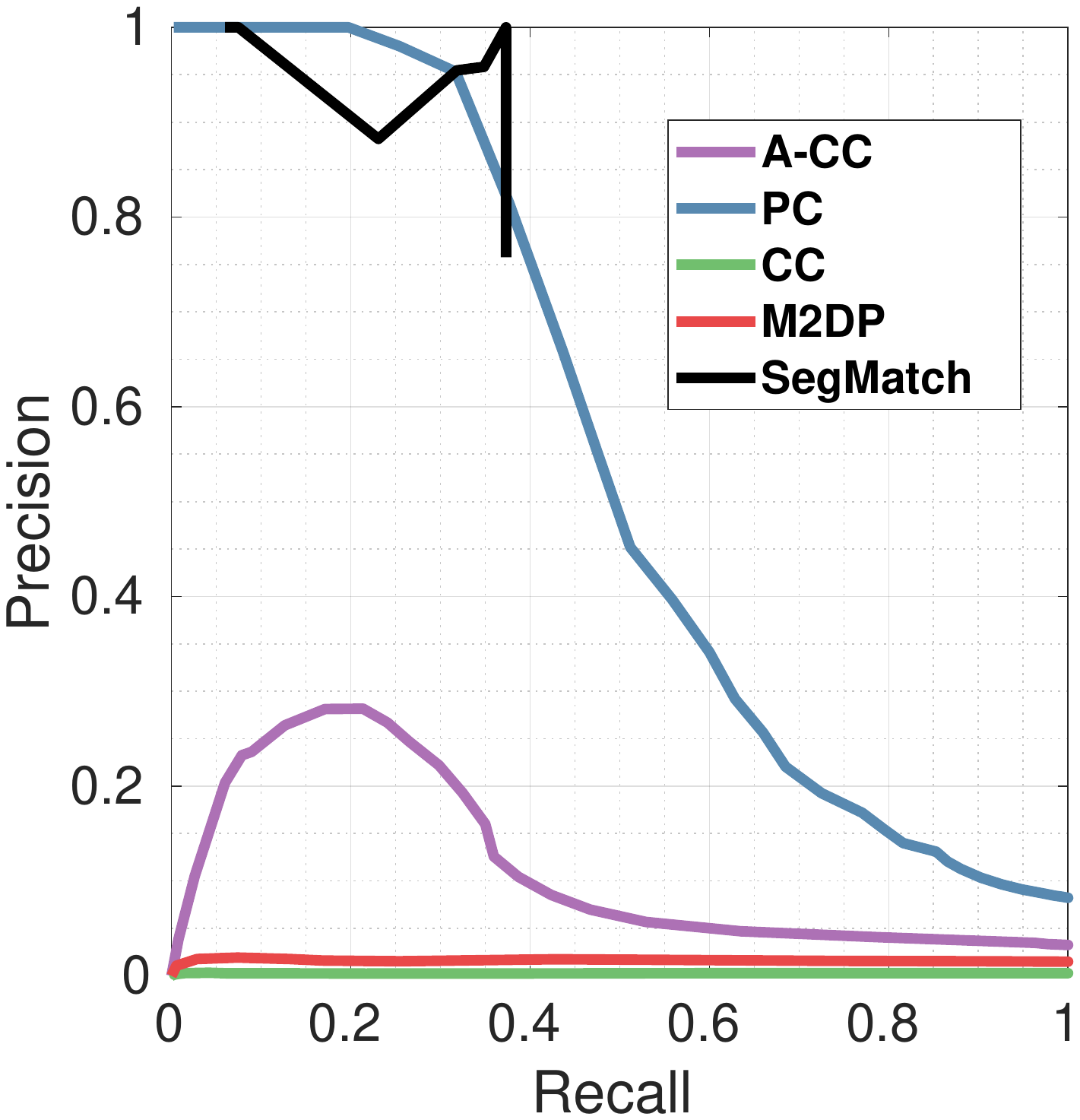}
    \label{fig:kitti08_pr}
  }%
  \subfigure[ DR curve ($\downarrow$)]{
    \includegraphics[width=0.2\textwidth, trim = 85 160 100 180, clip]{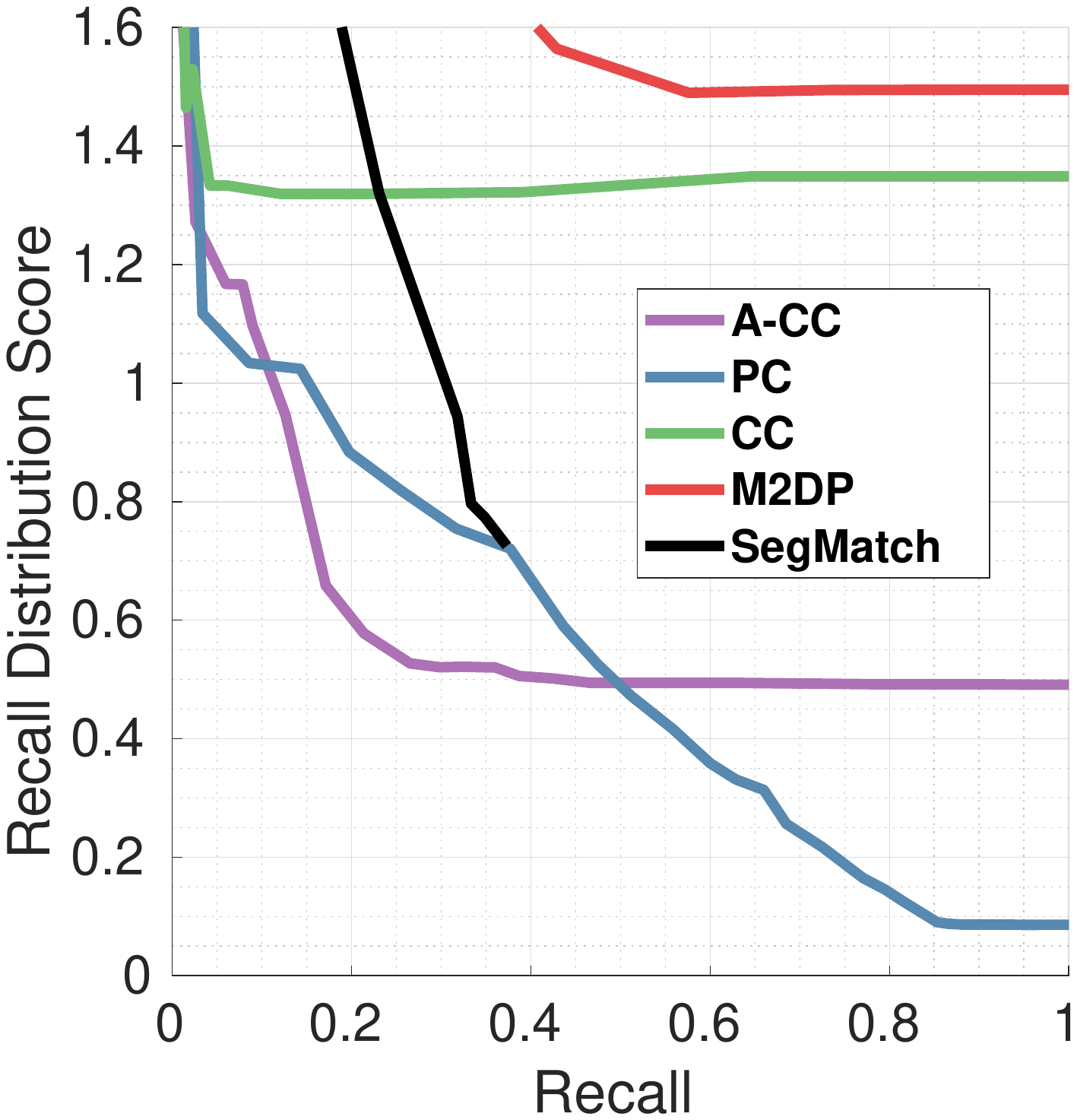}
    \label{fig:kitti08_dr} 
  }\\
  \subfigure[Trajectory and revisit distribution]{%
    \includegraphics[width=.57\textwidth, trim = 0 -50 0 0, clip]{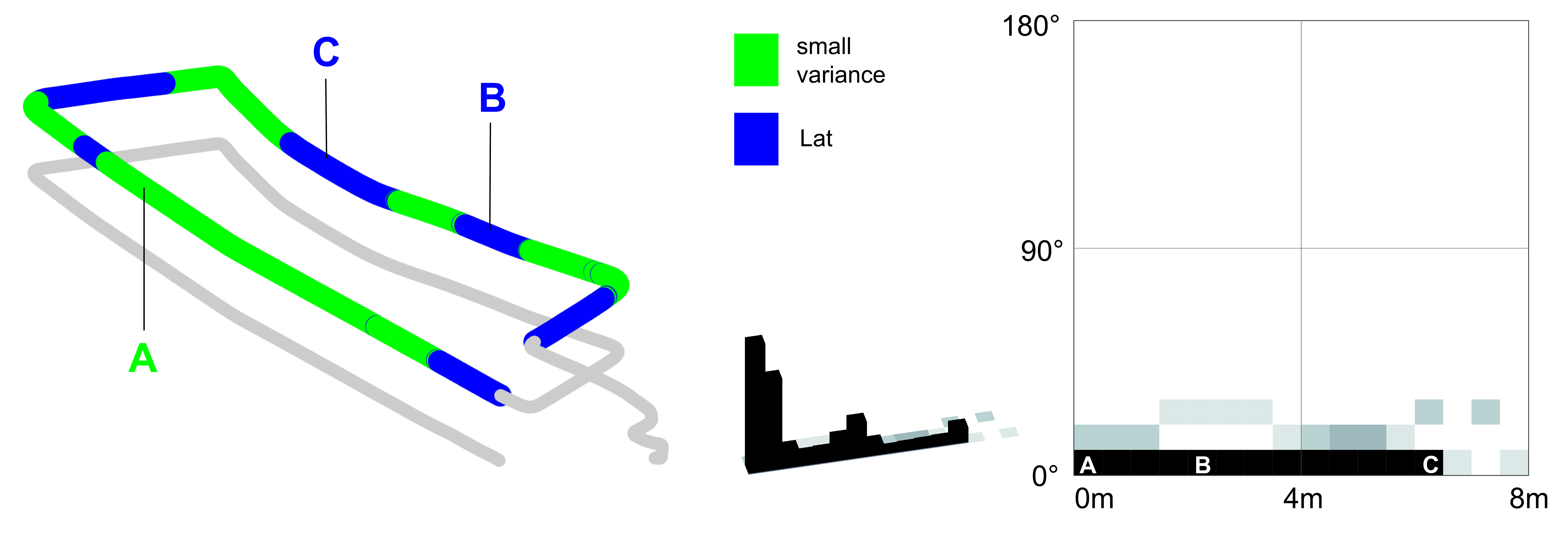}
    \label{fig:riverside02_traj}
  }%
  \subfigure[ PR curve ($\uparrow$)]{%
    \includegraphics[width=0.2\textwidth, trim = 85 155 100 180, clip]{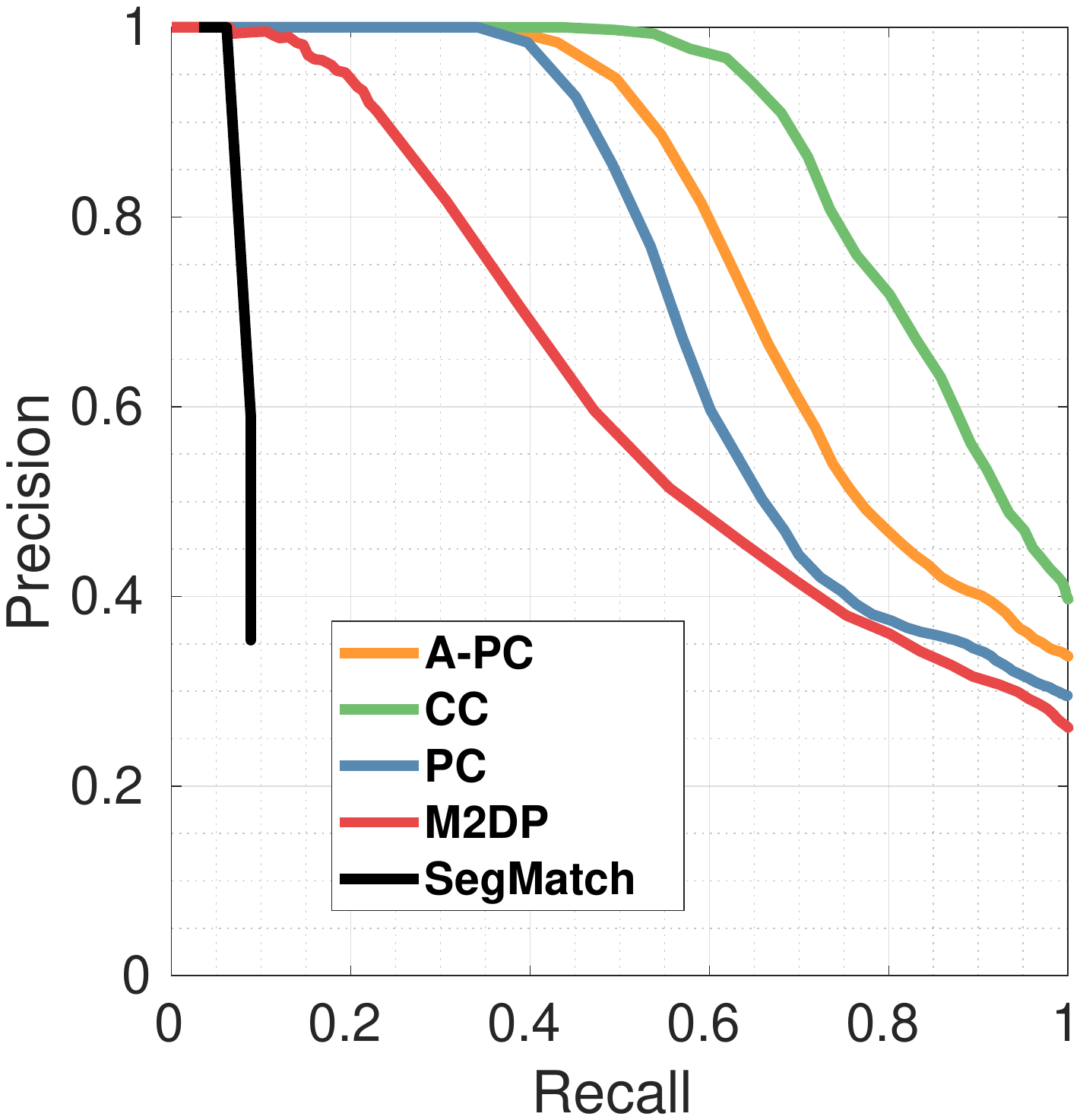}
    \label{fig:riverside02_pr}
  }%
  \subfigure[ DR curve ($\downarrow$)]{%
    \includegraphics[width=0.2\textwidth, trim = 85 160 100 180, clip]{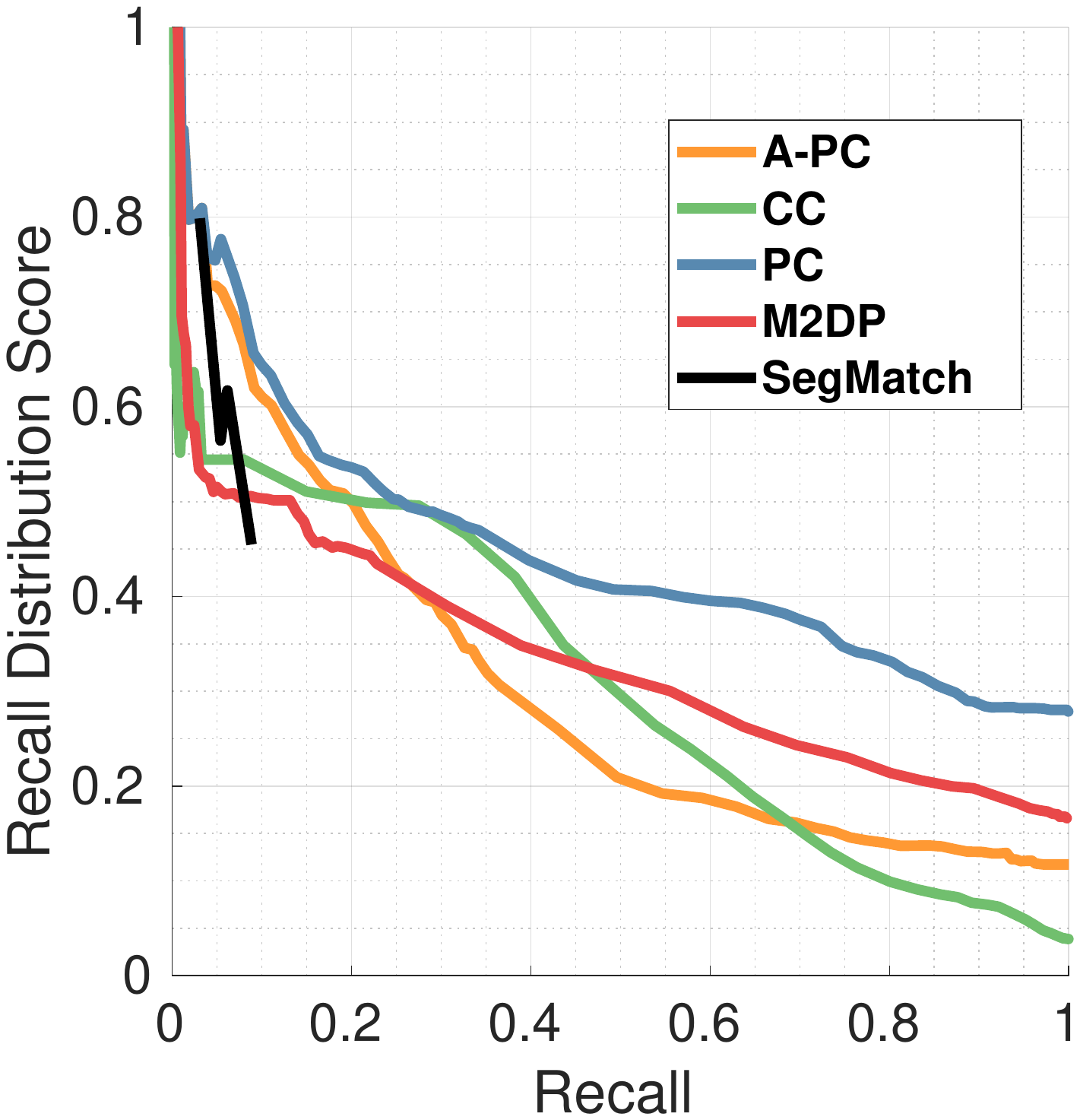}
    \label{fig:riverside02_dr}
  }%
  \end{minipage}

  \caption{
    Two sequences showing either predominantly rotational or lateral variance. \subref{fig:kitti08_traj} Most of the revisits occurred in the reversed direction for \texttt{KITTI 08}, with a concentrated distribution on the upper-left quadrant. \bl{\subref{fig:kitti08_pr}-\subref{fig:kitti08_dr}} \ac{PC} is the most robust method for \texttt{KITTI 08}. \subref{fig:riverside02_traj} \texttt{MulRan Riverside 02} contains lateral variations with little rotational change. \bl{\subref{fig:riverside02_pr}-\subref{fig:riverside02_dr} \ac{CC} is better capable of handling lateral variations.} Higher precision ($\uparrow$) and lower distribution ($\downarrow$) for all recalls indicate better performance.
  }
  \label{fig:exp_rot_or_lat}
\end{figure*}


\begin{figure*}[!t]
  \centering
  \includegraphics[width=0.98\textwidth]{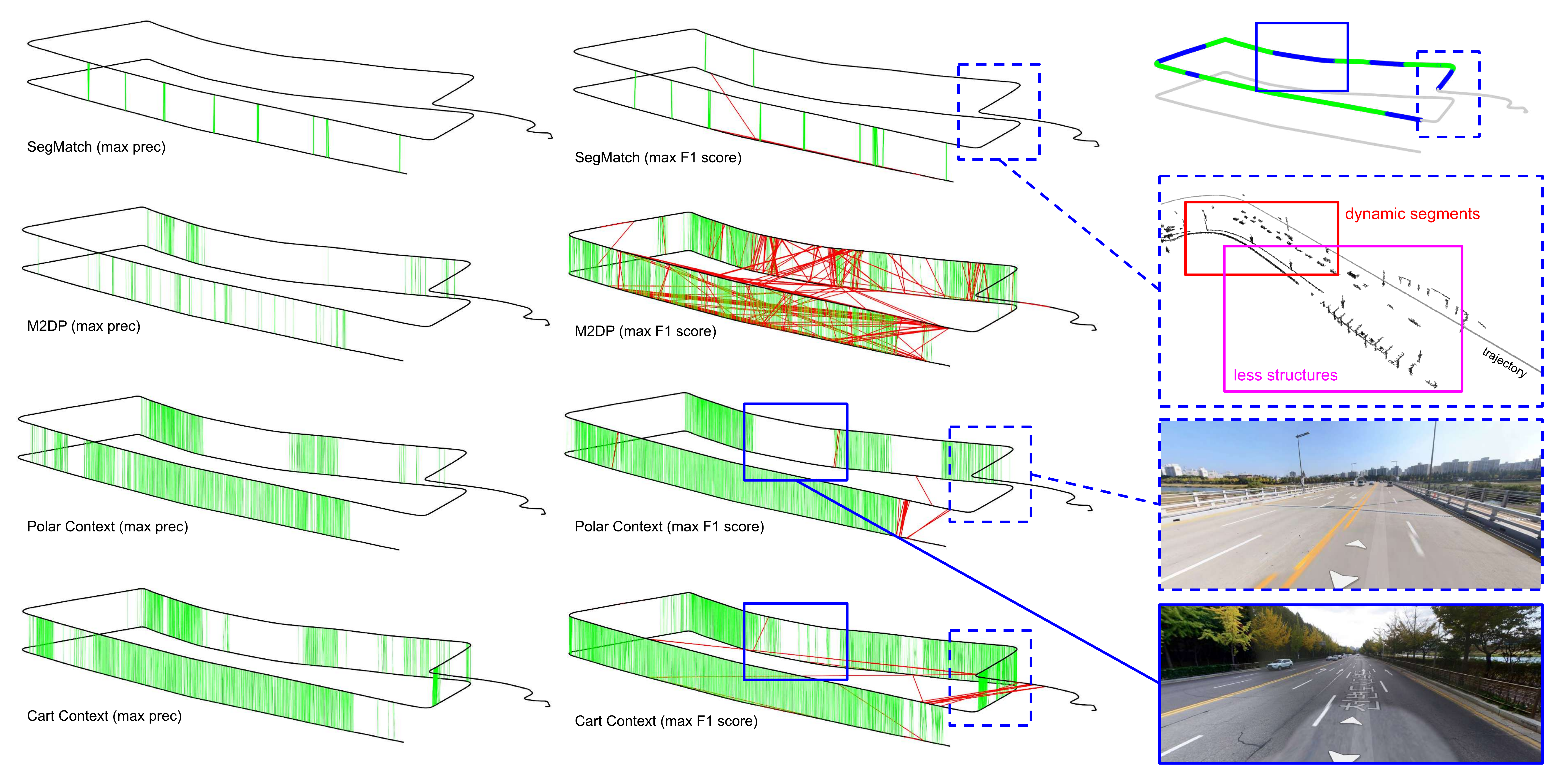}
  \caption{
    Time-elevation plots with true/false matches visualization for \texttt{Riverside 02}. We plotted correct (green) and incorrect (red) matches at the maximum precision (100\% for all methods) and the maximum F1 score. The solid blue box indicates the area with challenging multiple lane changes with repeated trees. The dotted blue box is the featureless bridge environment crossing a river. CC successfully found loops at those regions, while M2DP proposed many incorrect matches, and PC did not find loop-closures in these regions. 
  }
  \label{fig:matchedriver}
\end{figure*}


\begin{figure*}[!t]
  \centering\begin{minipage}{0.99\textwidth}\centering
  \subfigure[Trajectory and revisit distribution ]{%
    \includegraphics[width=.47\textwidth, trim = 0 -50 0 0, clip]{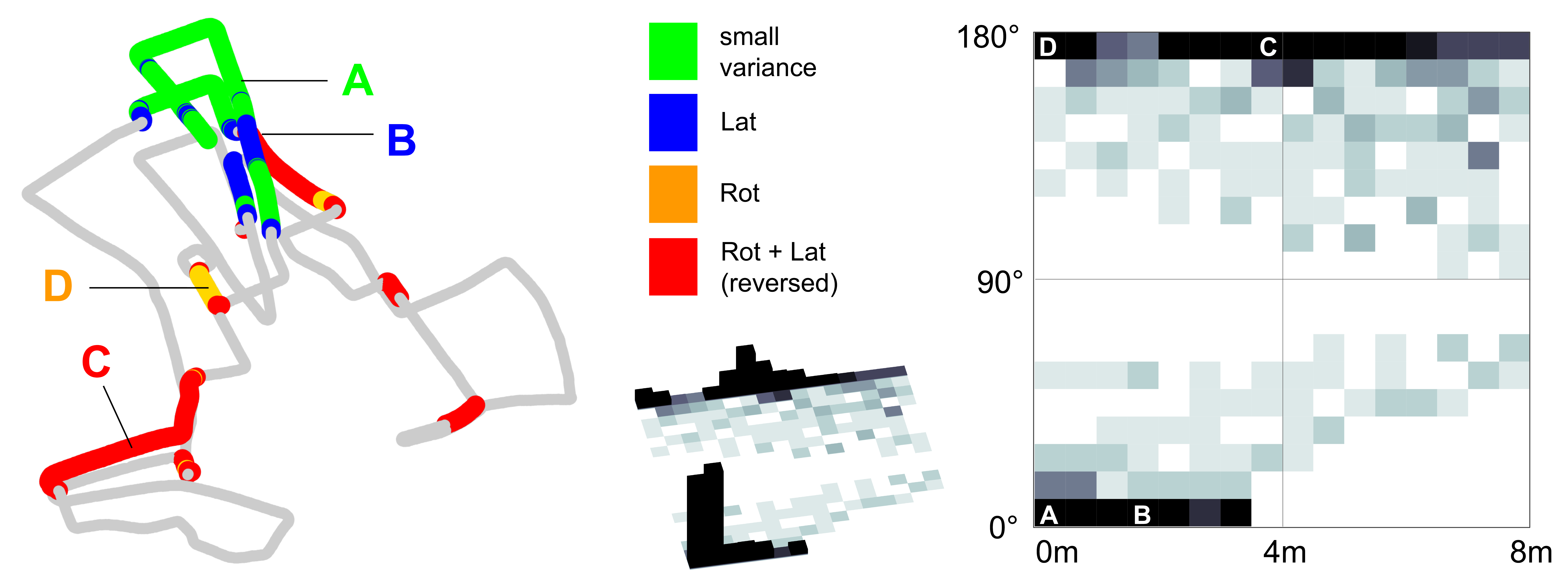}
    \label{fig:oxfordjan11_traj}
  }%
  \subfigure[ PR curve ($\uparrow$) ]{%
    \includegraphics[width=0.17\textwidth, trim = 85 155 100 180, clip]{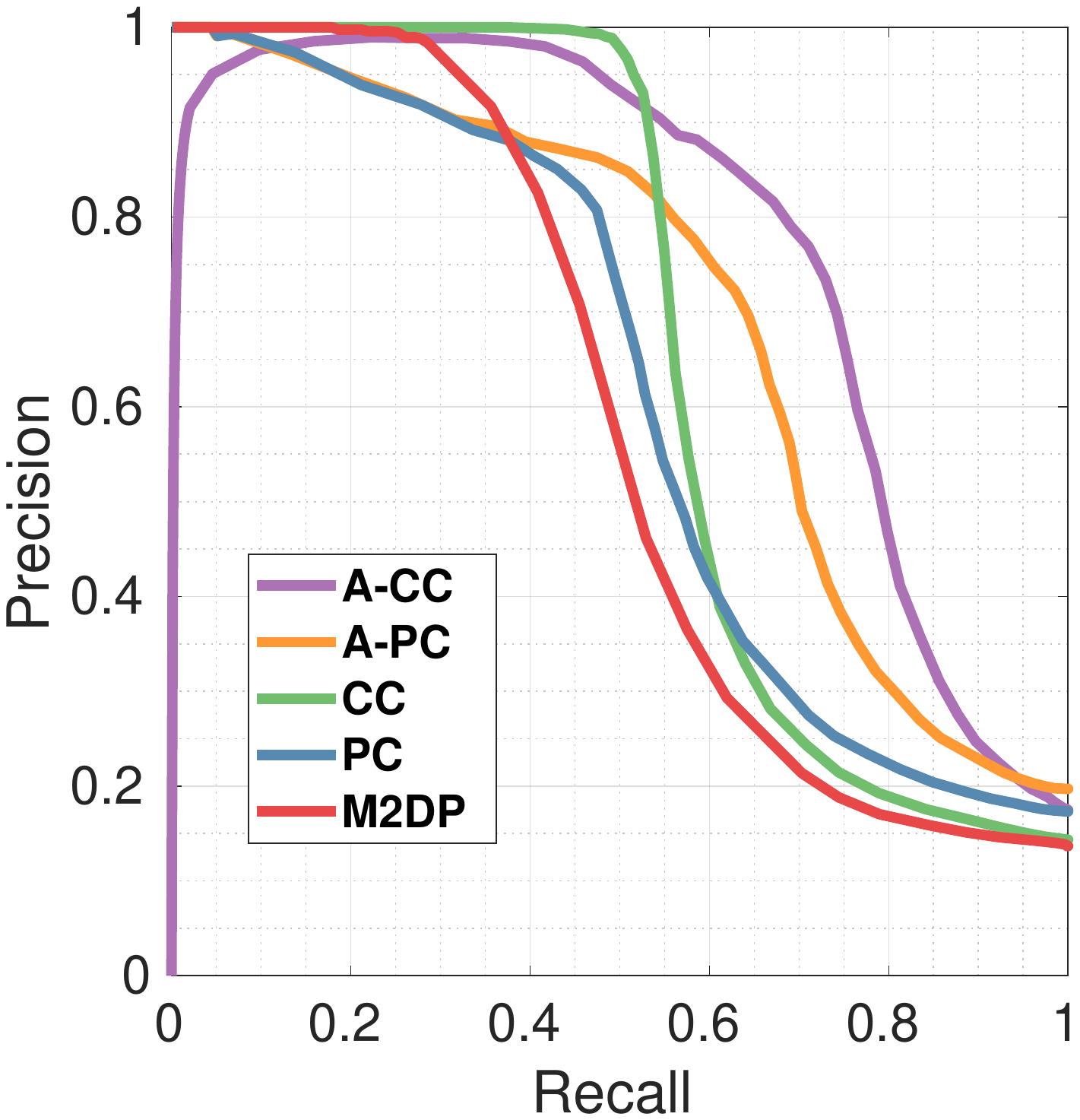}
    \label{fig:oxfordjan11_pr}
  }%
  \subfigure[ F1-R curve ($\uparrow$)]{%
    \includegraphics[width=0.17\textwidth, trim = 85 160 100 180, clip]{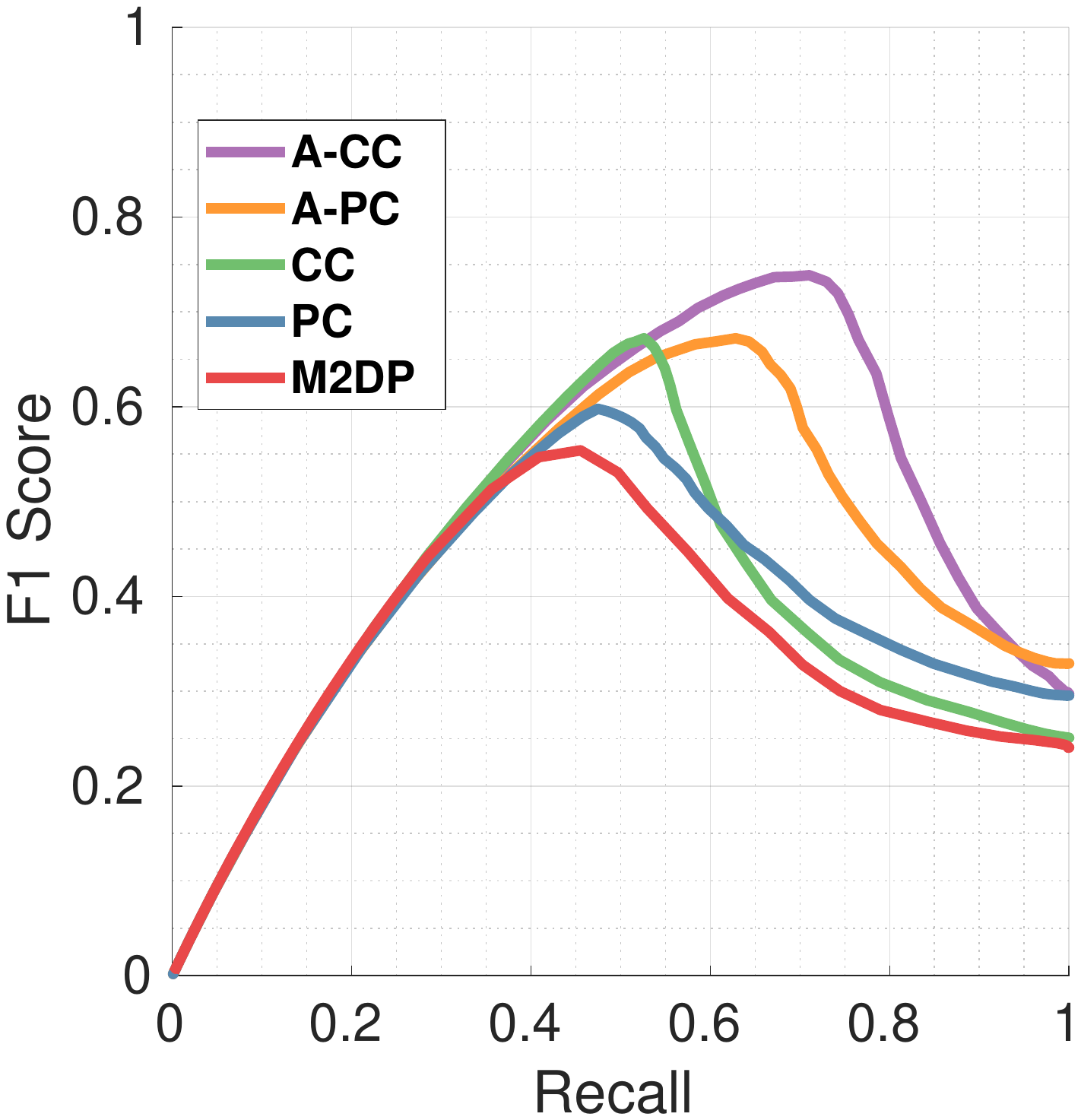}
    \label{fig:oxfordjan11_f1r}
  }%
  \subfigure[ DR curve ($\downarrow$)]{%
    \includegraphics[width=0.17\textwidth, trim = 85 160 100 180, clip]{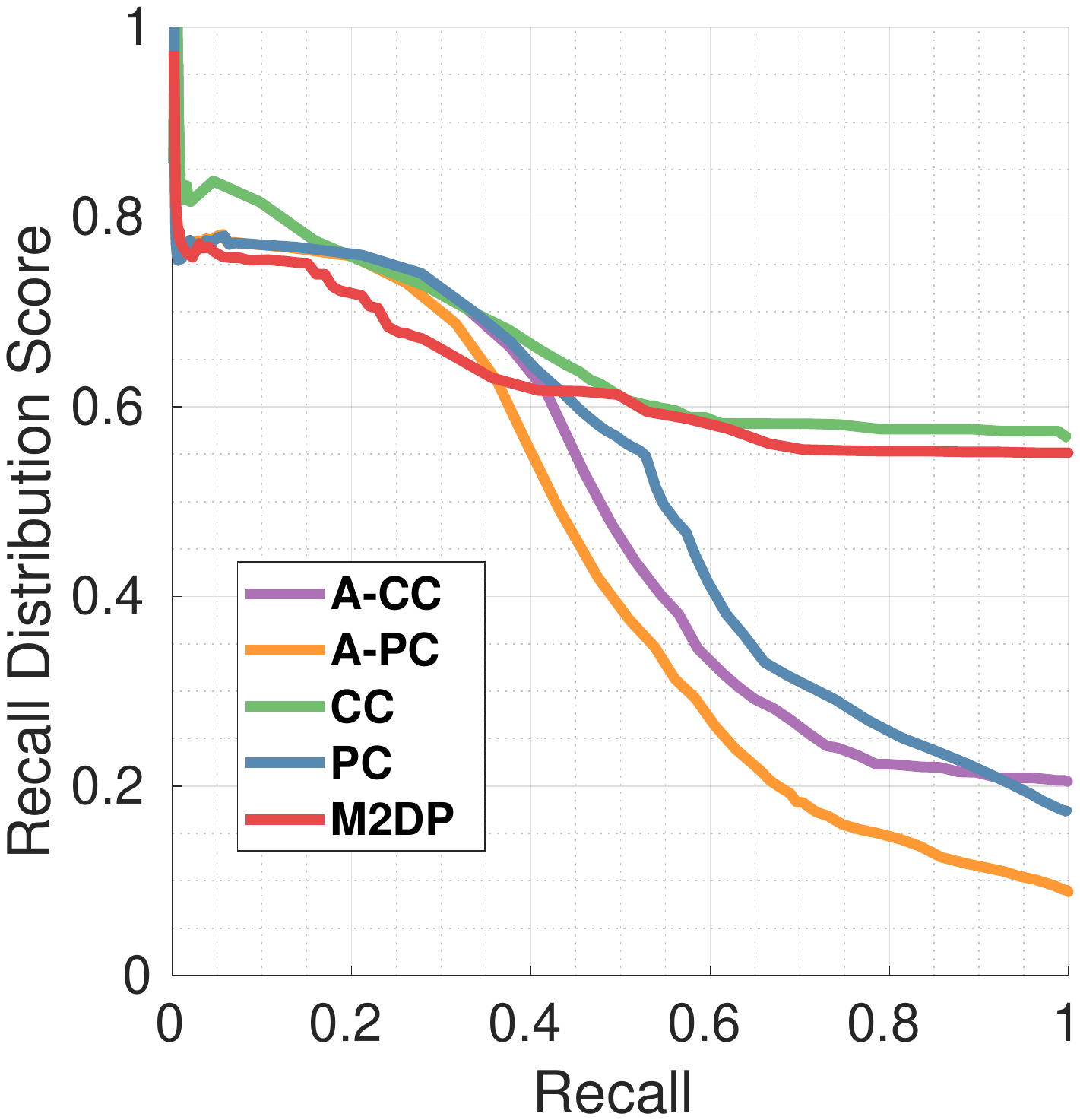}
    \label{fig:oxfordjan11_dr}
  }\\
  \end{minipage}
  \caption{\subref{fig:oxfordjan11_traj} \texttt{Oxford 2019-01-11-13-24-51} revealed concurrent rotational and lateral variance. The trajectory is color-coded by variance types. The true revisit distribution shows reversed revisits with lane changes. In addition to the PR and DR curves, we show the F1-R curve, which shows the change of F1 score with respect to the recall. \bl{\subref{fig:oxfordjan11_pr}-\subref{fig:oxfordjan11_dr} From the DR curve, we can easily see that augmentation not only increases the number of recalls with higher precision, but also increases the diversity of loop conditions.} 
  } 
  \label{fig:exp_oxford}
\end{figure*}

\begin{figure*}[!t]
  \centering\begin{minipage}{0.98\textwidth}\centering
  \def\mywidth{0.19\textwidth}%
  \subfigure[ M2DP ]{%
    \includegraphics[width=\mywidth, trim = 460 205 470 110, clip]{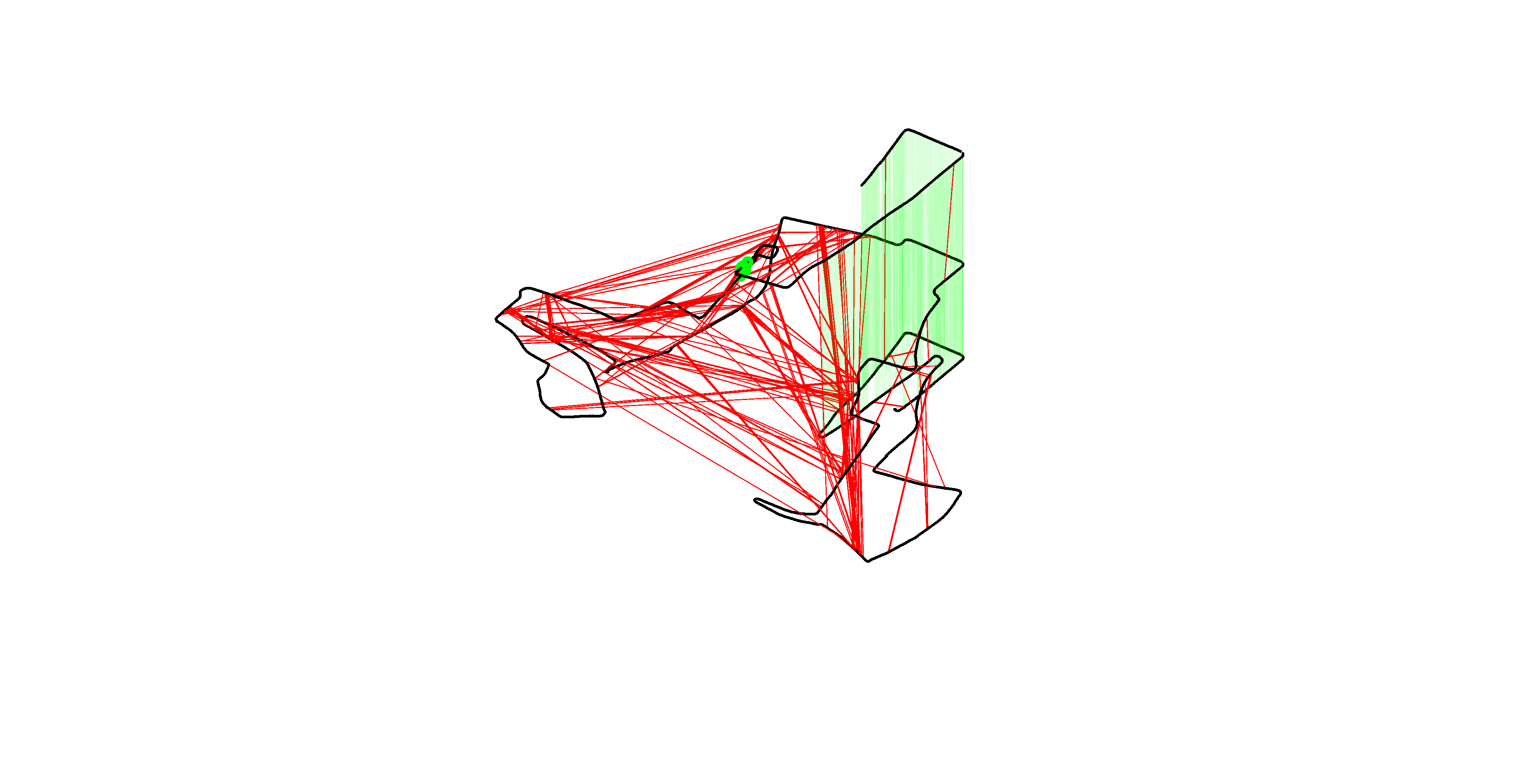}
    \label{fig:matchedoxford1}
  }%
  \subfigure[ PC ]{%
    \includegraphics[width=\mywidth, trim = 460 205 470 110, clip]{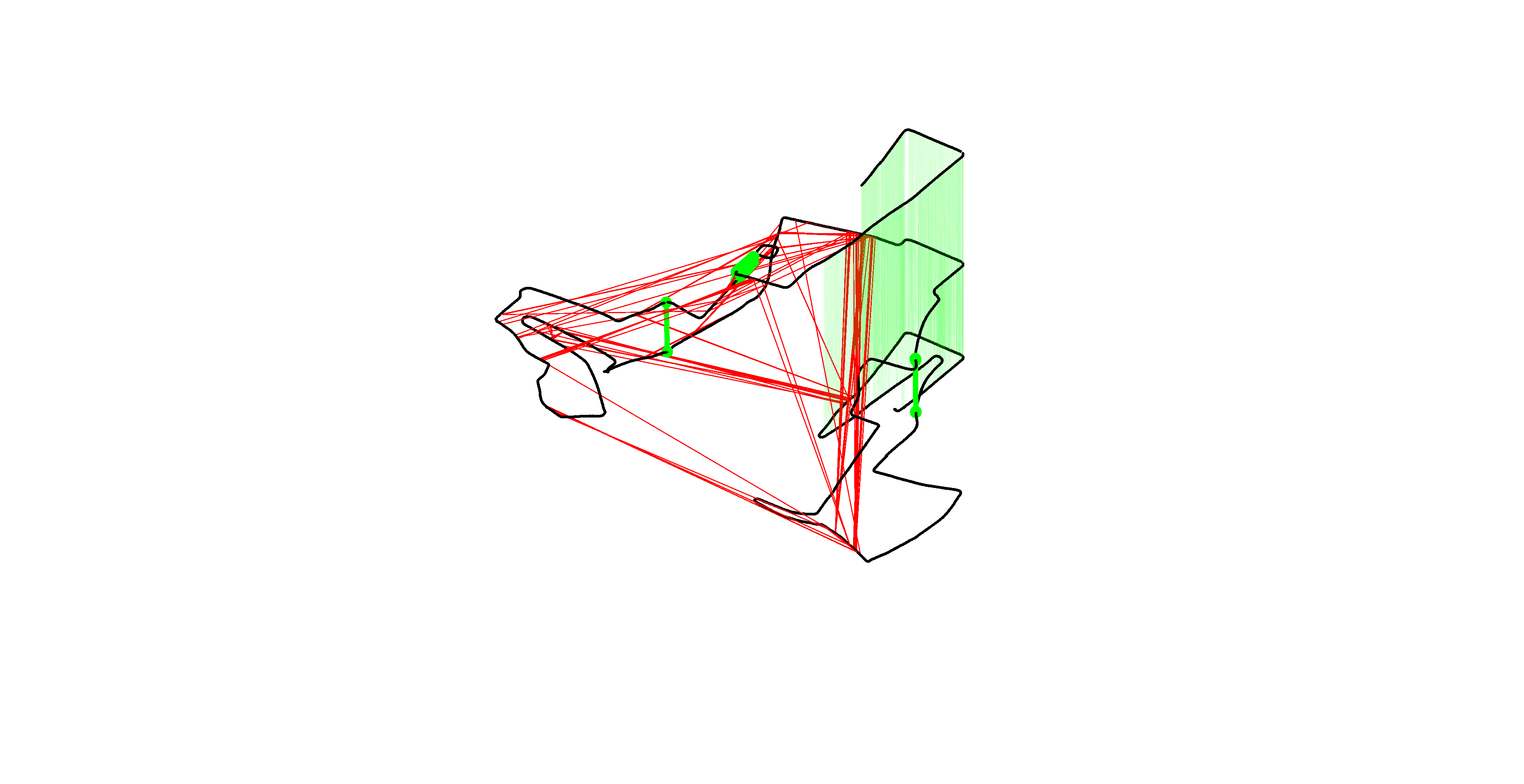}
    \label{fig:matchedoxford_pc}
  }
  \subfigure[ CC ]{%
    \includegraphics[width=\mywidth, trim = 460 205 470 110, clip]{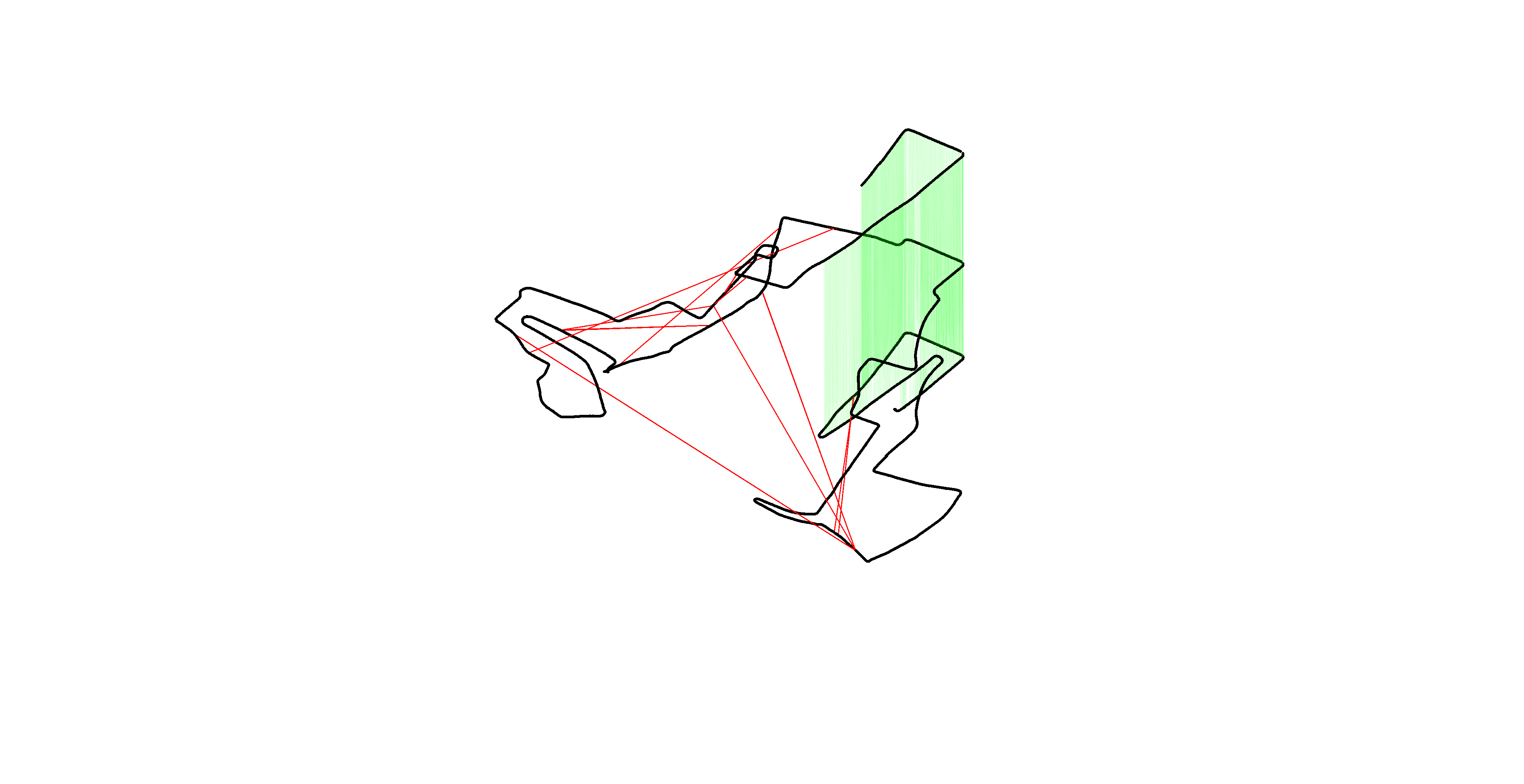}
    \label{fig:matchedoxford_cc}
  }%
  \subfigure[ A-PC ]{%
    \includegraphics[width=\mywidth, trim = 460 205 470 110, clip]{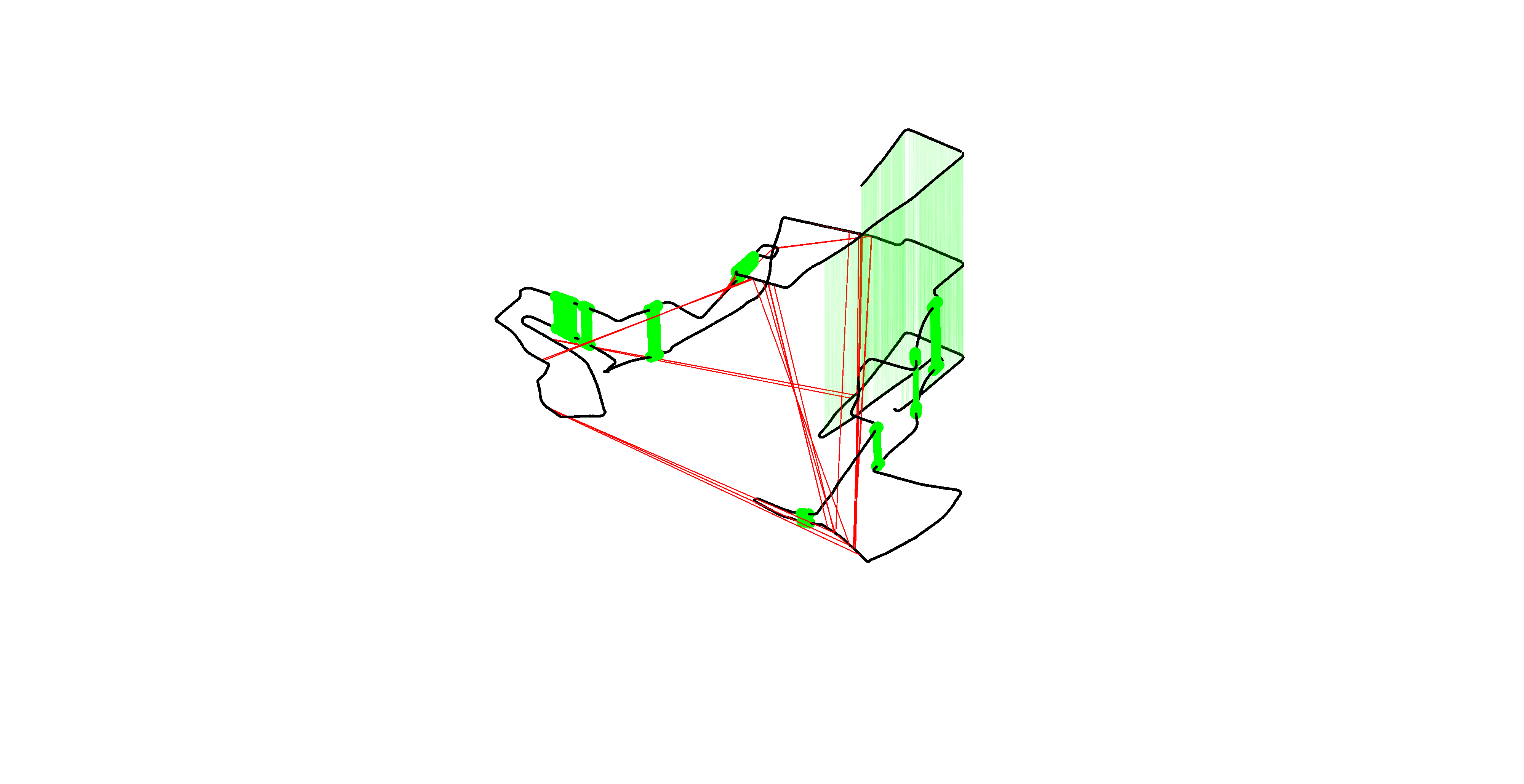}
    \label{fig:matchedoxford_apc}
  }%
  \subfigure[ A-CC ]{%
    \includegraphics[width=\mywidth, trim = 460 205 470 110, clip]{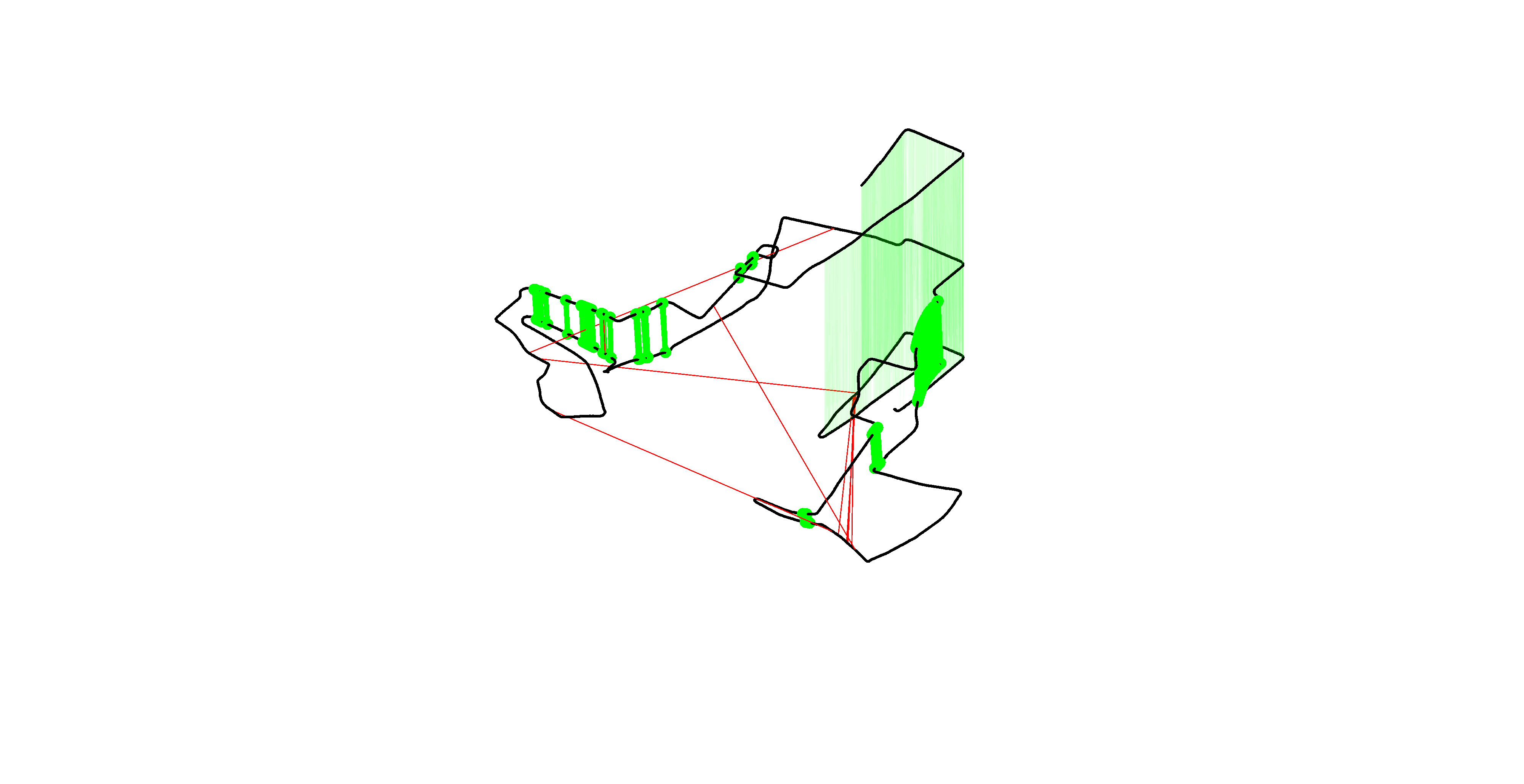}
    \label{fig:matchedoxford_acc}
  }%
  \end{minipage}

  \caption{Time-elevation match graph for \texttt{Oxford 2019-01-11-13-24-51}. Both true and false loop detections at recall of 50\% are visualized. The black line is the sequence trajectory whose height represents the time. The falsely connected matches are red, the true matches at \textit{easy} revisit are green, and the true matches at revisit with variance are drawn as green (bold).}
  \label{fig:matchedoxford}

\end{figure*}


\begin{figure*}[!t]
  \centering
  \begin{minipage}{0.27\textwidth}\centering
    \subfigure[Trajectory and revisit distribution ]{%
      \includegraphics[width=.95\textwidth, trim = 0 0 0 0, clip]{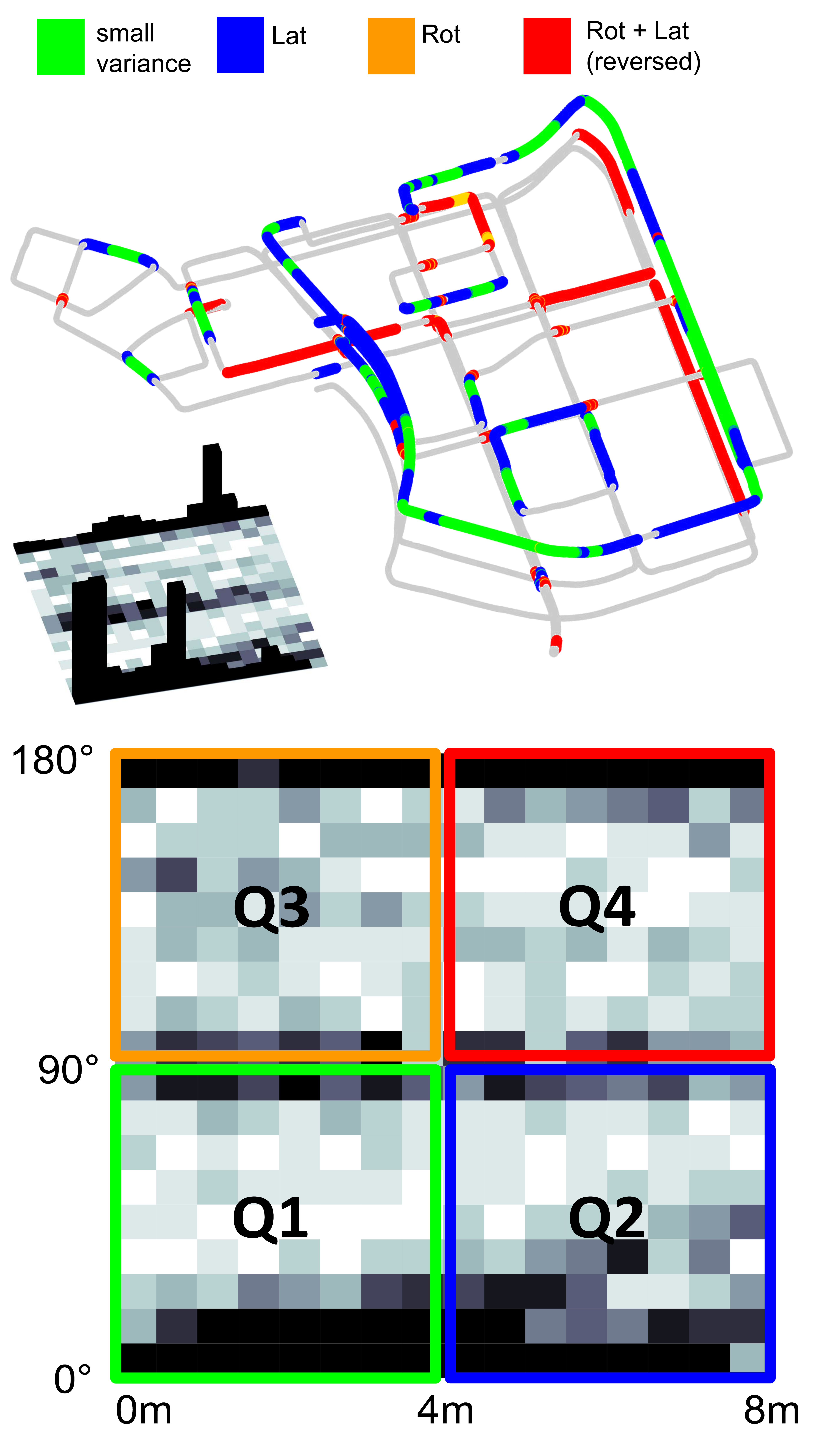}
      \label{fig:pangyo_traj}
    }%
  \end{minipage}
  \begin{minipage}{0.7\textwidth}\centering
    \subfigure[ PR curve ($\uparrow$)]{%
      \includegraphics[width=0.32\textwidth, trim = 85 155 100 180, clip]{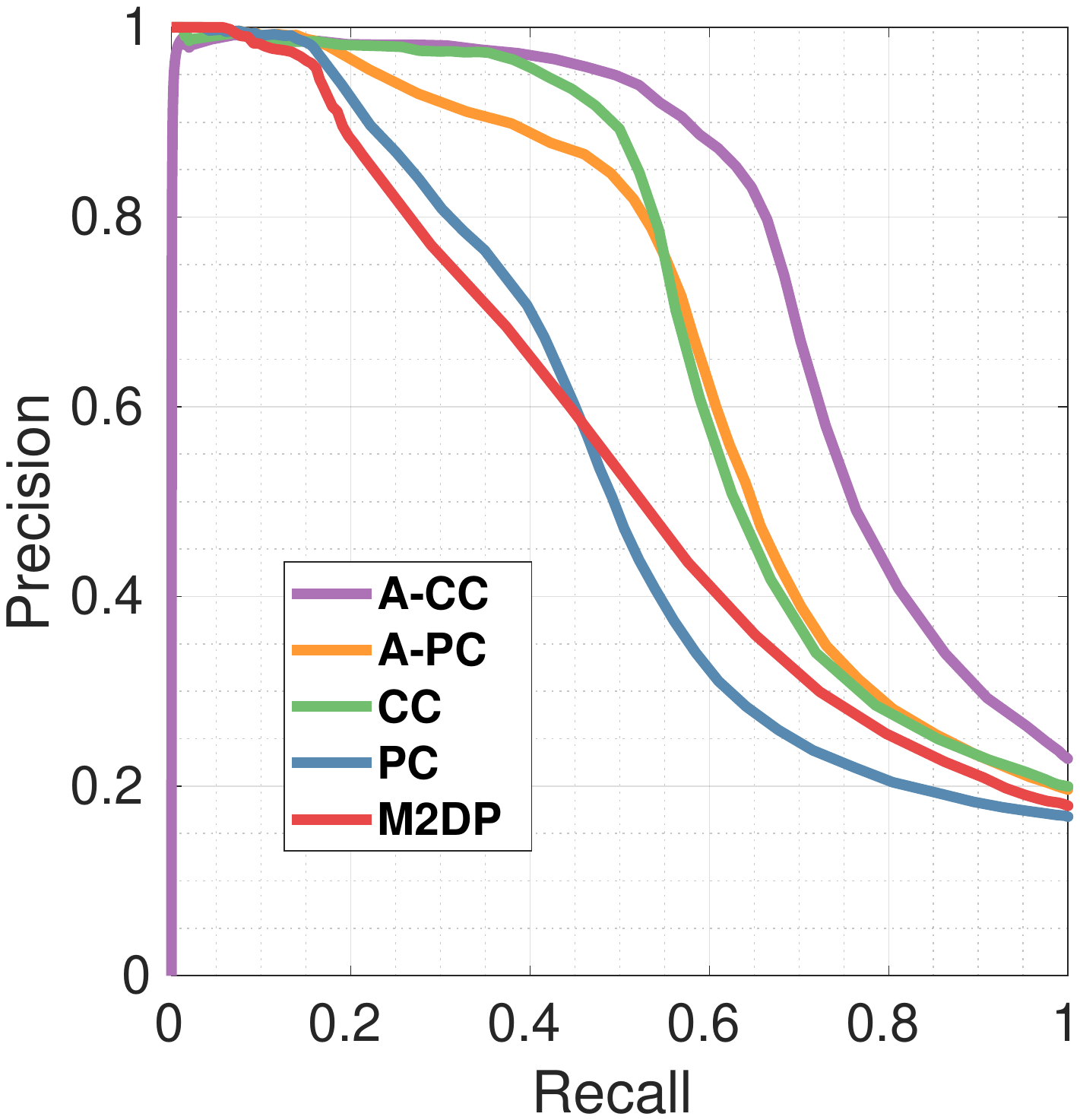}
      \label{fig:pangyo_pr}
    }%
    \subfigure[ F1-R curve ($\uparrow$)]{%
      \includegraphics[width=0.32\textwidth, trim = 85 160 100 180, clip]{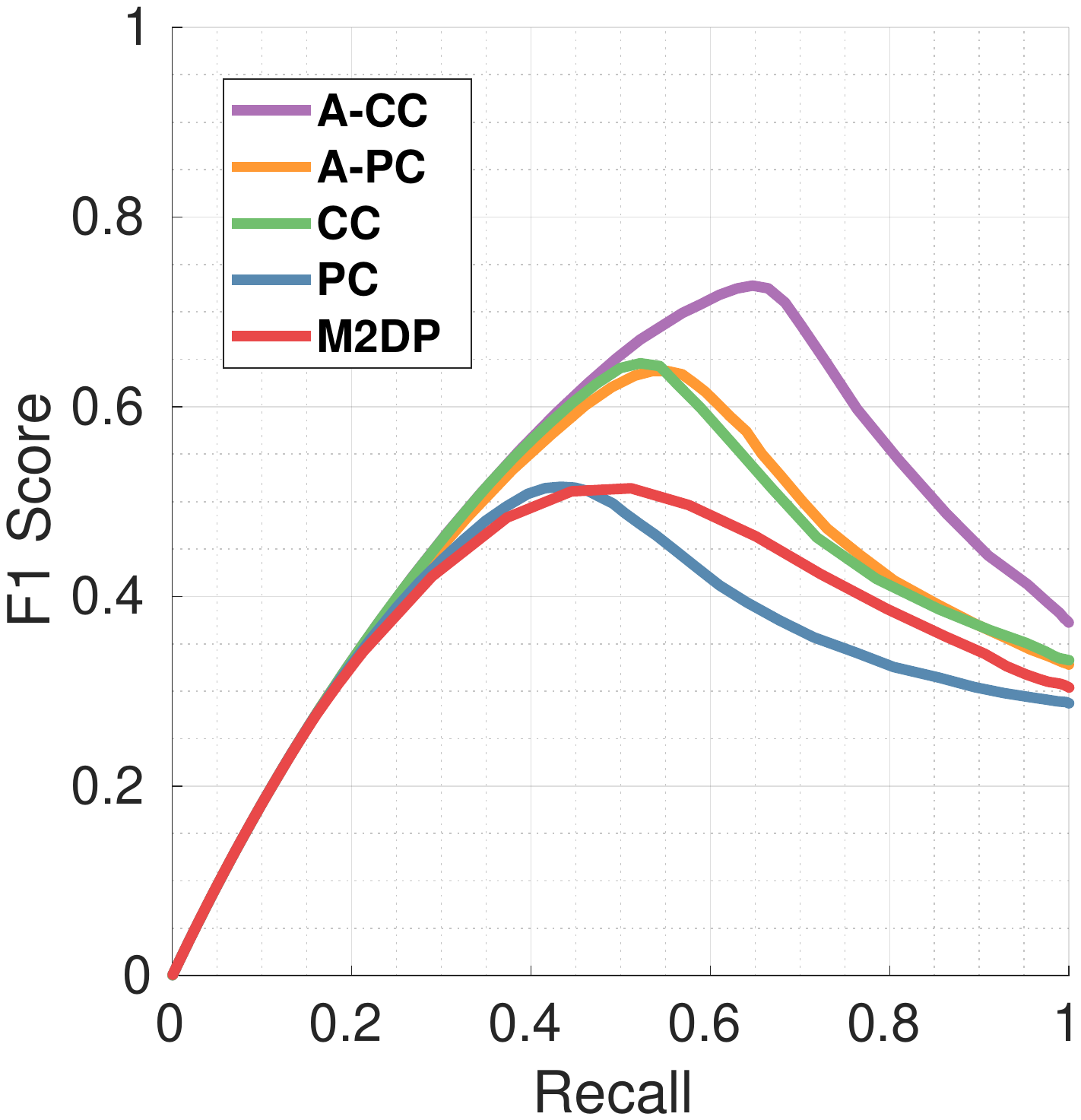}
      \label{fig:pangyo_f1r}
    }%
    \subfigure[ DR curve ($\downarrow$)]{%
      \includegraphics[width=0.32\textwidth, trim = 85 160 100 180, clip]{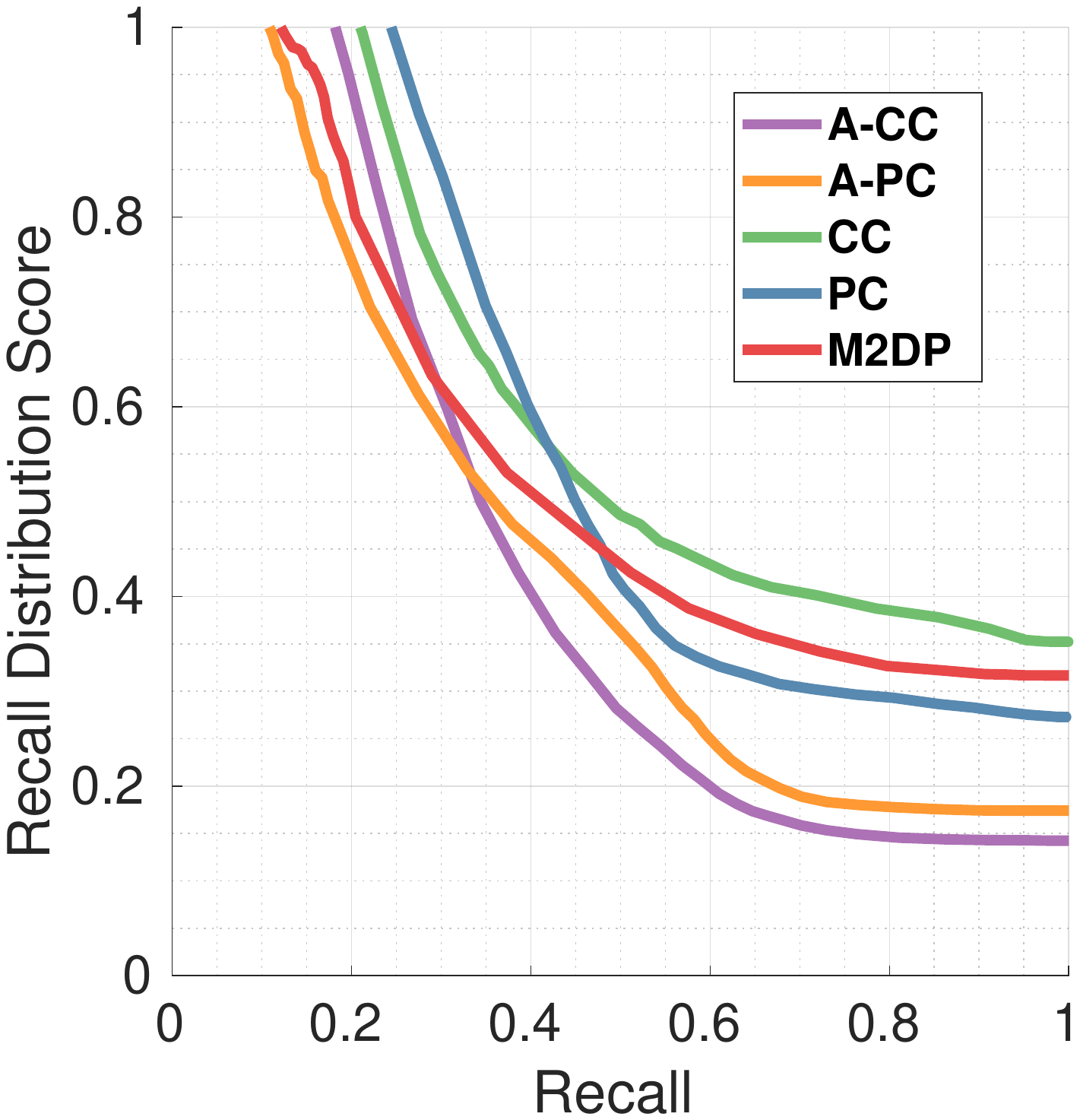}
      \label{fig:pangyo_dr}
    }\\
    \subfigure[ DR curve (\textit{Easy}) ]{%
      \includegraphics[width=.24\textwidth, trim = 60 155 95 155, clip]{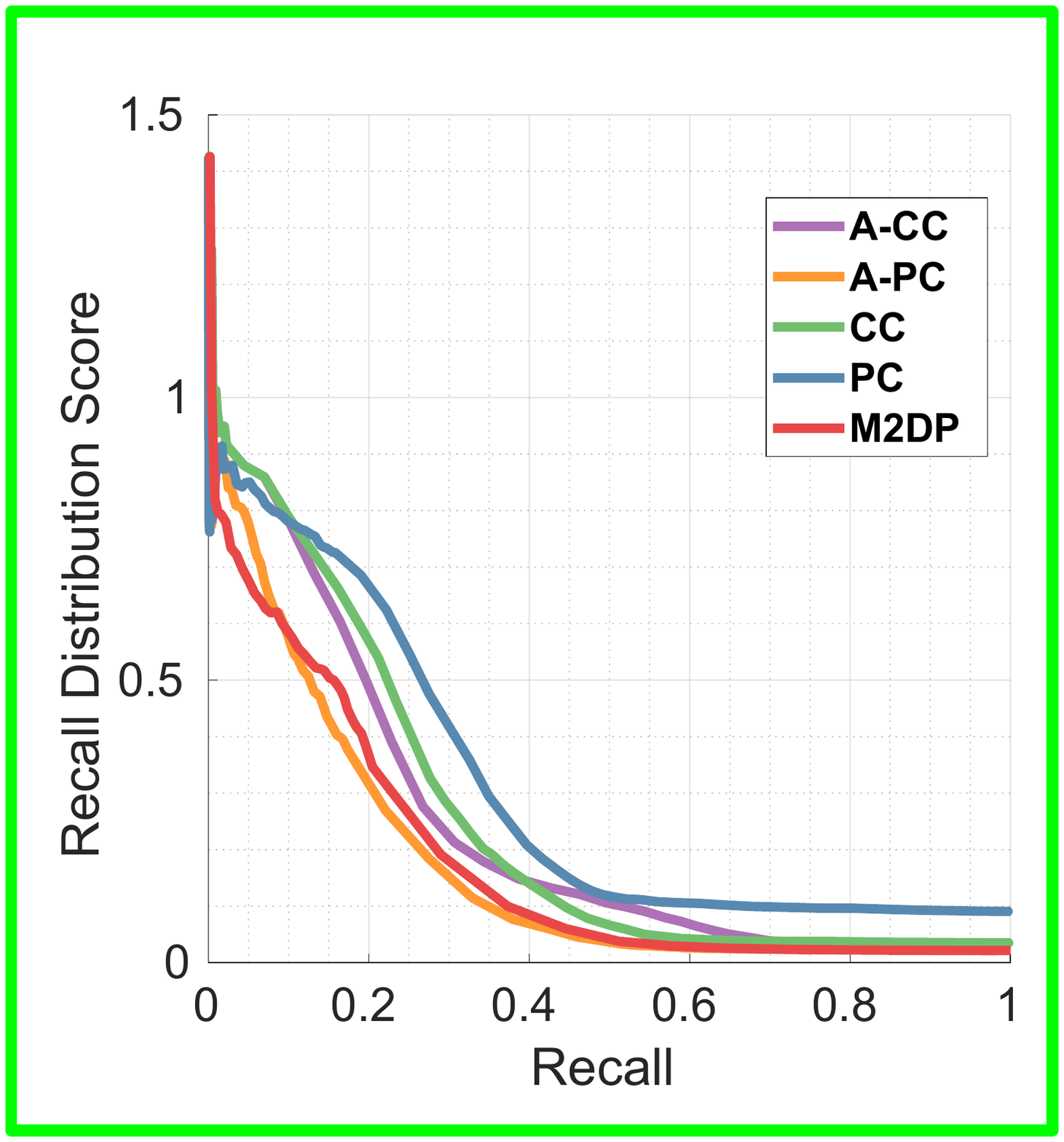}
      \label{fig:pangyo_dr1}
    }%
    \subfigure[ DR curve (\textit{Lat}) ]{%
      \includegraphics[width=0.24\textwidth, trim = 60 155 95 155, clip]{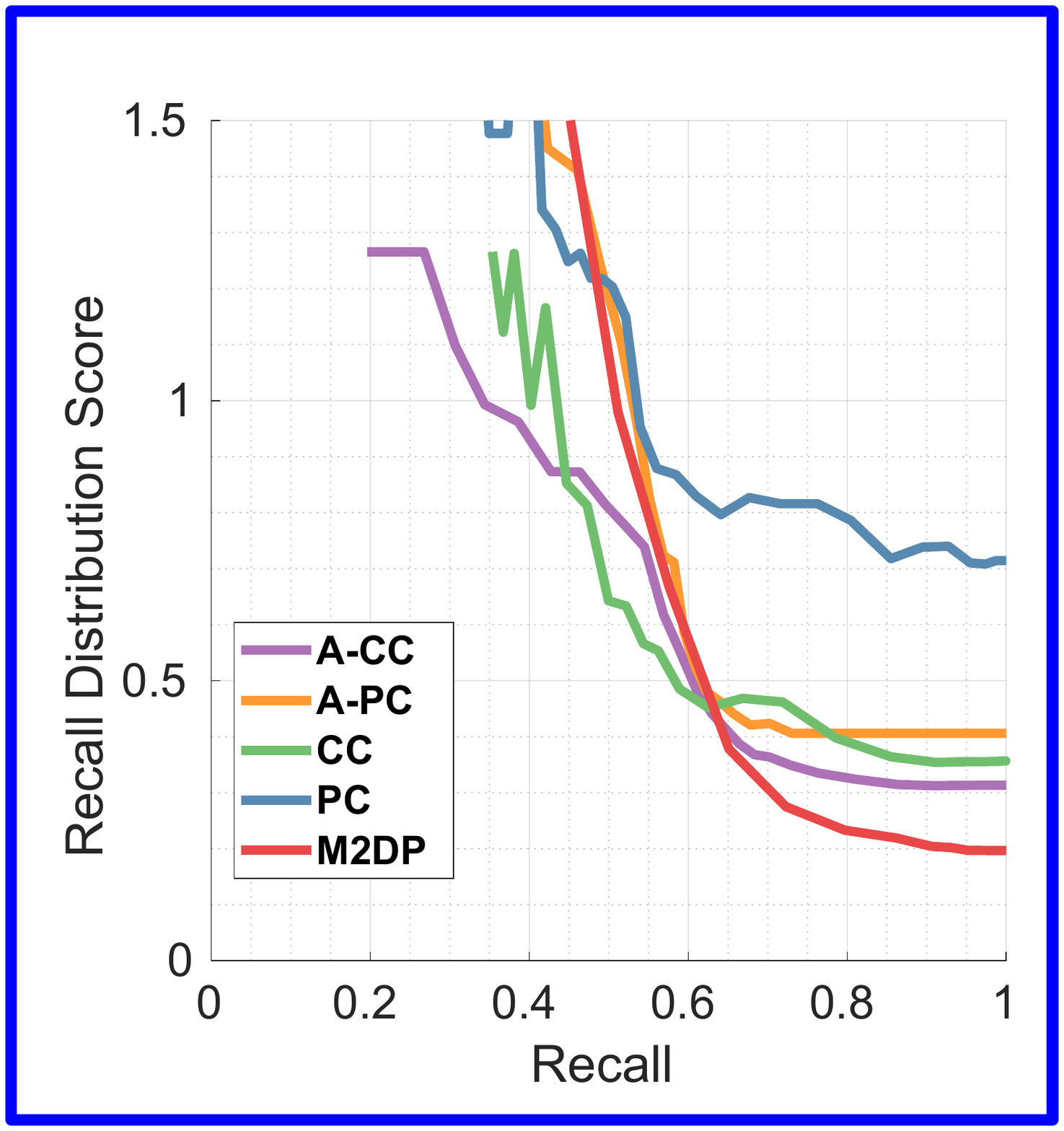}
      \label{fig:pangyo_dr2}
    }%
    \subfigure[ DR curve (\textit{Rot}) ]{%
      \includegraphics[width=0.24\textwidth, trim = 60 155 95 155, clip]{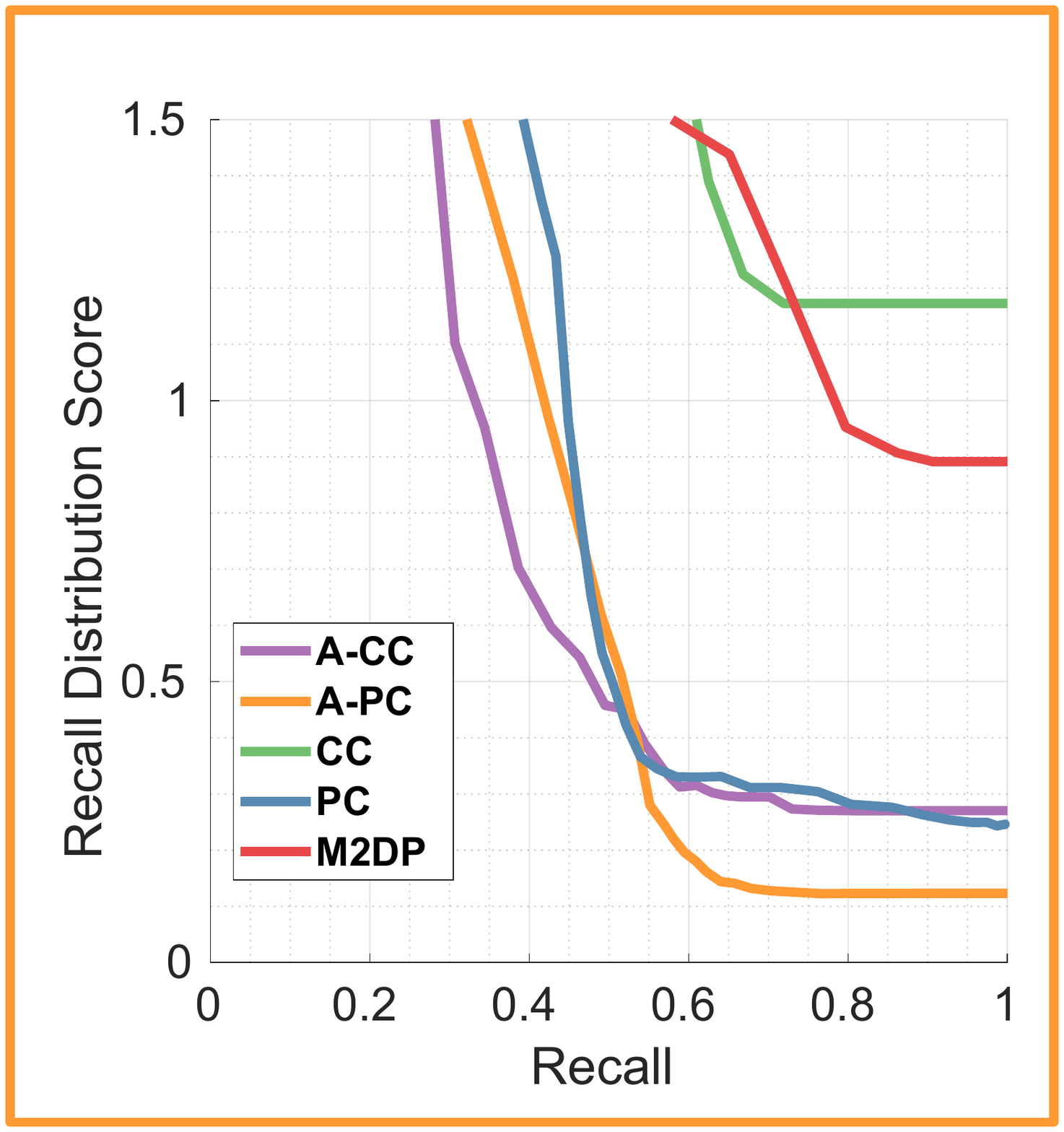}
      \label{fig:pangyo_dr3}
    }%
    \subfigure[ DR curve (\textit{Rot + Lat}) ]{%
      \includegraphics[width=0.24\textwidth, trim = 60 155 95 155, clip]{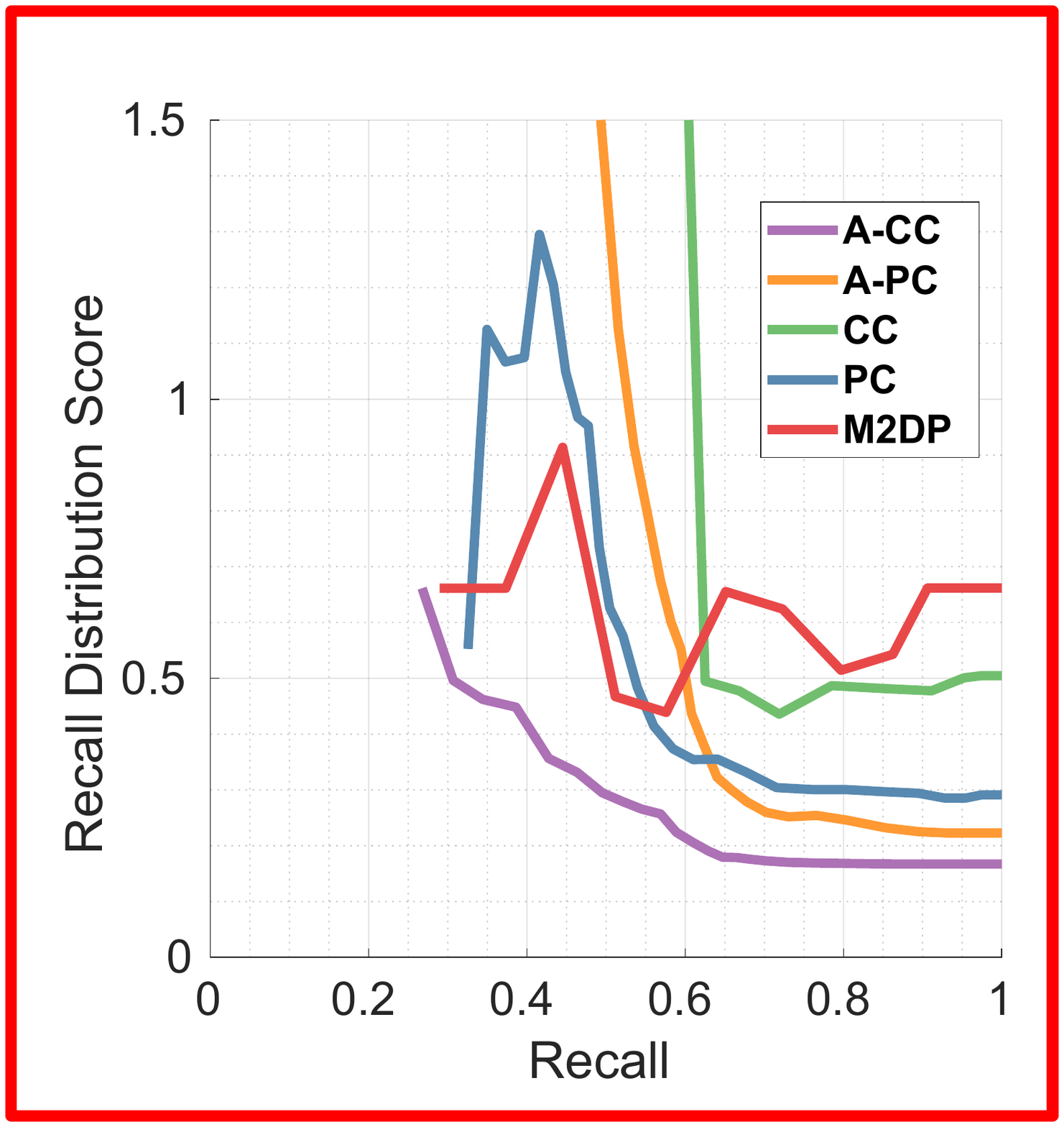}
      \label{fig:pangyo_dr4}
    }%
  \end{minipage}
  \caption{\subref{fig:pangyo_traj} Trajectory color-coded by variance types. Unlike other sequences, \texttt{Pangyo} includes all types of variance. Each quadrant indicates easy (Q1), lateral-dominant (Q2), rotational-dominant (Q3), and composite (Q4) cases. For the sequence, \subref{fig:pangyo_pr} PR curve, \subref{fig:pangyo_f1r} F1-R curve, and \subref{fig:pangyo_dr} DR curve are given. On the bottom row, we examine a detailed view for each quadrant from the top-down view of the revisit distribution. Each quadrant represents the predominant group of revisits in one type of variance.}
  \label{fig:exp_pangyo}

\end{figure*}

\section{Experimental Evaluation}
\label{sec:exp}

Next, we validated our spatial descriptor and place recognition algorithm on various datasets. As addressed in \figref{fig:exampletax} and \tabref{tab:taxonomy}, coping with multiple variations of a place is crucial for loop detection and global localization. To clearly state the associated invariance, we color-coded routes depending on the revisit types.

\subsection{Revisit with Small Variance}
\label{sec:easy}

Among the eight sequences in \tabref{tab:dataset}, \texttt{KITTI 00} and \texttt{MulRan KAIST 03} are relative \textit{easy} sequences, containing small rotational/translation variance and few dynamic objects. For \texttt{KITTI 00} (\figref{fig:kitti00_pr}), M2DP showed the highest performance with respect to both precision and recall. SegMatch revealed quite lower recall compared to the others; however, the distribution of the recognition was sufficient to construct a globally consistent map. In particular, SegMatch successfully recognized the loop at the middle crossroad, where a composite change (both \textit{rotational and lateral}) existed (see \figref{fig:matchedkitti1}), while the other methods failed to do so. Both \ac{PC} and \ac{CC} showed similar performance because \texttt{KITTI 00} barely has any rotations or lane changes at the loops. The \ac{PC} matched pairs at the 100\% precision are visualized in \figref{fig:matchedkitti2}. In \texttt{MulRan KAIST 03} (\figref{fig:kaist03_pr}), all of the methods successfully recognized the loops because this sequence is for a campus environment with almost no lane changes and few dynamic objects.

\subsection{Revisit with Rotational or Lateral Variance}
\label{sec:rot_or_lat}

Next, we examined sequences showing dominant variance in either the rotational or lateral direction. Their performance is summarized in \figref{fig:exp_rot_or_lat}.

\subsubsection{\texttt{KITTI 08}} This sequence only contains reverse revisits, with half of them further including simultaneous lane change. This appears as the concentrated distribution of revisit events in \figref{fig:kitti08_traj}. M2DP and \ac{CC} failed due to the severe rotational variance while \ac{PC} showed substantially better precision. SegMatch yielded enough precision, but the recall is limited. For this sequence with rotational variance, we examined the \ac{A-CC} to see the improvement of the augmentation.


\subsubsection{\texttt{MulRan Riverside 02}} In this sequence, the vehicle revisits a place with multiple lane changes but in the same direction. This variance is clearly captured in \figref{fig:riverside02_traj}. In terms of precision-recall, \ac{CC} outperformed the others techniques by large margin (\figref{fig:riverside02_pr}). The time-elevation graph in \figref{fig:matchedriver} shows true/false matches for the sequence. \ac{CC} outperformed the others in challenging regions with few false positives (red), which can potentially be treated using existing robust back-ends \cite{sunderhauf2012switchable, agarwal2013robust, rosinol2019kimera}. As with the improvement of A-CC in \texttt{KITTI 08}, the augmentation (\ac{A-PC}) improved the \ac{PC} under lateral variance.



\subsection{Concurrent Rotational and Lateral Variance}
\label{sec:rotlatinv}

The more complex case includes concurrent rotational and lateral variance. We used \texttt{Oxford} and \texttt{Pangyo} to evaluate performance under composite variance. We excluded SegMatch for the composite cases because of their high dependency on odometry. Enhancing other prior modules for place recognition is beyond the scope of this paper.

\subsubsection{\texttt{Oxford}}
\label{sec:rotlatinvoxford}

As shown in \figref{fig:exp_oxford}, the performances of the original \ac{PC} and \ac{CC} without augmentation are steeply limited at a certain recall, even with increased thresholds. Interestingly, the unrecognized recalls at this steep point matched the ratio of non-same direction revisits (\unit{43}{\%}) shown in \tabref{tab:dataset}. Applying associated augmentation to \ac{PC} and \ac{CC} showed improved precisions at the higer recalls, with large margins for both descriptors. Overall, the \ac{A-CC} generally showed higher precision than the \ac{A-PC} did. In \figref{fig:matchedoxford}, true/false matches for each method are visualized. For a fair comparison, we pinned the recall at 50\% for all methods to measure each method's accuracy and effectiveness quantitatively.

Note the importance of the distribution shown in \figref{fig:exp_oxford}. According to this plot, CC outperforms PC in terms of precision and maximum F1 score, except for the distribution score. This indicates that the increased precision of CC is concentrated in easy regions, while PC can detect difficult loops that may critically contribute to SLAM performance (see \figref{fig:matchedoxford_pc}). However, the restricted performance of CC was alleviated by A-CC, as can be seen in the improved distribution score as shown in \figref{fig:oxfordjan11_dr}. This improvement is also depicted in \figref{fig:matchedoxford_acc}, in which A-CC detects well-distributed loop-closures.

\subsubsection{\texttt{NAVER LABS Pangyo}}
\label{sec:latinvpangyo}

This \texttt{Pangyo} sequence includes sporadic lane changes during revisits, accompanied by rotational change. This composite variance (both \textit{rotational and lateral}) is inevitable in an urban environment when the reverse route necessarily involves a lane change. The \texttt{Pangyo} sequence encompasses abundant types of variance as can be seen in \figref{fig:exp_pangyo}. Overall, augmentation yielded substantial improvement when the revisit underwent composite variance. A-PC showed the best performance for rotational change (\figref{fig:pangyo_dr3}) and M2DP was meaningful for lateral variance. However, under concurrent rotational and lateral variance, A-CC proved its validity over other methods.

\begin{figure}[!t]
\centering
  \subfigure[Trajectory ]{%
    \includegraphics[width=0.24\columnwidth, trim = 0 0 0 0, clip]{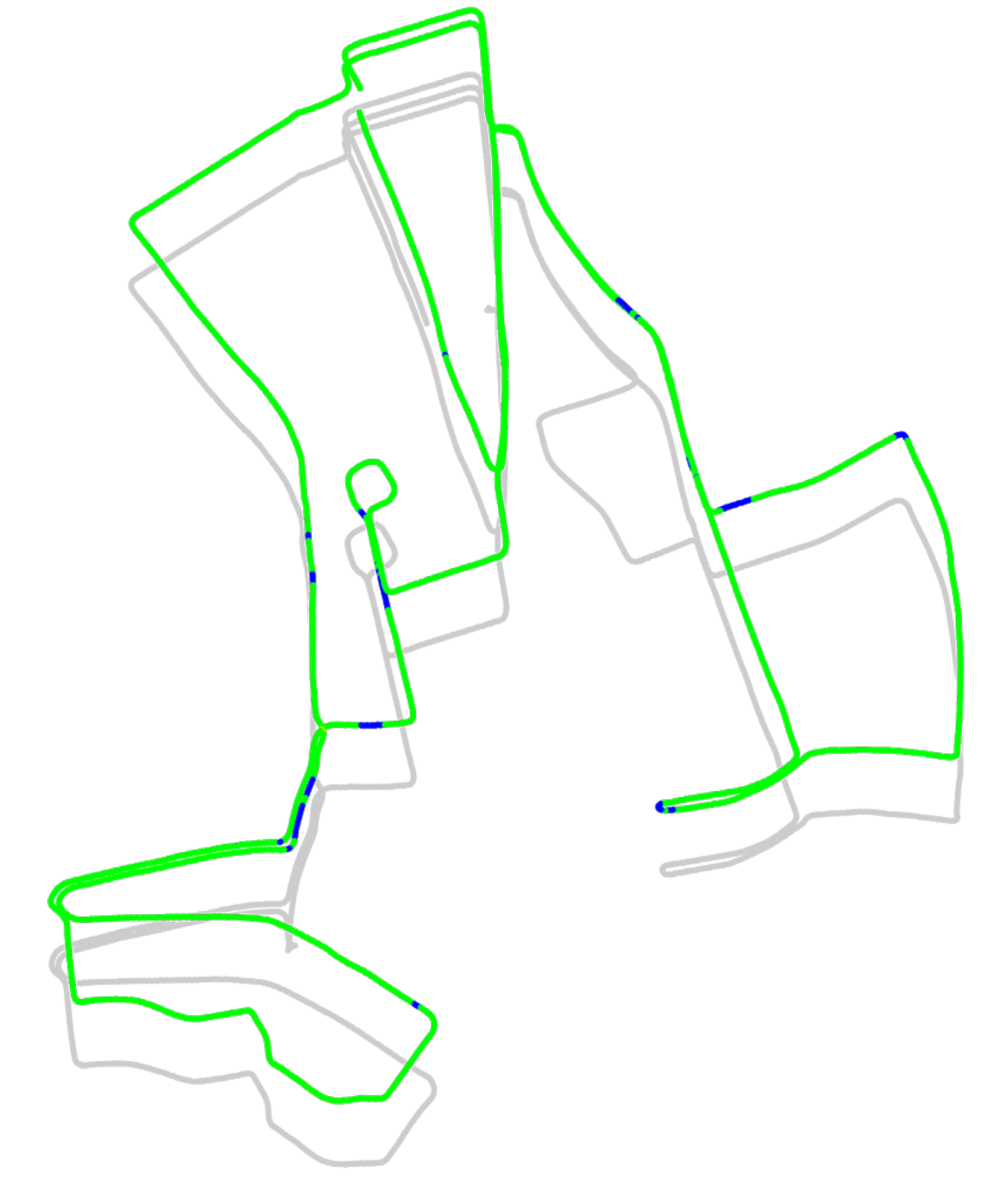}
    \label{fig:oxfordmulti_traj}
  }%
  \subfigure[Lateral displacement]{%
    \includegraphics[width=0.33\columnwidth, trim = 0 -200 0 0, clip]{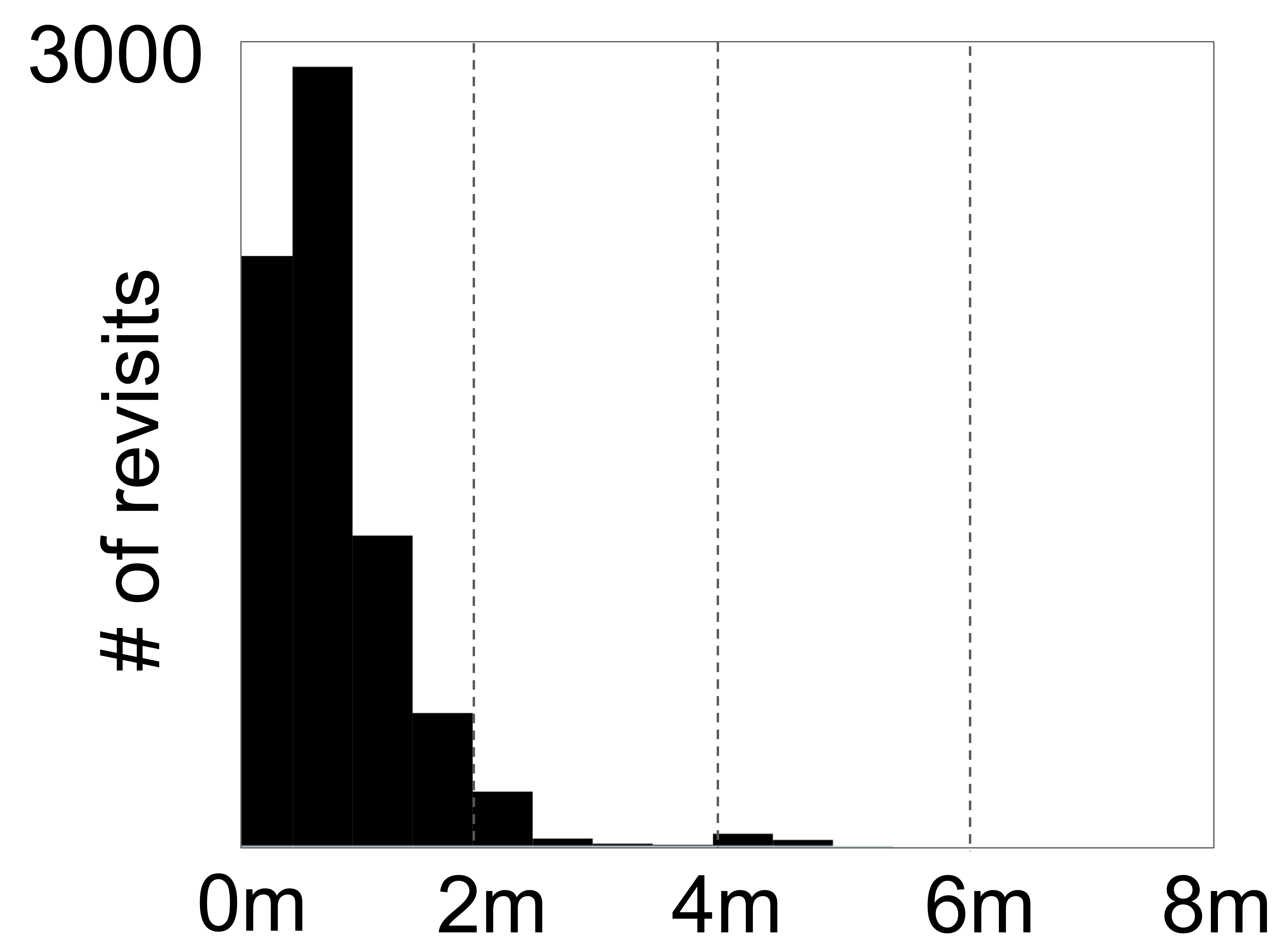}
    \label{fig:oxfordmulti_distbution}
  }%
  \subfigure[ PR curve ($\uparrow$)]{%
      \includegraphics[width=0.35\columnwidth, trim = 0 0 0 0, clip]{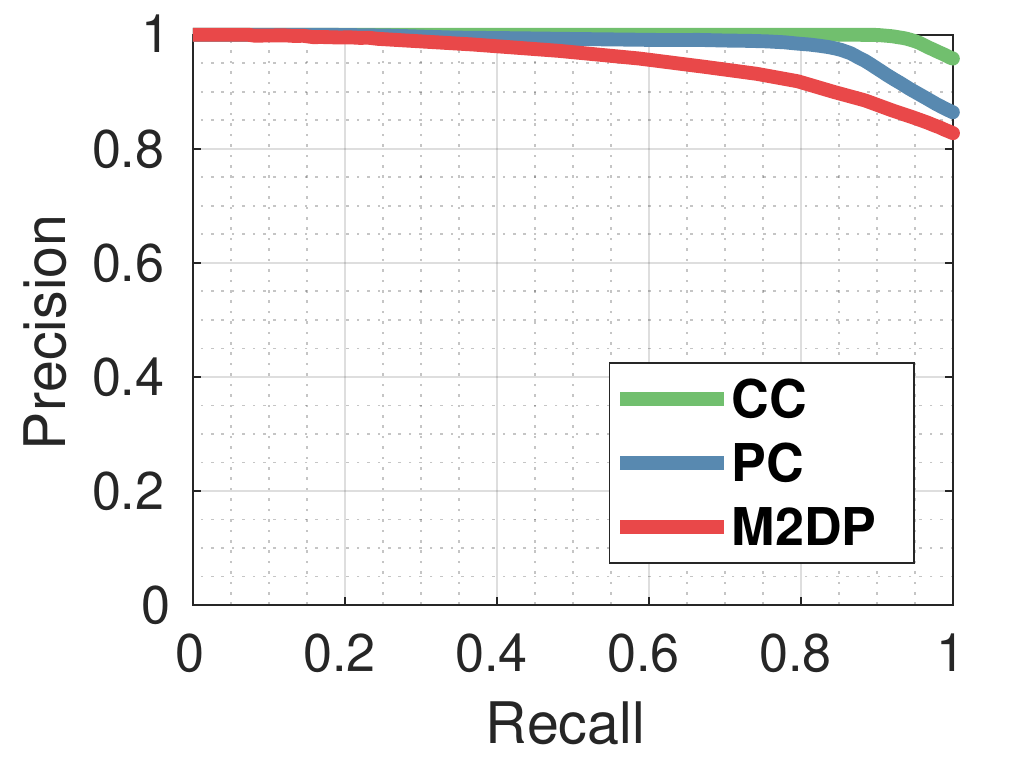}
      \label{fig:oxfordmulti_pr}
  }\\
  \subfigure[ Trajectory ]{%
    \includegraphics[width=0.24\columnwidth, trim = 0 -150 0 0, clip]{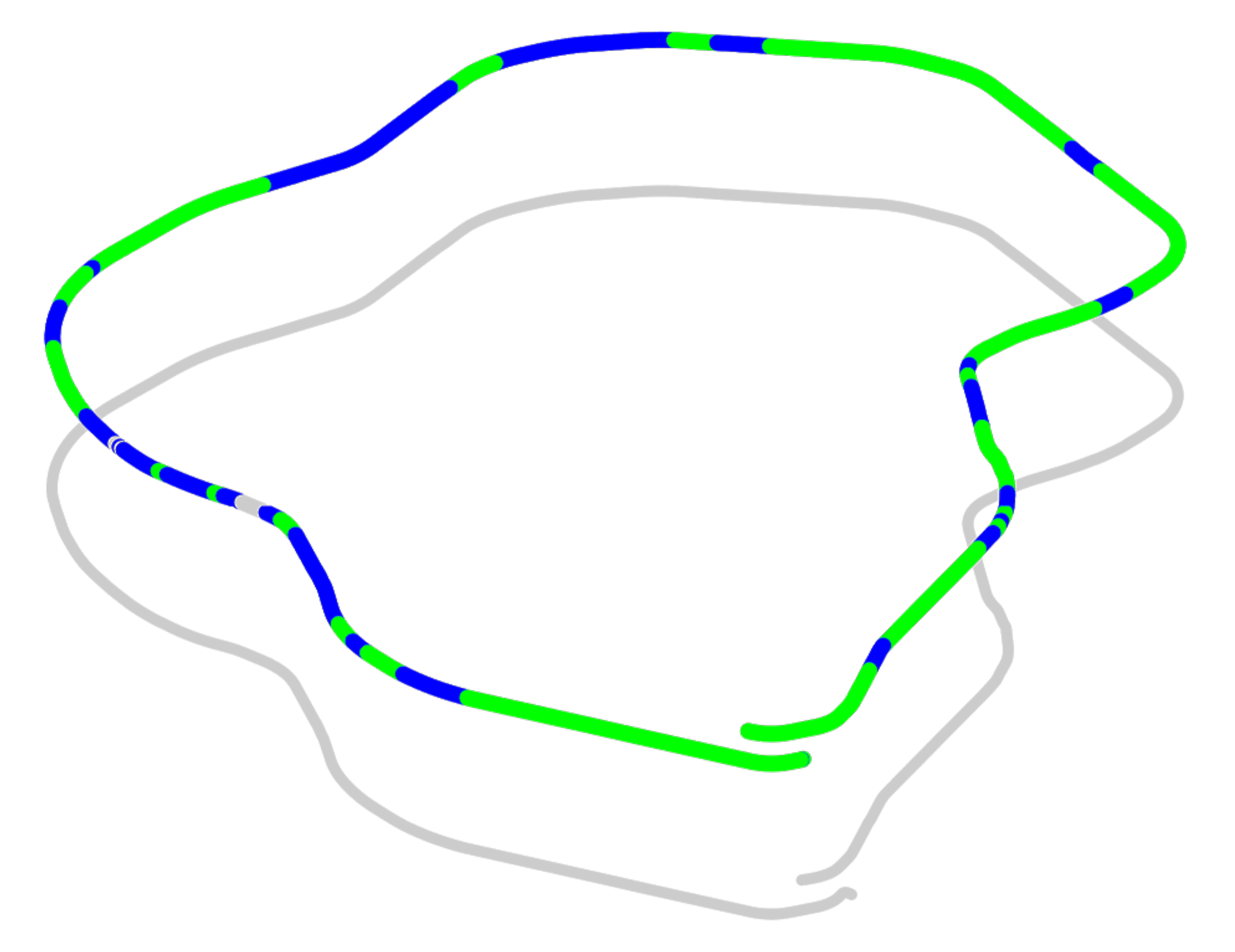}
    \label{fig:sejongmulti_traj}
  }%
  \subfigure[Lateral displacement]{%
    \includegraphics[width=0.33\columnwidth, trim = 0 -180 0 0, clip]{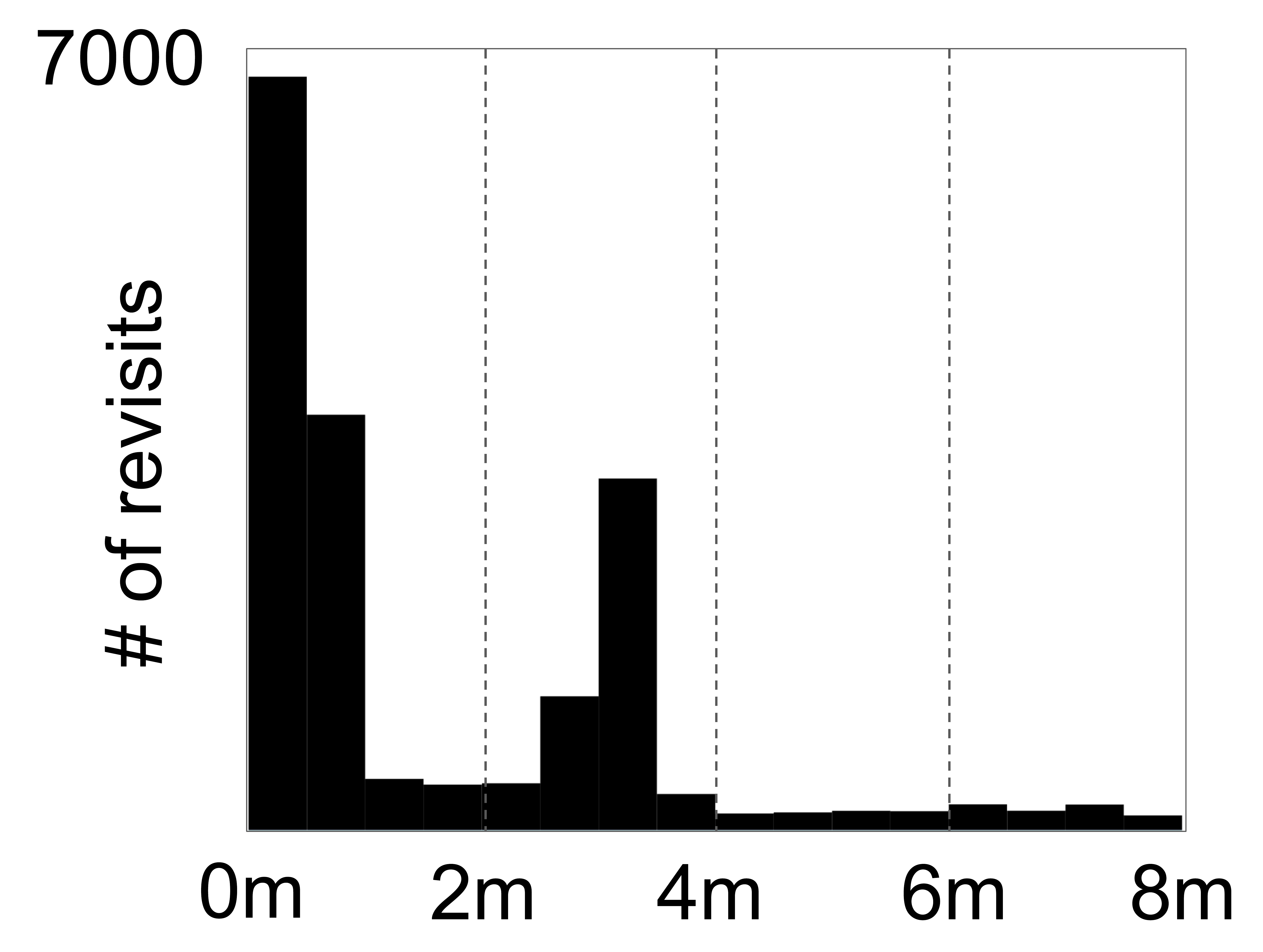}
    \label{fig:sejongmulti_distribution}
  }%
  \subfigure[ PR curve ($\uparrow$) ]{%
    \includegraphics[width=0.35\columnwidth, trim = 0 0 0 0, clip]{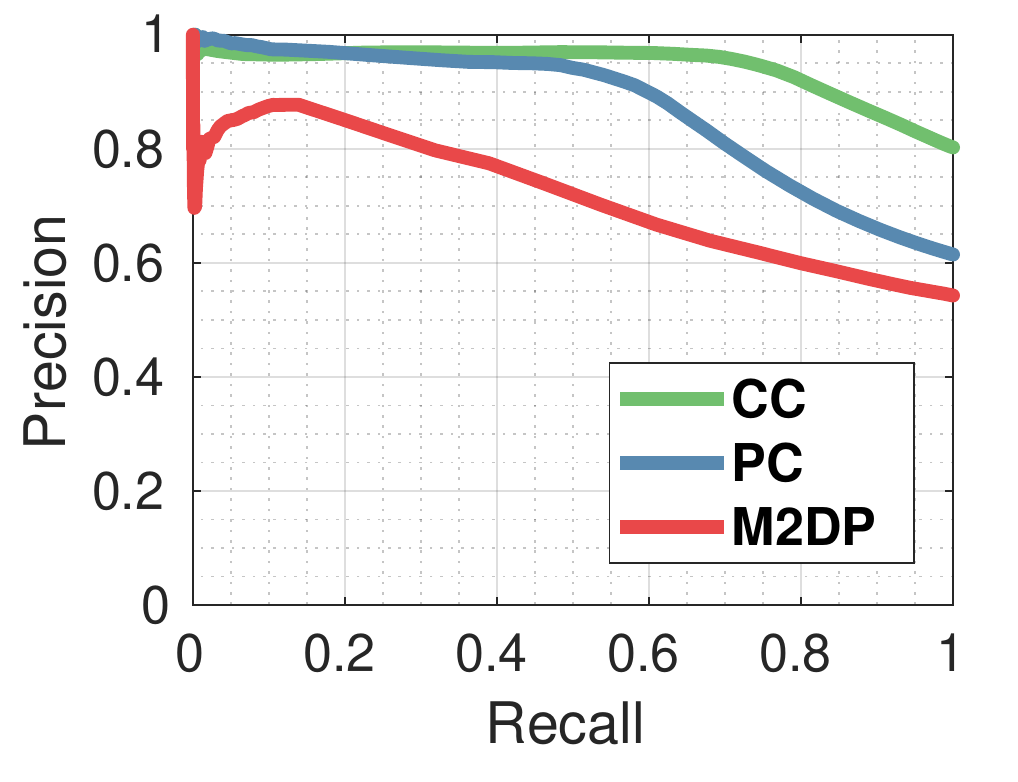}
    \label{fig:sejongmulti_pr}
  }%
  \caption{Loop-closure detection under multi-session scenarios. (Top) A pair from the \texttt{Oxford} dataset and (bottom) a pair from the \texttt{Sejong} of \texttt{MulRan} dataset were used. The revisits mostly had lateral changes. Single peak in \figref{fig:oxfordmulti_distbution} and double peaks in \figref{fig:sejongmulti_distribution} describes the number of laterally displaced revisits in each sequence.}
  \label{fig:exp_multisession}

\end{figure}

\begin{figure}[!t]
  \centering
  \def\mywidth{0.32\columnwidth}%
  \subfigure[ M2DP ]{%
    \centering
    \includegraphics[width=\mywidth, trim = 520 130 590 150, clip]{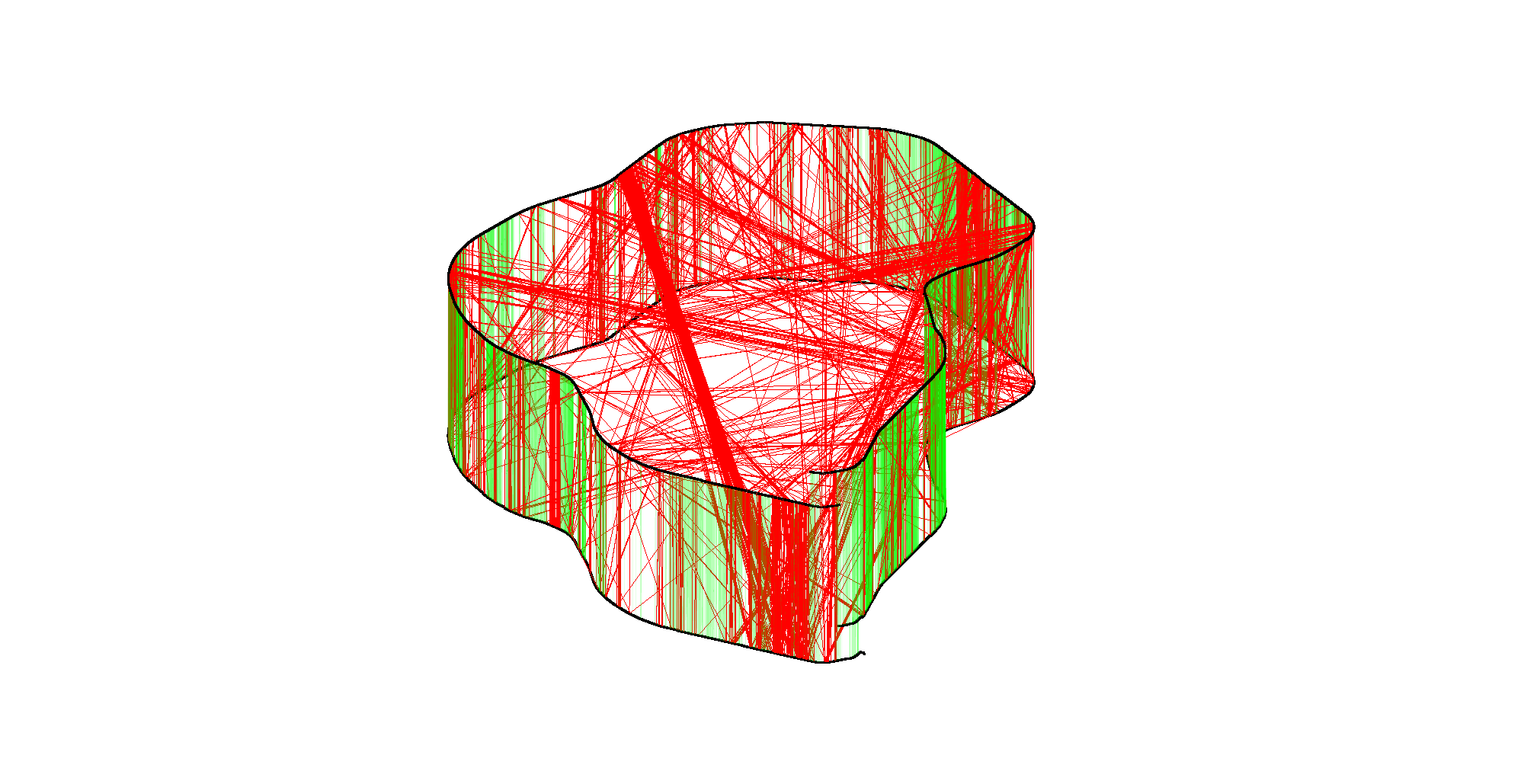}
    \label{fig:matchedsejong1}
  }%
  \subfigure[ Polar Context ]{%
    \centering
    \includegraphics[width=\mywidth, trim = 360 80 430 110, clip]{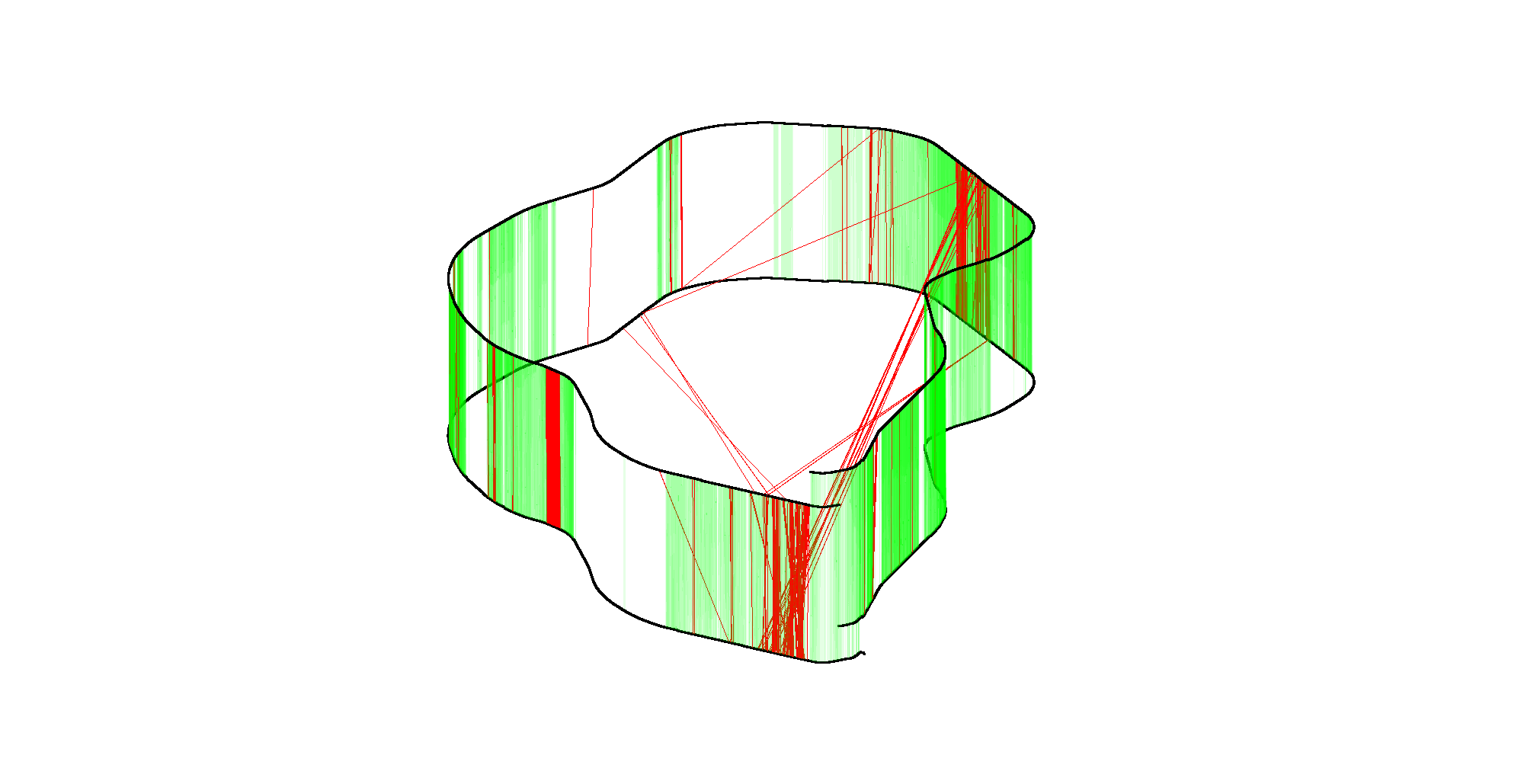}
    \label{fig:matchedsejong2}
  }%
  \subfigure[ Cart Context ]{%
    \centering
    \includegraphics[width=\mywidth, trim = 360 80 430 110, clip]{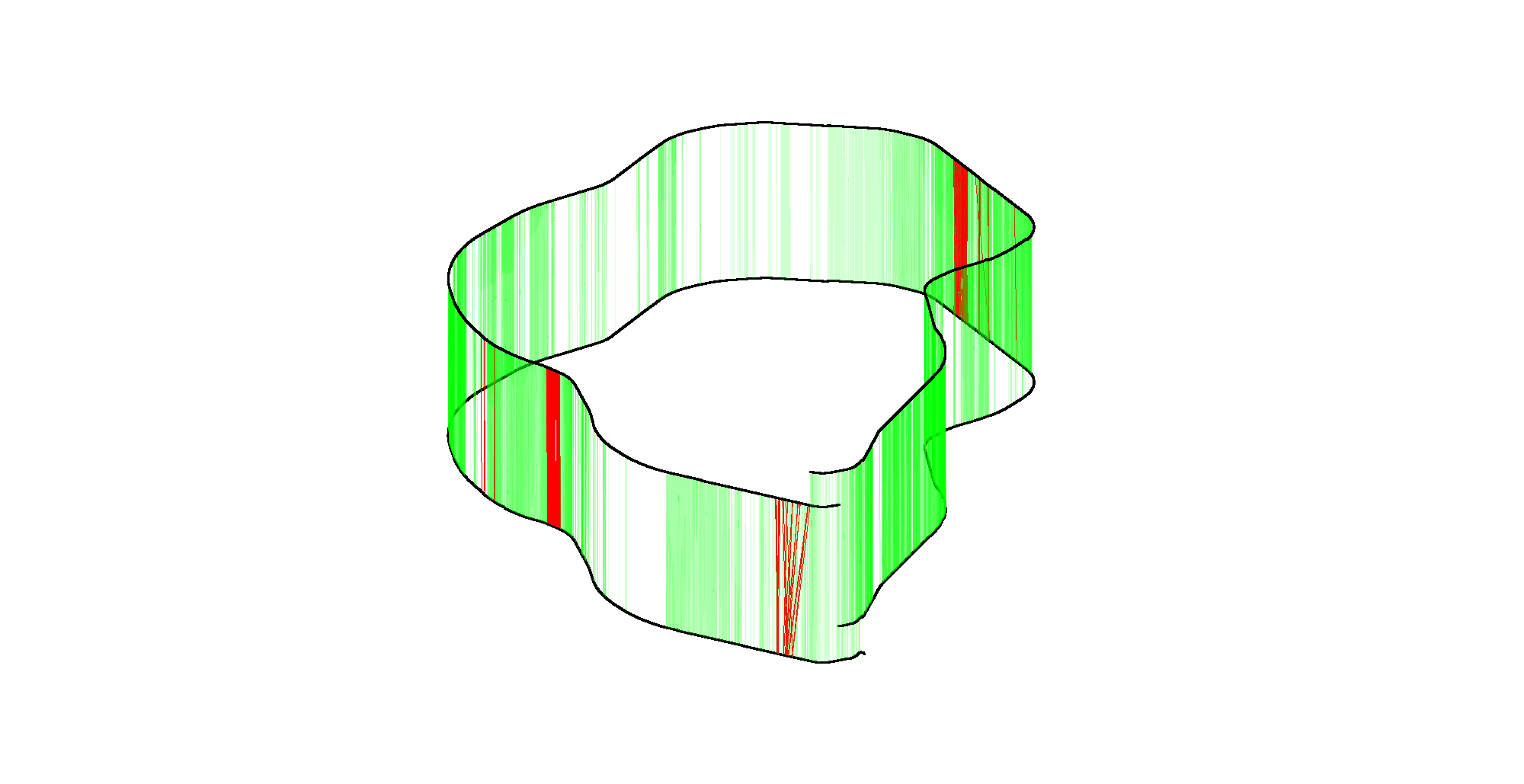}
    \label{fig:matchedsejong3}
  }%
  \caption{Inter-session place recognition visualization registering the query sequence (\texttt{Sejong 02}) to the map sequence (\texttt{Sejong 01}). A set of true (green)/false (red) loop detection results at recall of 50\% are visualized. The black line is the sequence trajectory whose height represents the time. The performance during severe lane change (blue in  \figref{fig:sejongmulti_traj}) is of interest. Only \ac{CC} had green matches even under major lateral variance while suppressing perceptual aliasing (red lines).}
  \label{fig:matchedsejong}
  
\end{figure}


\subsection{Multi-session Capability}
\label{sec:evalmultisession}

\begin{figure*}[!t]
  \centering
  \begin{minipage}{0.9\textwidth}
  \subfigure[Initial relative rotation inference (A-PC) ]{%
    \includegraphics[width=0.72\textwidth, trim = 0 0 0 0, clip]{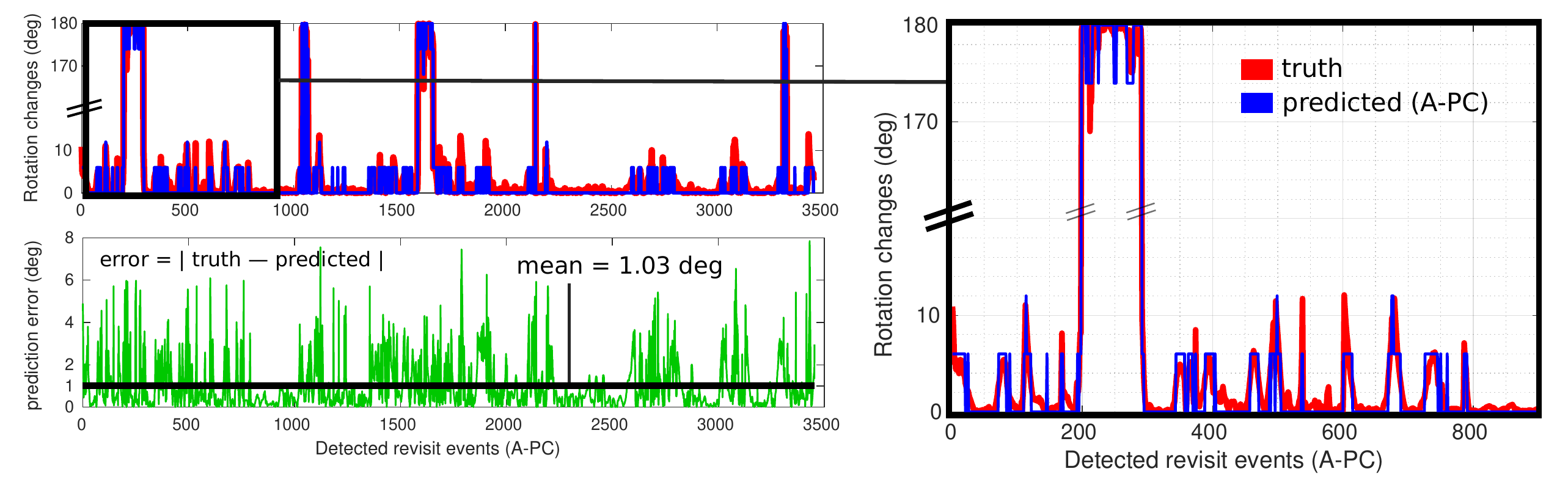}
    \label{fig:semilocrot1}
  }%
  \subfigure[RMSE and SCD distance]{%
    \includegraphics[width=0.23\textwidth, trim = 0 0 0 0, clip]{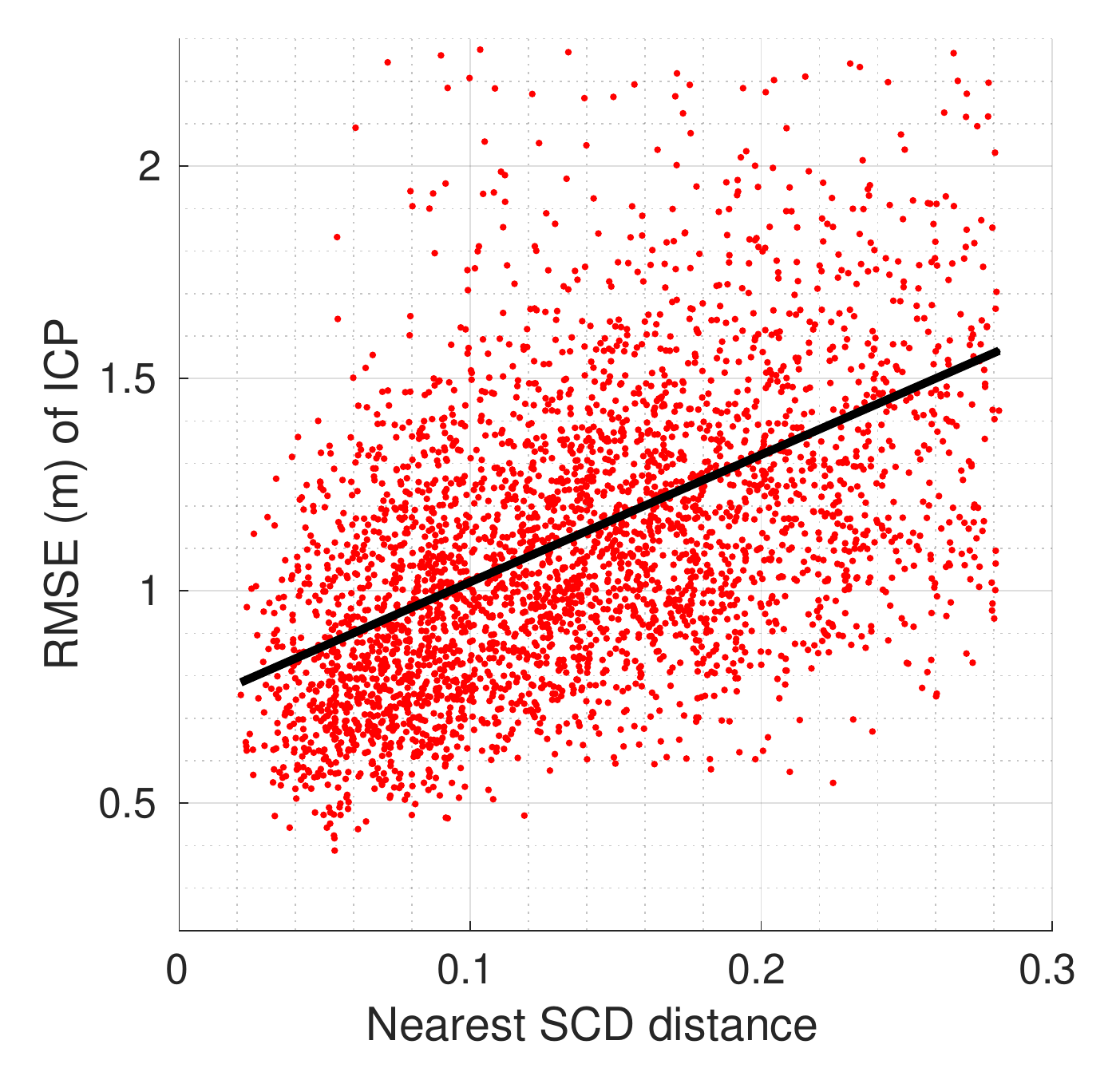}
    \label{fig:semilocrot2}
  }\\
  \subfigure[Initial relative translation inference (A-CC) ]{%
    \includegraphics[width=0.72\textwidth, trim = 0 0 0 0, clip]{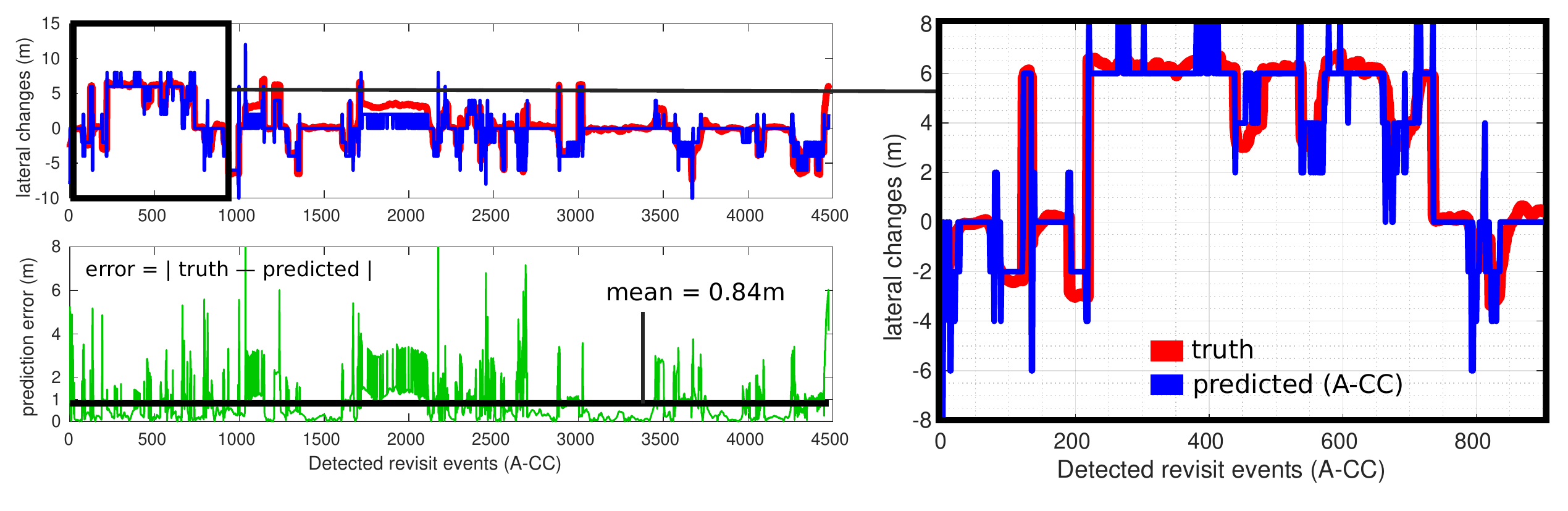}
    \label{fig:semiloclat1}
  }%
  \subfigure[RMSE and SCD distance]{%
    \includegraphics[width=0.23\textwidth, trim = 0 -30 0 0, clip]{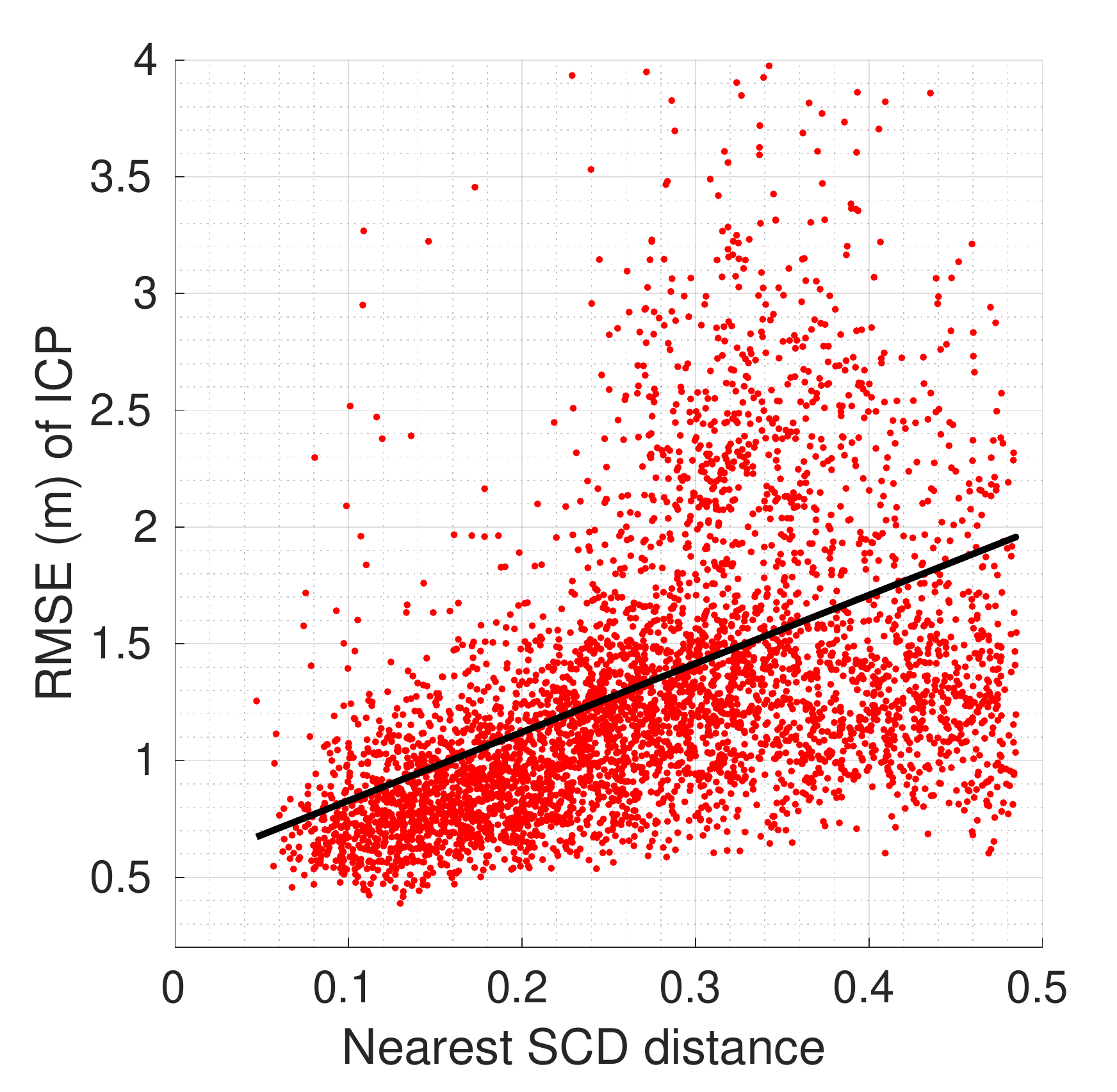}
    \label{fig:semiloclat2}
  }%
  \end{minipage}
  \caption{Semi-metric localization at max F1 score on \texttt{Pangyo} sequence. \subref{fig:semilocrot1} and \subref{fig:semiloclat1} Relative 1D pose estimation via aligning key matching. Relative rotation (A-PC) and lateral translation (A-CC) are depicted in blue while the red line indicates the true relative displacement obtained from the ground truth. Error is illustrated in green below the estimation plot. \subref{fig:semilocrot2} and \subref{fig:semiloclat2}, plotting RMSE between relative pose after ICP and ground truth against the scan context descriptor distance reveals correlation.}
  \label{fig:semiloc}

\end{figure*}

So far, we have investigated revisits within a single session. Here, we consider place recognition in multi-session scenarios toward long-term autonomy. To validate our methods in multi-session scenarios, we chose two sequences from a dataset with sufficient temporal differences.

The first pair was \texttt{Oxford 2019-01-15-13-06-37} to \texttt{Oxford 2019-01-11-13-24-51}. We used \texttt{Oxford
2019-01-11-13-24-51} as a map and tested the loop-closure performance of \texttt{Oxford 2019-01-15-13-06-37} as a query sequence. Another pair used for testing multi-session loop-closure showed a larger temporal gap of two months. We chose \texttt{Sejong 01} in the \texttt{MulRan} dataset as a map using \texttt{Sejong 02} as a query.

As can be seen in \figref{fig:exp_multisession}, these revisits mostly included lateral variance but with a temporal gap. For the \texttt{Oxford} pair, all of the methods successfully detected loops. The \texttt{Sejong} pair was more challenging because the lateral change included multiple lane changes. The loop-closure results for the \texttt{Sejong} pair are further visualized in \figref{fig:matchedsejong}. M2DP seemed to show meaningful performance but included wrong loop-closures. CC the showed best performance for the multi-session scenarios. Obvious improvements could be made via augmentation, although this was excluded from the multi-session scenarios.

\subsection{Metric Localization Evaluation and Quality Assessment}
\label{sec:icp}

Together with the retrieved place, the proposed method is capable of estimating the relative 1D pose between query and map places. This is important when the topological place retrieval is combined with the metric localization because this initial estimate can be exploited in further metric refinement.

From the aligning key registration, we estimate a 1D relative pose (i.e., rotation for PC and lateral displacement for CC). Using the ground truth pose provided in \texttt{Pangyo} sequence, we plot the estimated 1D relative pose against the true relative pose from the ground truth. As can be seen in \figref{fig:semilocrot1} and \figref{fig:semiloclat1}, the estimation yielded a meaningful relative pose inference of \unit{1.03}{\degree} for A-PC and \unit{0.84}{m} for A-CC on average.

We can further examine the quality of this metric localization using the full descriptor similarity score. In the proposed method, we utilized this similarity score as the second barometer to exclude the retrieval with large SCD distance (i.e., small similarity). To assess the metric evaluation quality, we present a scatter plot between the RMSE of the relative estimation from ICP and the SCD distance in \figref{fig:semiloc}.

\begin{table}[!b]

  \caption{
    \bl{ATEs (Mean / Max) of odometry and Scan Context integrated SLAM system.}
  }
    \centering
  \scriptsize
  \resizebox{0.95\columnwidth}{!}{
    \begin{tabular}{l|cc|cc}
    \hline
        & \multicolumn{2}{c|}{ \texttt{KAIST 03} }  & \multicolumn{2}{c}{ \texttt{Riverside 02}  }\\
    \toprule[1.1pt]
    Methods        & Trans. (m)     &  Rot. (deg)   & Trans. (m)         &  Rot. (deg)  \\
    \midrule[0.5pt]
    LeGO-LOAM      & 20.7 / 42.7 & 4.9 / 9.9    & 47.7 / 130.5    &  6.9 / 12.8  \\
    SC-LeGO-LOAM   & 3.4 / 8.8   & 2.2 / 8.2    & 15.2 / 50.5     &  4.2 / 8.7  \\
    \bottomrule[1.2pt]
  \end{tabular}
}
\vspace{2mm}
\label{tab:evo}
\end{table}

\label{sec:slam}
\subsection{External Module Dependence and SLAM Integration}

Being lightweight and independent to an external module would be needed in a global localizer. We aimed to develop a stand-alone module without requiring prior information such as odometry. During evaluation, we found that SegMatch's place recognition performance is affected by the odometry quality. We also empirically discovered that the SegMatch hardly made recalls for harsh environments such as the \texttt{Riverside 02 (MulRan)} sequence when a good quality of frame-to-frame odometry barely obtainable due to many dynamic objects. In the previous evaluations, although we fed the ground truth as the odometry to ensure their best performance, the performance was restricted for less structured and repeated environments.

The proposed implementation is lightweight provided in a single C++ and header file pair. Thus, ours is easy to combine with any keyframe-based pose-graph \ac{SLAM} system because the Scan Context-based place recognition's atomic element is a single keyframe measurement. Along our open source place recognition module\footnote{https://github.com/gisbi-kim/scancontext}, we also made a real-time LiDAR SLAM system publicly available. It is written in C\texttt{++} and named SC-LeGO-LOAM\footnote{https://github.com/gisbi-kim/SC-LeGO-LOAM} integrated with LeGO-LOAM \cite{shan2018lego}. \bl{As in \tabref{tab:evo}, the Scan Context-based loop detection and pose-graph optimization with iSAM2 \cite{kaess2012isam2} successfully alleviated the odometry trajectory's drifts. For a detailed demonstration, we refer to the attached multimedia file.}

\subsection{Computational Cost}
\label{sec:timecost}

\begin{figure}[!h]

  \centering
  \subfigure[\texttt{KITTI 00}. Mean \unit{7.36}{ms}]{%
    \includegraphics[width=0.99\columnwidth,  trim = 0 0 0 0, clip]{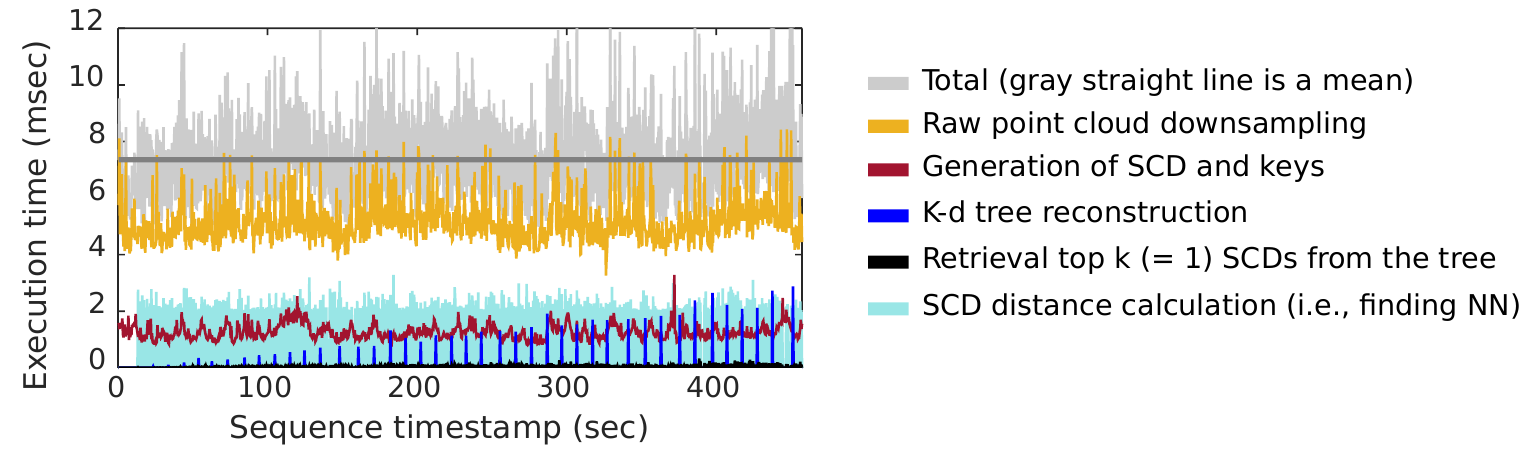}
    \label{fig:timecost1}
  }
  \subfigure[\texttt{NAVER LABS Pangyo} ($\sim$ \unit{31}{km}). Mean \unit{7.31}{ms}]{%
    \includegraphics[width=0.99\columnwidth,  trim = 0 0 0 0, clip]{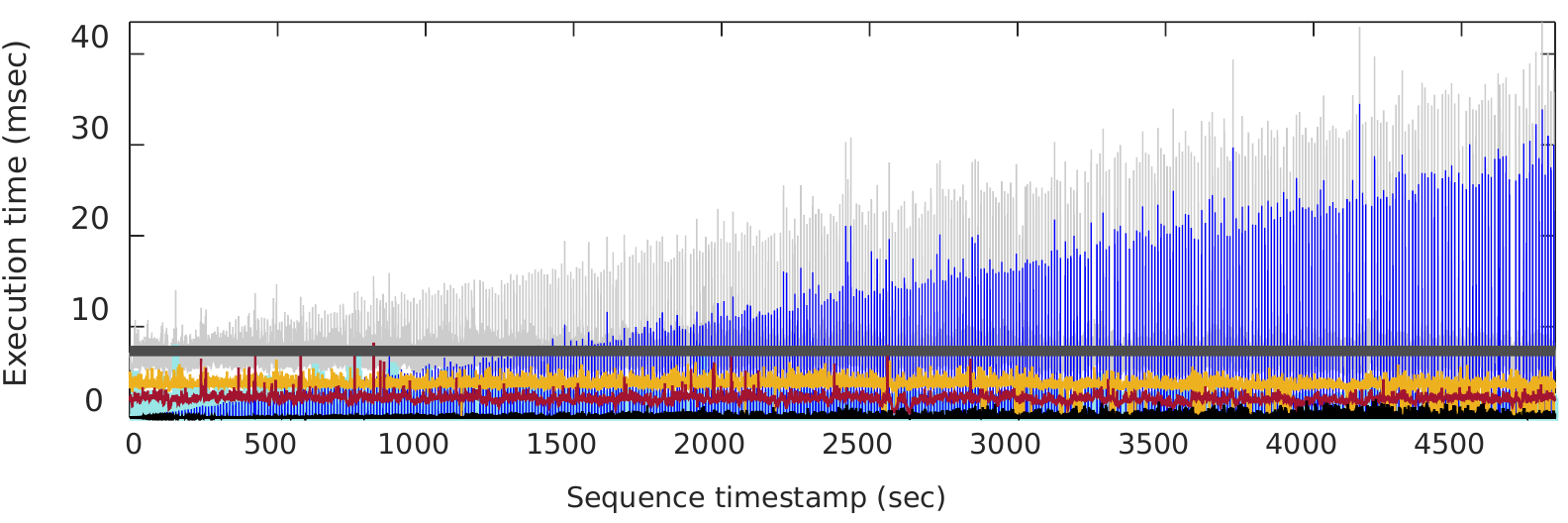}
    \label{fig:timecost2}
  }
  \caption{
    Computational time visualization per module when tested over \texttt{KITTI00} and \texttt{Pangyo}. Conservative batch tree update every other 10 seconds was performed.}
  \label{fig:timecost}
   
\end{figure}

The proposed place descriptor \bl{generation} and recognition modules are both fast. The per module computational costs are visualized in \figref{fig:timecost} for two sequences. \ac{PC}'s computation costs are reported in \figref{fig:timecost} because the computational costs for PC and CC are almost the same under a similar resolution and only the coordinate selections differed. These time consumptions are measured while running the Scan Context integrated real-time LiDAR \ac{SLAM} (\secref{sec:slam}) on Intel i9-9900 CPU (3.10GHz) and 64G RAM.



As can be seen in \figref{fig:timecost2}, the mean computational time is less than \unit{10}{ms}. The most time consuming task is the k-d tree reconstruction, which is performed periodically in batches. However, a graph plots the conservative case when we repeatedly rebuild the tree every other \unit{10}{secs}. This could be elongated depending on the application to reduce the total cost and is not even required for the multisession scenario.

The mean execution time is even shorter at \texttt{Pangyo} despite its large scale because \texttt{Pangyo} used 32-ray \ac{LiDAR} with fewer points than \texttt{KITTI 00}. This also indicates that the overall computational complexity is $\mathcal{O}(1)$ though periodic batch tree rebuilding scales linearly with the map $\mathcal{O}(N)$, with $N$ being the number of nodes in the map.

\bl{
The time cost comparisons with the other methods are given in \tabref{tab:times}. The timing for ours and M2DP were measured using Matlab, while SegMatch was copied from \cite{dube2017segmatch}. SegMatch spent most of its time on segmentation. M2DP was the most lightweight. A-PC only requires extra description times during the augmentation phase as no extra costs managing retrieval keys are needed. Despite requiring GPU (GTX 1080 Ti), PointNetVLAD (\secref{sec:dl}) was more expensive than ours. The retrievals were fast for M2DP and PointNetVLAD because a fixed-length vector's comparison of Euclidean distance is very lightweight.
}

\begin{table}[!h]
  \vspace{-3mm}
  \caption{ 
    \bl{Time cost comparison. All units are in ms. } 
  }
    \centering
  \resizebox{0.99\columnwidth}{!}{
  \begin{tabular}{l|cc|ccc}
  \hline
  \toprule[1.0pt]
                &  Ours (PC)   & Ours (A-PC)          & M2DP    & SegMatch   & PointNetVLAD \\
  \midrule[.6pt]
  Desription    & 1.6             & 4.8               & 4.3     & 430.2        & 33.3 \\
  Retrieval     & 6.7             & 6.7               & 1.5       & 365.8      & 0.7 \\
  \midrule[.3pt]
  Total   & 8.3             & 11.5              & 5.8     & 796.0        & 34.0  \\ 
  \bottomrule[1.2pt]
\end{tabular}
  }
\vspace{2mm}
\label{tab:times}
\vspace{-3mm}
\end{table}

\section{Discussion}
\label{sec:discussion}

Beyond the evaluation of the proposed global localization method, we provide ablation studies and interpretations.

\subsection{Descriptor Resolution}
\label{sec:res}

We examined the descriptor resolution and corresponding performances. As in \tabref{tab:abl_res}, the lower resolution yielded better performance. \bl{Therefore, we used the baseline resolution for the following subsections.}

\newcommand{\R}[1]{{\multirow{2}{*}{#1}}}

\begin{table}[!h]
  \vspace{-2mm}
  \centering
  \caption{Performance comparison with respect to the descriptor resolution \bl{at \texttt{Oxford 2019-01-11-13-24-51}}. P: precision, R: recall, D: KL-D at F1 max. The baseline resolution is marked with *.}
  \resizebox{0.85\columnwidth}{!}
  {
  \begin{tabular}{c|ccc|c|ccc}
    \hline
    \multicolumn{4}{c|}{\acf{PC}}                             & \multicolumn{4}{c}{\acf{CC}}\\
    \toprule[1.2pt]
    Resolution              &P ($\uparrow$) & R ($\uparrow$) & D ($\downarrow$)          & Resolution               & P ($\uparrow$) & R ($\uparrow$) & D ($\downarrow$) \\
    \midrule[0.9pt]
    10$\times$30            & \R{0.65} & \R{0.41} & \R{0.56}  & 40$\times$20             & \R{0.94} & \R{0.50} & \R{0.62}\\
    (\unit{8}{m}, 12$^\circ$) &          &          &           & (\unit{5}{m}, \unit{4}{m}) &          &          &\\\hline
    20$\times$60*           & \R{0.81} & \R{0.47} & \R{0.58}  & 40$\times$40*            & \R{0.93} & \R{0.53} & \R{0.60}\\
    (\unit{4}{m}, 6$^\circ$)  &          &          &           & (\unit{5}{m}, \unit{2}{m}) &          &          &\\\hline
    20$\times$120           & \R{0.76} & \R{0.50} & \R{0.57}  & 40$\times$60             & \R{0.90} & \R{0.52} & \R{0.61}\\
    (\unit{4}{m}, 3$^\circ$)  &          &          &           & (\unit{5}{m}, \unit{1.3}{m}) &          &          &\\\hline
    40$\times$60            & \R{0.77} & \R{0.50} & \R{0.58}  & 60$\times$40             & \R{0.92} & \R{0.52} & \R{0.60}\\
    (\unit{2}{m}, 6$^\circ$)  &          &          &           & (\unit{3.3}{m}, \unit{2}{m}) &        &          &\\\hline
    40$\times$120           & \R{0.76} & \R{0.50} & \R{0.59}  & 80$\times$80             & \R{0.92} & \R{0.51} & \R{0.61}\\
    (\unit{2}{m}, 3$^\circ$)  &          &          &           & (\unit{2.5}{m}, \unit{1}{m}) &        &          &\\\hline
    60$\times$180           & \R{0.74} & \R{0.49} & \R{0.61}  & 120$\times$120            & \R{0.91} & \R{0.52} & \R{0.60}\\
    (\unit{1.3}{m}, 2$^\circ$) &          &          &           & (\unit{1.6}{m}, \unit{0.6}{m}) &          &          &\\
    \bottomrule[1.2pt]
  \end{tabular}
  }
  \vspace{2mm}
  \label{tab:abl_res}
\end{table}

\subsection{Analysis on Retrieval key Performance}
\label{sec:topk}

\textbf{Candidate numbers.} In \secref{sec:pr}, we leveraged the k-d tree to propose $k$ candidates for retrieval and \bl{only a single answer is selected after the full descriptor-based false positive rejection (\secref{sec:fullscd})}. In this subsection, we examine the effect of $k$ on performance. \bl{We first note that the increase in $k$ does not mean to relax the success criteria, but rather the number of candidates in the first step of our algorithm.} Interestingly, \bl{as in \tabref{tab:topk},} all statistics outperformed others when we only chose the best candidate. \bl{The full descriptor may suffer confusion showing the best performance at $k=1$. Though this may seem contrary to a general belief for better performance under more candidates, the result indicates the reduced spatial discernibility of the full descriptor. Based on this investigation, we used $k=1$ for all experiments conducted earlier.}

\begin{table}[!h]
  \vspace{-2mm}
  \caption{Performance comparison \bl{with respect to the number of first stage's candidates} \bl{at \texttt{Oxford 2019-01-11-13-24-51}}. P: precision, R: recall, D: KL-D at F1 max. The baseline resolution is marked with *.}
  \centering
  \resizebox{0.65\columnwidth}{!}{
  \begin{tabular}{l|ccc|ccc}
  \hline
      & \multicolumn{3}{c|}{\acf{PC}}  & \multicolumn{3}{c}{\acf{CC}}\\
  \toprule[1.2pt]
  K   & P ($\uparrow$) & R ($\uparrow$) & D ($\downarrow$)   & P ($\uparrow$) & R ($\uparrow$) & D ($\downarrow$)\\
  \midrule[0.6pt]
  1*  & \B{0.81} & 0.48 & \B{0.58} & \B{0.93} & \B{0.53} & \B{0.60}\\
  10  & 0.76 & \B{0.49} & 0.60             & 0.87 & 0.51 & 0.61\\
  50  & 0.72 & 0.49 & 0.59             & 0.76 & 0.51 & 0.61\\
  100  & 0.71 & 0.49 & 0.59             & 0.70 & 0.49 & 0.62\\
  \bottomrule[1.2pt]
\end{tabular}
}
\label{tab:topk}
\end{table}

\bl{\textbf{Retrieval key vs. full descriptor brute-force search.} Additionally, we analyzed how the performance varies if an entire database is compared (i.e., brute-force) using the full descriptor-based distance \equref{eq:prealign2}. Through the multiple tests in \tabref{tab:bf}, the performance difference between the retrieval key-based and the brute-force search is negligible although the brute-force search requires heavier computations following $\mathcal{O}(n)$ (e.g., almost 1 second for 4500 frames of \texttt{KITTI 00}).
}

\begin{table}[!h] 
  \vspace{-5mm}
  \caption{
    \bl{
      Full descriptor similarity (brute-force search) and retrieval key comparison. The measure is an area under a precision-recall curve (AUC). 
    }
  }
    \centering
  \scriptsize
  \resizebox{\columnwidth}{!}{
    \begin{tabular}{l|cc|cc}
    \hline
        & \multicolumn{2}{c|}{\acf{PC}}  & \multicolumn{2}{c}{\acf{CC}}\\
    \toprule[1.1pt]
    Sequences   & Retrieval key &  Full descriptor  & Retrieval key &  Full descriptor  \\
    \midrule[.6pt]
    \texttt{KITTI 00}      & 0.84 & 0.85 & 0.80 & 0.34  \\
    \texttt{KAIST 03}      & 0.99 & 0.99 & 0.99 & 0.99  \\
    \texttt{Riverside 02}  & 0.72 & 0.74 & 0.88 & 0.84  \\
    \texttt{KITTI 08}      & 0.55 & 0.46 & 0.00 & 0.00  \\
    \bottomrule[1.2pt]
  \end{tabular}
}
\label{tab:bf}
\end{table}

\begin{figure}[!h]
  \centering
  \includegraphics[width=0.7\columnwidth, trim = 0 10 -60 0, clip]{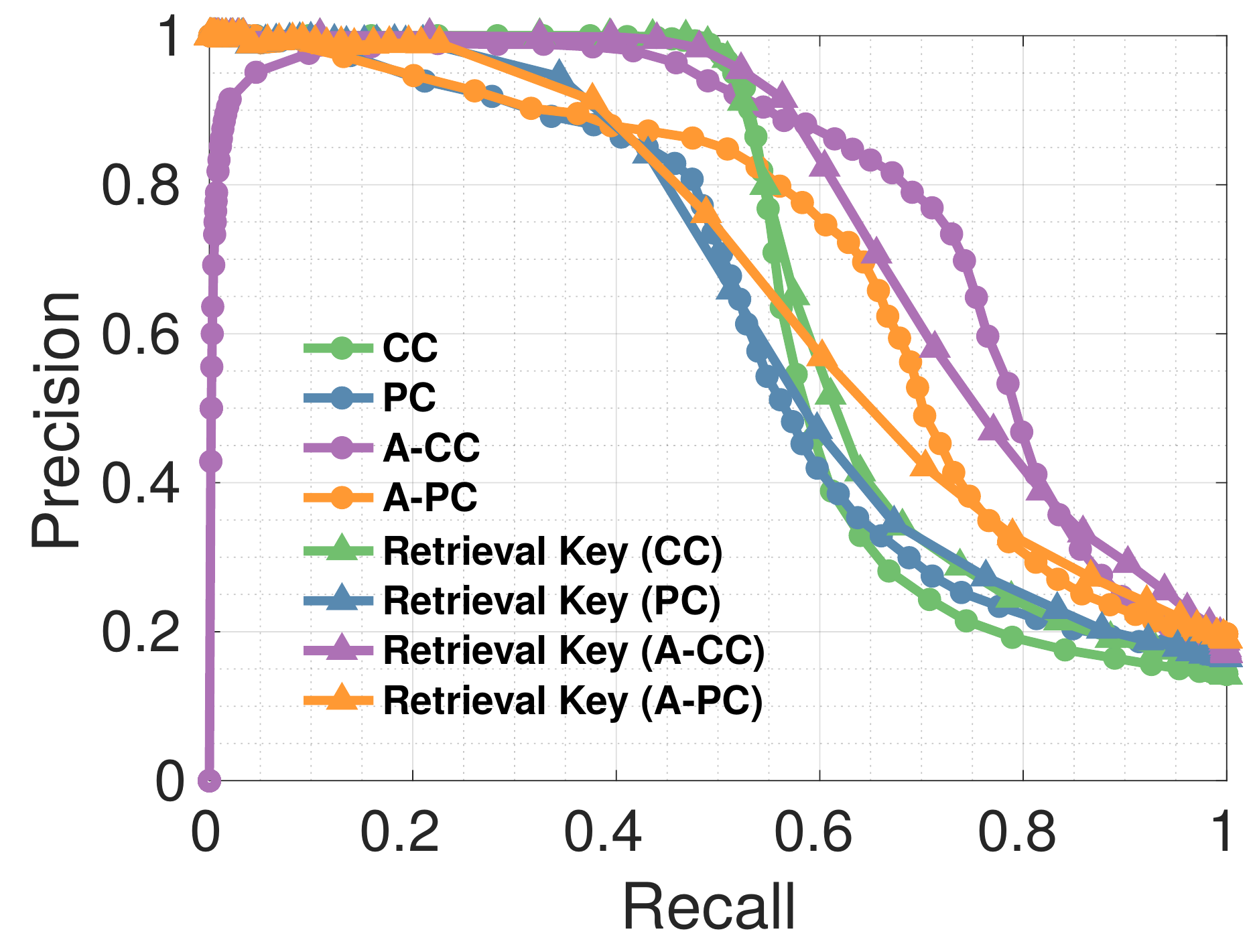}
  \caption{Effect of the retrieval key in terms of the PR curve \bl{(\texttt{Oxford 2019-01-11-13-24-51})}. The same SCD is depicted with the same color. A-CC and A-PC showed improvements when additional full descriptor similarity was considered.}
  \label{fig:keyeffect}
\end{figure}

\textbf{Full descriptor's effect.} \bl{Nevertheless the confusion of the full descriptor-based similarity proven in \tabref{tab:topk}, this additional similarity validation enhanced the precision for the augmentation cases as well as its semi-metric localization capability.} In \figref{fig:keyeffect}, the augmented Scan Context's precision was improved by eliminating less accurate matches via the supplemental similarity verification using a full descriptor.

\subsection{Correctness Criteria}
\label{sec:thres}
\bl{
The performance tends to be improved when more tight criteria are applied as in \tabref{tab:thres}. This phenomenon occurred because laterally displaced queries, which are generally difficult to recognize, are considered as correct rejections when they are missed. However, precise localization was more difficult in reversed revisits (i.e., \texttt{KITTI 08} in \tabref{tab:thres}) because the previously correctly recognized queries (e.g., within \unit{4}{m}$-$\unit{8}{m}) are considered false alarms. From these findings, \unit{8}{m} was used for the criteria of correctly recognized places during our main evaluations in \secref{sec:exp} and the ablations in \secref{sec:discussion} to successfully cope with laterally translated revisits. Even if a place is recognized from a slightly distant place (e.g., \unit{4}{m}$-$\unit{8}{m} apart), the proposed method can close a loop successfully because it provides a semi-metric localization result.
}

\begin{table}[!h]
  \vspace{-2mm}
  \caption{
    \bl{
      Performances (AUC) with respect to correctness threshold. The baseline(*) threshold is \unit{8}{m}. 
    }
  }
    \centering
  \scriptsize
  \resizebox{\columnwidth}{!}{
    \begin{tabular}{l|cccc|cccc}
    \hline
        & \multicolumn{4}{c|}{\acf{PC}}  & \multicolumn{4}{c}{\acf{CC}}\\
    \toprule[1.1pt]
    Sequences   & \unit{8}{m}* & \unit{4}{m} & \unit{2}{m} & \unit{1}{m} & \unit{8}{m}* & \unit{4}{m} & \unit{2}{m} & \unit{1}{m}  \\
    \midrule[.6pt]

    \texttt{KITTI 00}      & 0.84 & 0.88 & 0.91 & \textbf{0.94} & 0.81 & 0.85 & 0.87 & \textbf{0.88} \\  
    \texttt{KAIST 03}      & 0.99 & 0.99 & \textbf{0.99} & 0.96 & 0.99 & 0.99 & \textbf{0.99} & 0.96 \\
    \texttt{Riverside 02}  & 0.72 & 0.73 & \textbf{0.79} & 0.73 & 0.88 & \textbf{0.89} & 0.87 & 0.81 \\
    \texttt{KITTI 08}      & \textbf{0.55} & 0.46 & 0.38 & 0.23 & 0.00 & 0.01 & 0.01 & 0.00 \\
    \bottomrule[1.2pt]
  \end{tabular}
}
\vspace{-2mm}
\label{tab:thres}
\end{table}

\subsection{Robustness to Roll-Pitch and Height Perturbations}
\label{sec:rollpitch}

\bl{
The previously used datasets are mainly from wheeled platforms with little roll-pitch and height perturbations. However, sensor measurement variation can occur in terms of rotational and height variations between two scans. In this regard, we added the additional experiments on roll-pitch perturbed simulations and a real-world hand-held LiDAR experiment. We randomly pre-rotated an input scan with respect to both roll and pitch for the simulation. In the real-world hand-held LiDAR dataset, the height of the measurement origins varies slightly while a human navigator walks.
}

\bl{
  \textbf{Simulations.} The degree of pre-rotations is divided into three levels: [\unit{-5}{\degree}, \unit{5}{\degree}], [\unit{-10}{\degree}, \unit{10}{\degree}], and [\unit{-15}{\degree}, \unit{15}{\degree}]. The simulations are conducted for two sequences \texttt{KAIST 03} and \texttt{Riverside 02}, and performance losses are clearly observed for all methods. For the \texttt{KAIST 03} sequence, M2DP showed smaller performance drops than ours. However, in \texttt{Riverside 02}, the performance degradations became clear with respect to the degree of perturbation for all three methods. We believe that this topic, robust place recognition under severe roll-pitch variations, has still not been studied much, and it could be a valuable future academic research topic.
}

\begin{figure}[!h]

  \centering
  \def\mywidththird{0.3\columnwidth}%
  \def\mywidth{0.4\columnwidth}%
  \def\mysmallwidth{0.25\columnwidth}%

  \subfigure[ PC - \texttt{KAIST 03} ]{%
    \centering
    \includegraphics[width=\mywidththird, trim = 160 280 190 295 , clip]{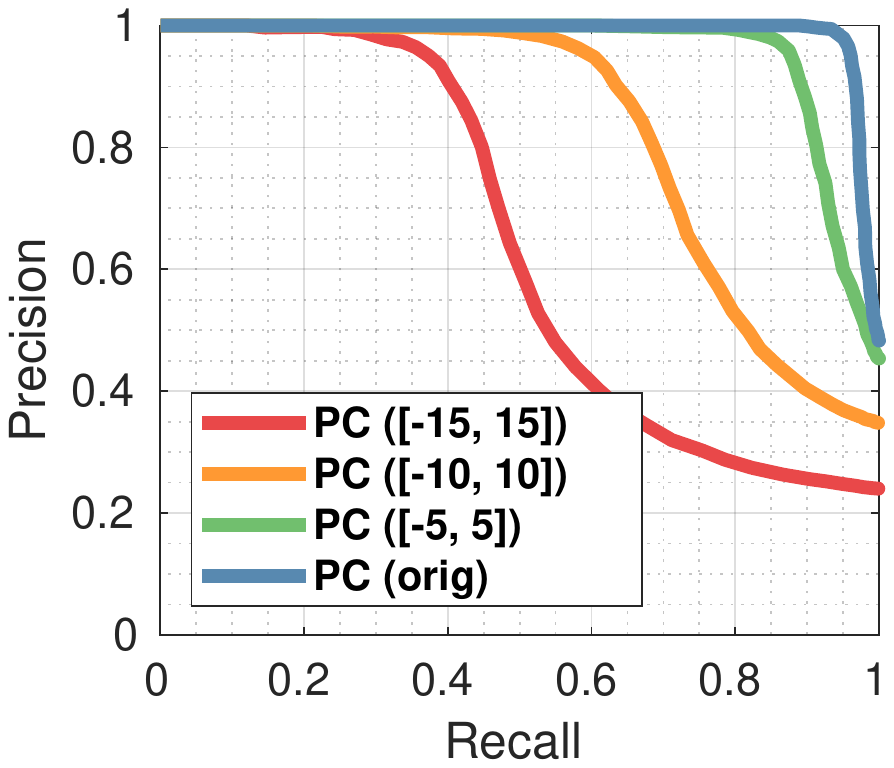}
    \label{fig:rollpitchsimul1}
  }%
  \subfigure[ CC - \texttt{KAIST 03} ]{%
    \centering
    \includegraphics[width=\mywidththird, trim = 160 280 190 295 , clip]{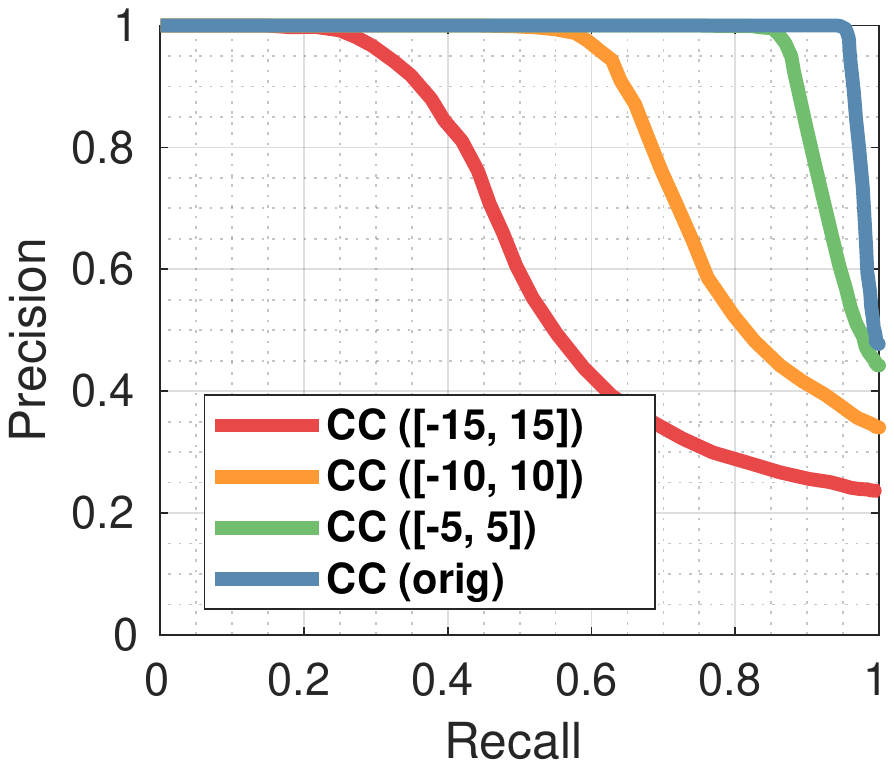}
    \label{fig:rollpitchsimul2}
  }%
  \subfigure[ M2DP - \texttt{KAIST 03} ]{%
    \centering
    \includegraphics[width=\mywidththird, trim = 160 280 190 295 , clip]{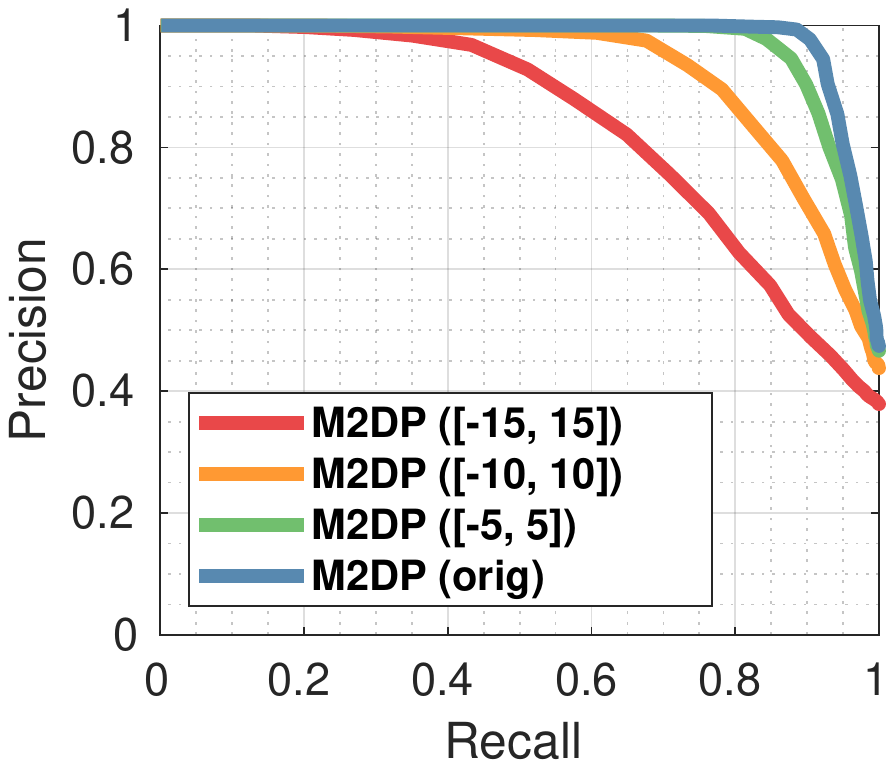}
    \label{fig:rollpitchsimul3}
  } 

  \subfigure[ PC - \texttt{Riv. 02} ]{%
  \centering
  \includegraphics[width=\mywidththird, trim = 160 280 190 295 , clip]{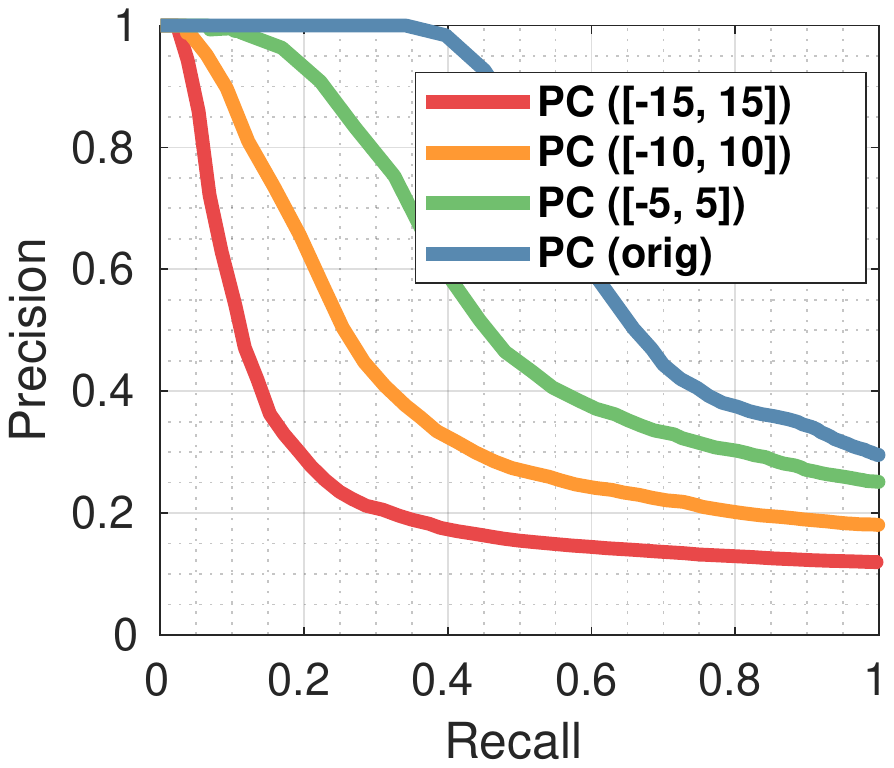}
  \label{fig:rollpitchsimul4}
}%
\subfigure[ CC - \texttt{Riv. 02} ]{%
  \centering
  \includegraphics[width=\mywidththird, trim = 160 280 190 295 , clip]{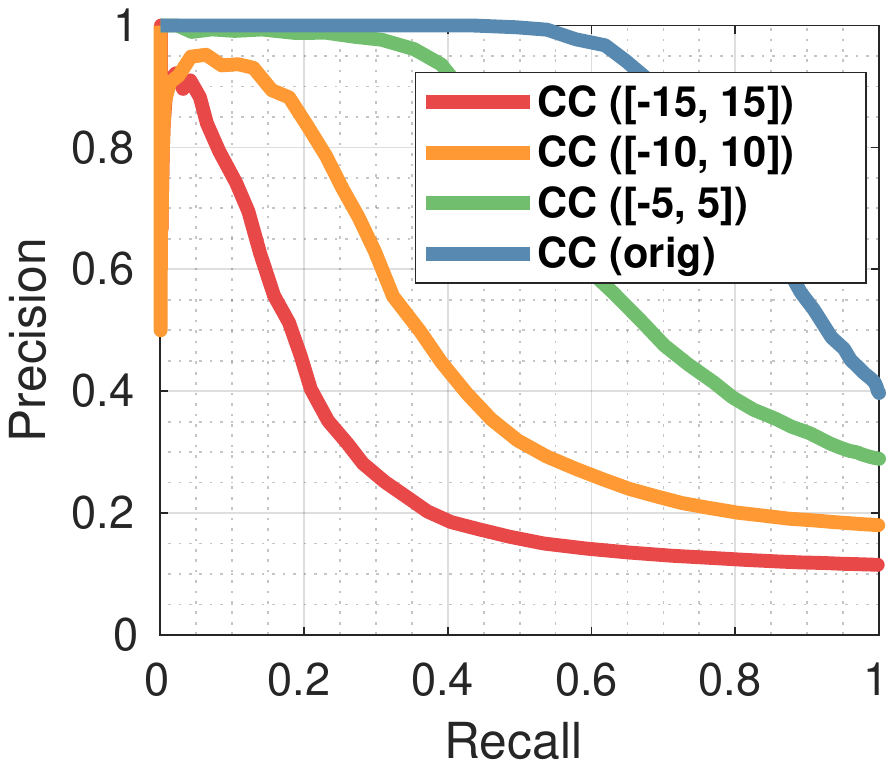}
  \label{fig:rollpitchsimul5}
}%
\subfigure[ M2DP - \texttt{Riv. 02} ]{%
  \centering
  \includegraphics[width=\mywidththird, trim = 160 280 190 295 , clip]{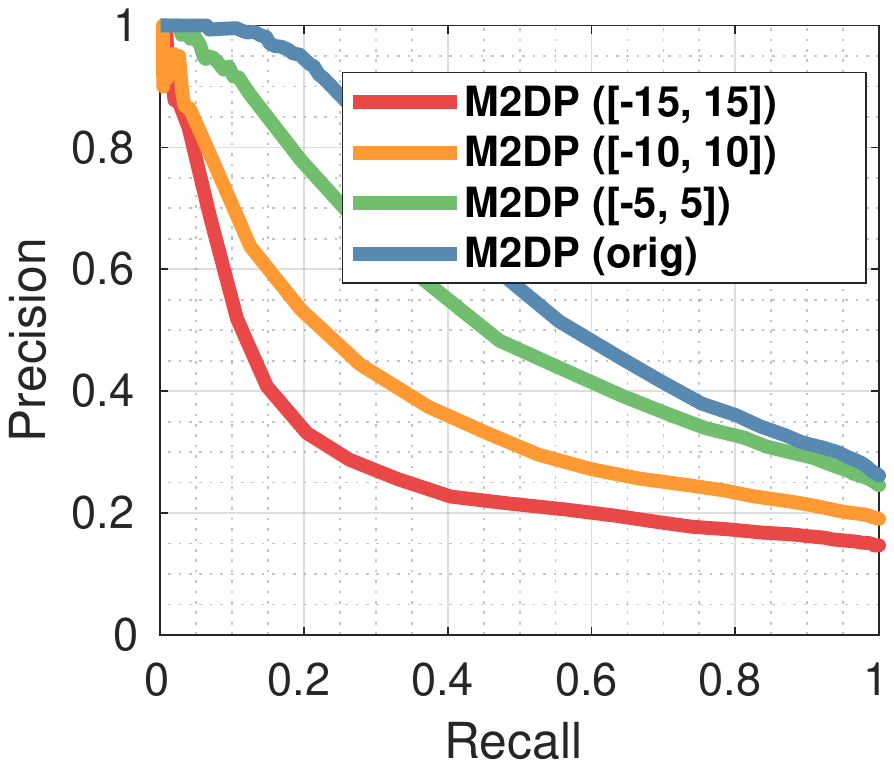}
  \label{fig:rollpitchsimul6}
} 

  \subfigure[ \texttt{KA Urban Campus 1} ]{%
    \centering
    \includegraphics[width=\mywidth, trim = 120 245 150 240, clip]{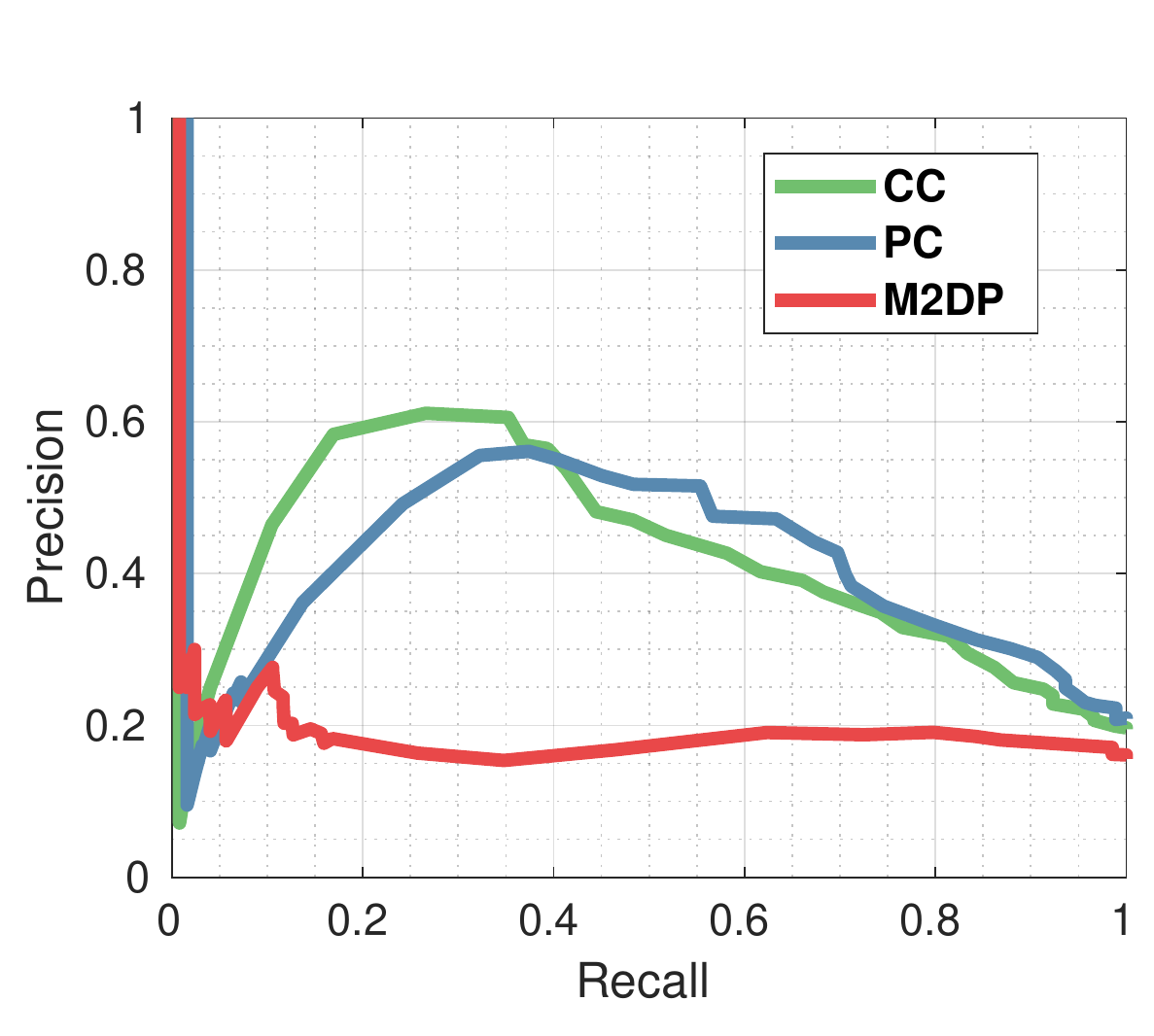}
    \label{fig:rollpitch3}
  }
  \subfigure[ Trajectory ]{%
    \centering
    \includegraphics[width=\mysmallwidth, trim = -40 -60 0 0, clip]{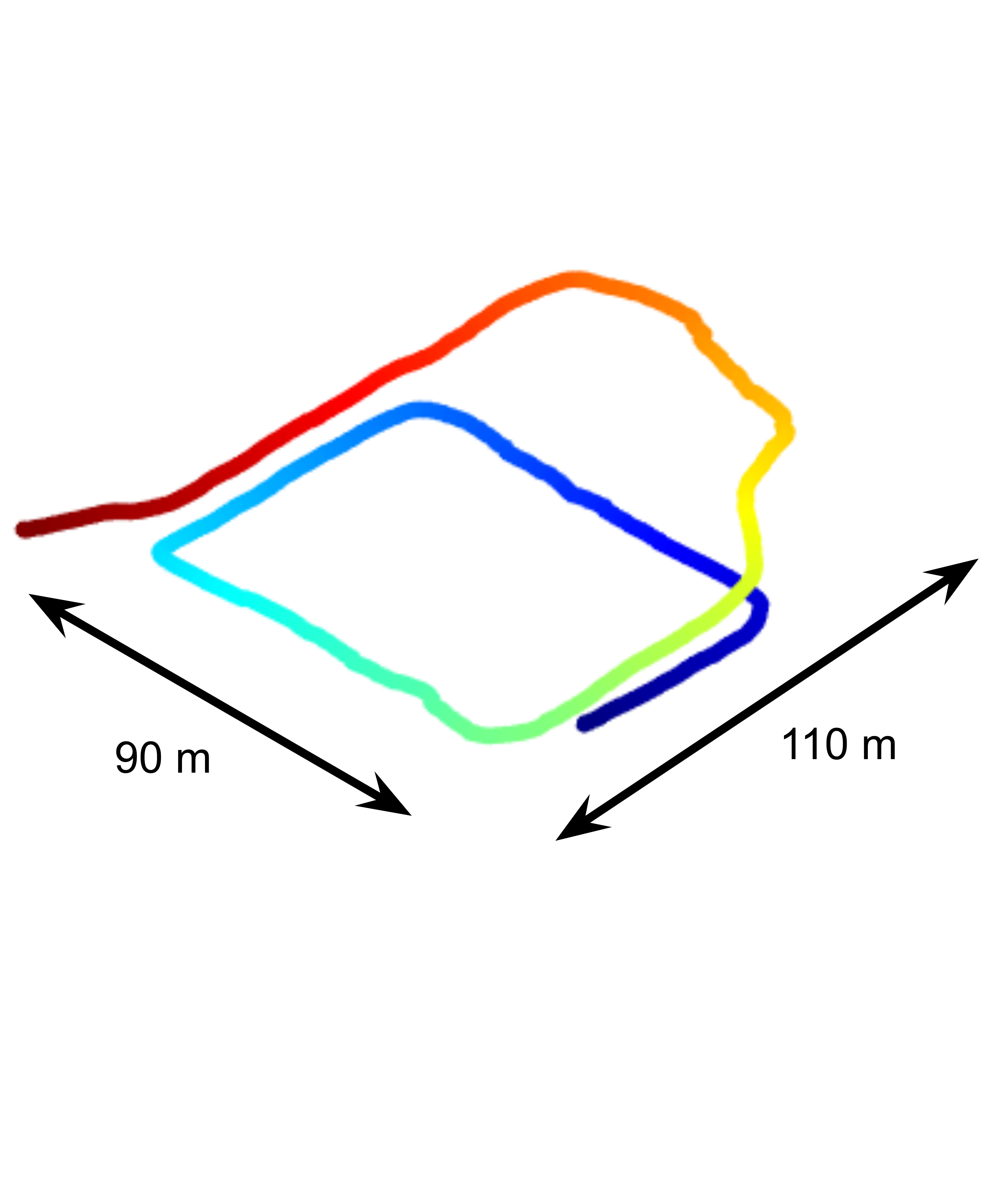}
    \label{fig:rollpitch4}
  }
  \subfigure[ Example scans]{%
    \centering
    \includegraphics[width=\mysmallwidth, trim = 0 0 0 0, clip]{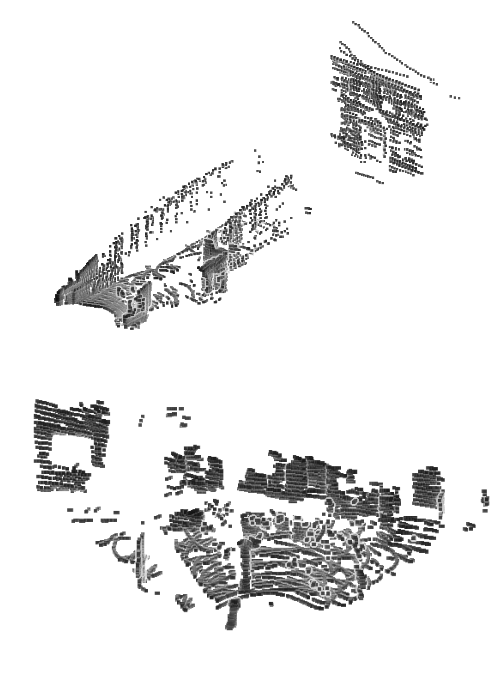}
    \label{fig:rollpitch5}
  }%

  \caption{ 
    \bl{
    \subref{fig:rollpitchsimul1}$-$\subref{fig:rollpitchsimul6} Perturbation simulation results. \subref{fig:rollpitch3}, \subref{fig:rollpitch4}, \subref{fig:rollpitch5} A real-world hand-held LiDAR dataset result, its time-elevated trajectory, and two example scans.
    } 
    }
  \label{fig:rollpitch}
  
\end{figure}

\bl{
  \textbf{Hand-held data.}
  Second, the result of real-world hand-held LiDAR data is given in \figref{fig:rollpitch3}. We used \texttt{KA Urban Campus 1} sequence provided in LiLi-OM \cite{li2021towards}, which was acquired from a slowly walking human navigator. It has the same direction revisits and narrow front horizontal FOVs ($\sim$\unit{70}{\degree}). In this real-world data, ours outperformed M2DP by a large margin and showed mild (e.g., human walking) roll-pitch and height perturbations are acceptable. Hence, the proposed method may not be restricted in wheeled platforms and work for a hand-held traverse under mild roll-pitch motions.
}

\subsection{Comparison to Deep learning-based methods}
\label{sec:dl}
\begin{figure}[!t]

  \centering
  \def\mywidth{0.32\columnwidth}%
  \subfigure[ \texttt{KAIST 03} ]{%
    \centering
    \includegraphics[width=\mywidth, trim = 190 285 190 285 , clip]{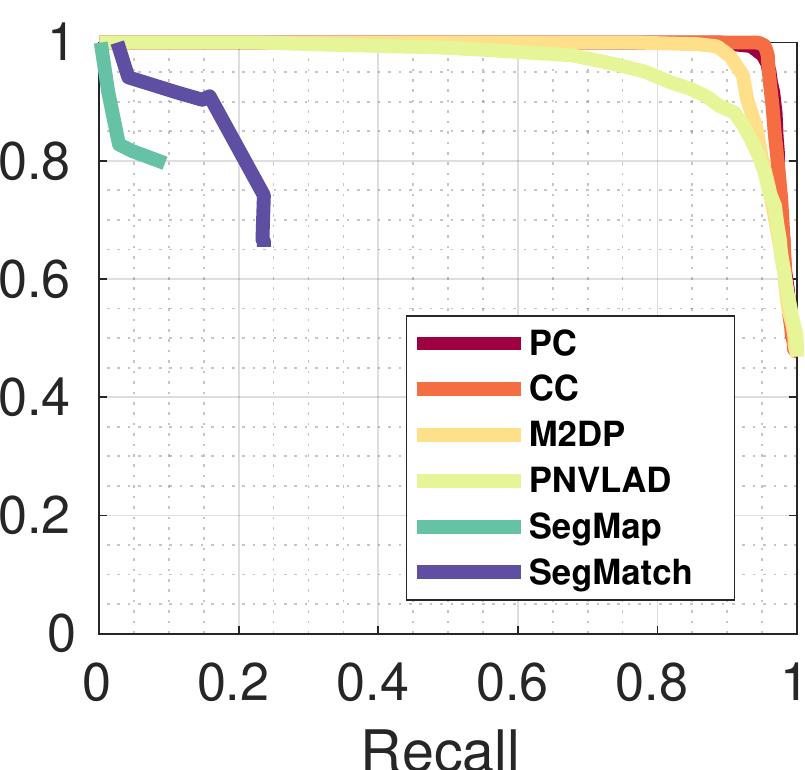}
    \label{fig:dl1}
  }%
  \subfigure[ \texttt{Riverside 02} ]{%
    \centering
    \includegraphics[width=\mywidth, trim = 190 285 190 285 , clip]{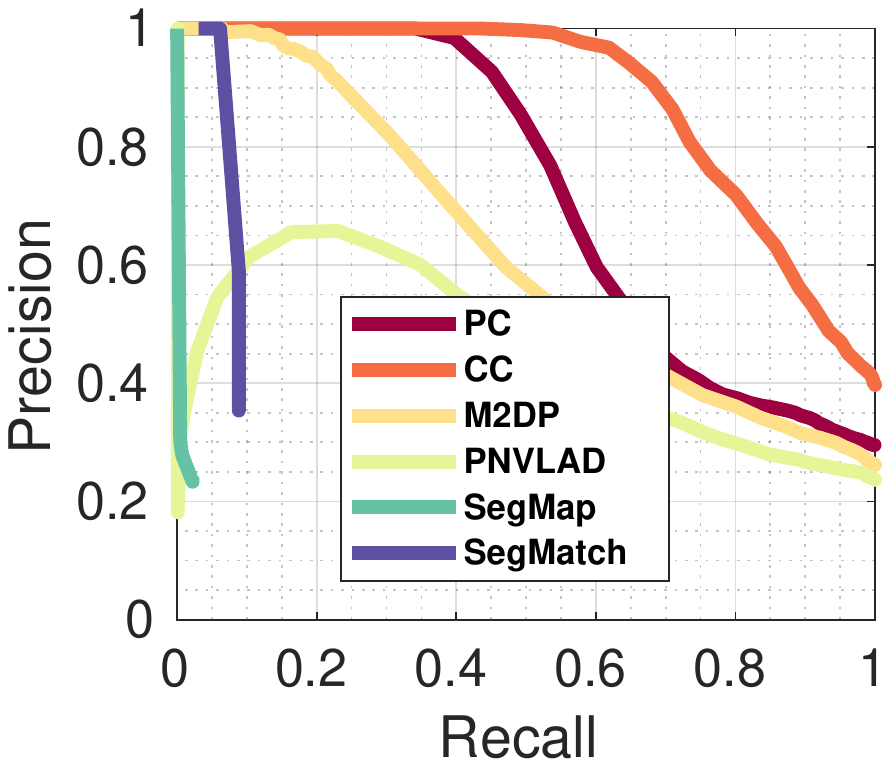}
    \label{fig:dl2}
  }%
  \subfigure[ \texttt{KITTI 08} ]{%
    \centering
    \includegraphics[width=\mywidth, trim = 190 285 190 285, clip]{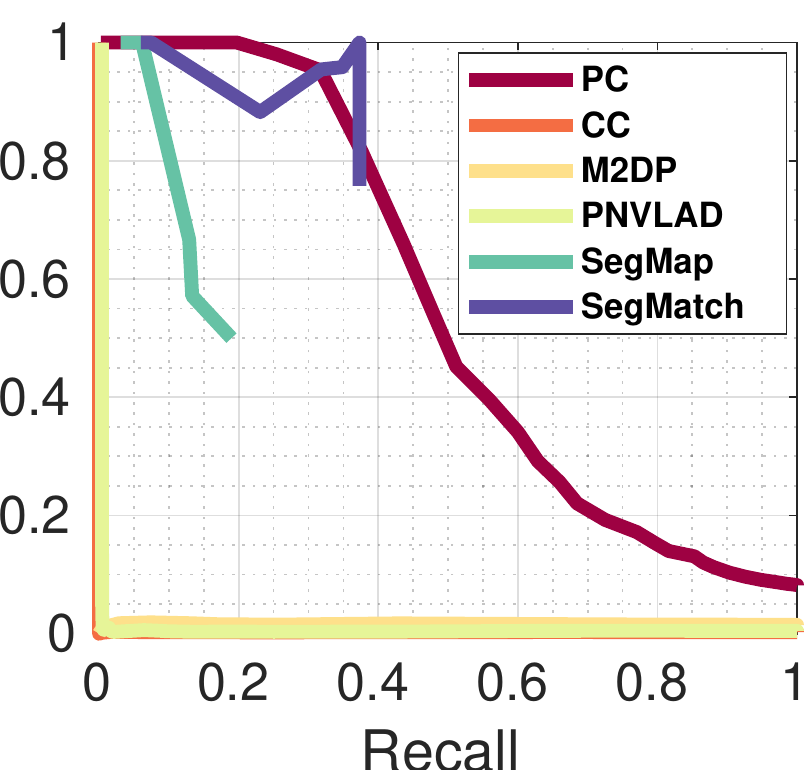}
    \label{fig:dl3}
  }%

  \caption{ 
    \bl{
      Comparisons with deep learning-based methods, SegMap and PointNetVLAD.
    }
  }
  \label{fig:dl}
  
\end{figure}

\bl{
We also provide comparisons to recent deep learning-based approaches, SegMap \cite{dube2020segmap} and PointNetVLAD \cite{angelina2018pointnetvlad}\footnote{\bl{For the input processing, we follow \cite{kim2019}. An input is a ground-removed, zero-centered 4096 points within a [\unit{-25}{m}, \unit{25}{m}] cubic region. }}. For both methods, we used pre-trained weights the authors released. SegMap revealed hampered performance compared to than SegMatch. This could be due to the limitation in generalization capability over unseen environments. PointNetVLAD showed comparable performance in the environment under little rotational and translational variations (i.e., \figref{fig:dl1}) but failed when the variation increased (\figref{fig:dl2} and (\figref{fig:dl3}), respectively).}

\subsection{Failure Cases}
\label{sec:summary}

\begin{figure}[!h]
  \centering
  \begin{minipage}{0.9\columnwidth}
  \subfigure[Successful cases]{%
    \includegraphics[width=\textwidth]{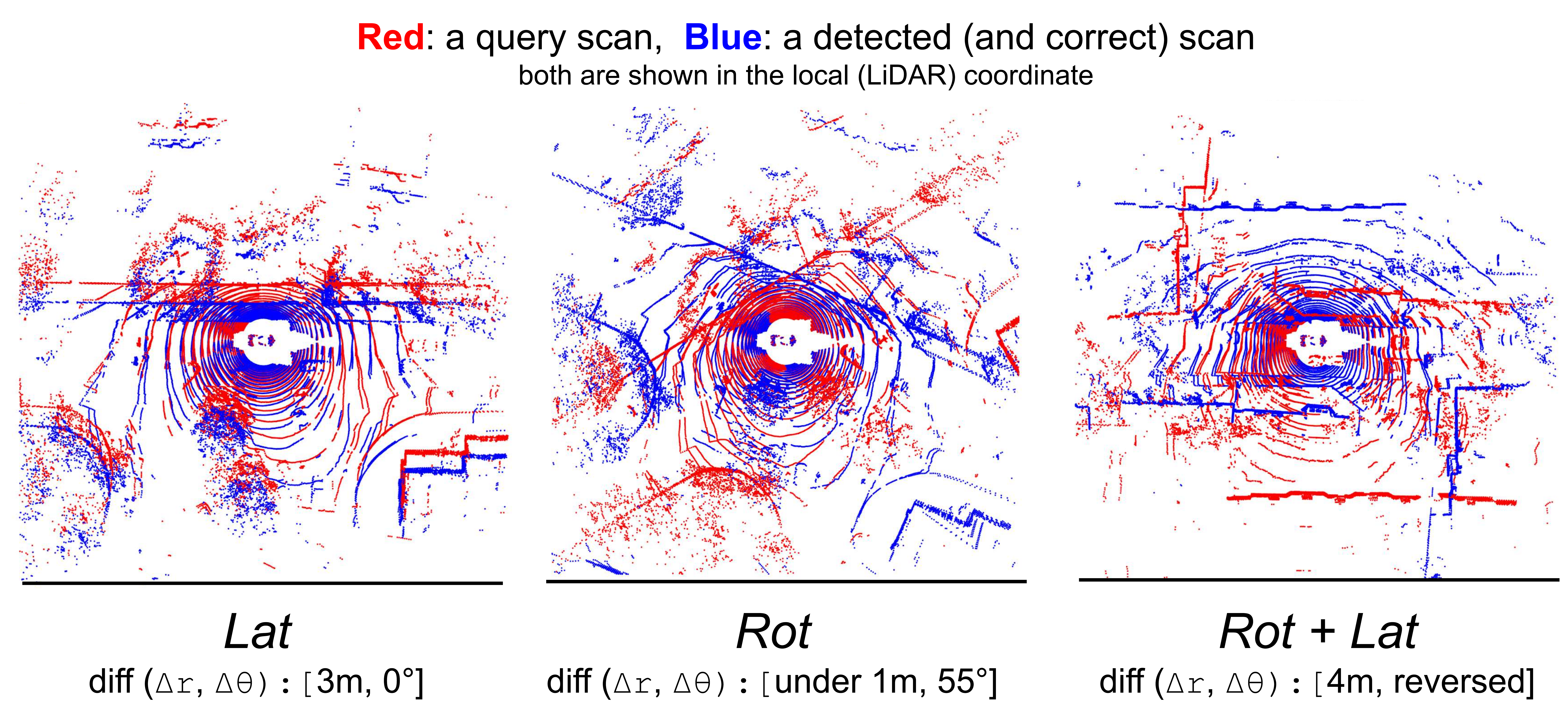}
    \label{fig:succ}
  }\\
  \centering
  \subfigure[Failure case: Corridor-like place]{%
    \includegraphics[width=0.8\textwidth, trim = 350 0 350 0, clip]{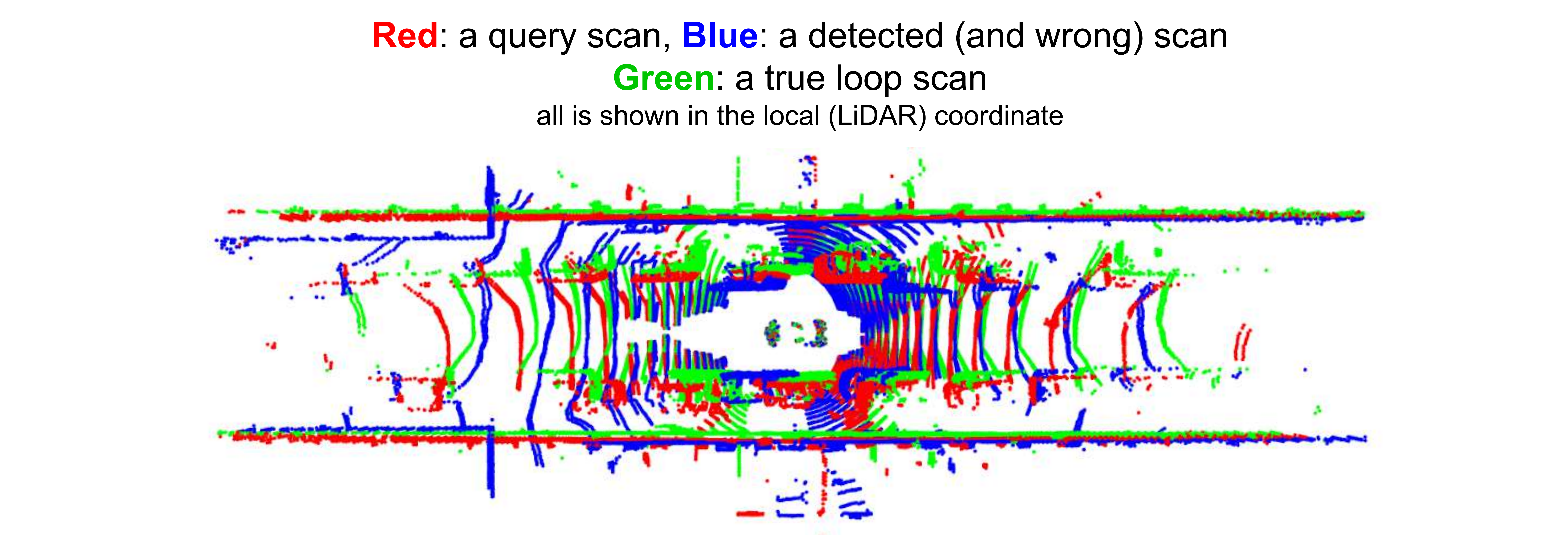}
    \label{fig:fail1}
  }\\
  \centering
  \subfigure[Failure case: Occlusion]{%
    \includegraphics[width=\textwidth]{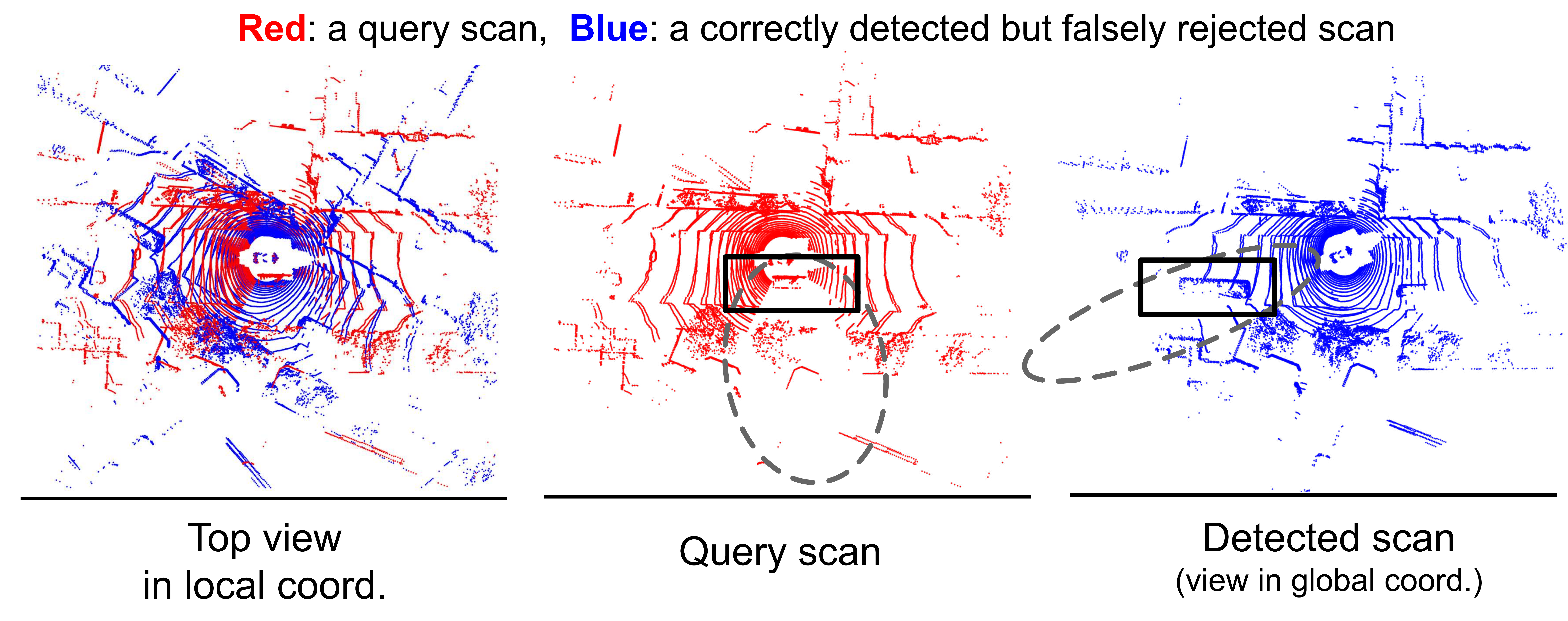}
    \label{fig:fail2}
  }%
  \end{minipage}
  \caption{\subref{fig:succ} Successful examples in \texttt{Oxford 2019-01-11-13-24-51} acquired by A-PC. \subref{fig:fail1} Perceptually aliased place from a corridor-like place. \subref{fig:fail2} When both query ($\sim$\unit{60}{\degree} loss) and map ($\sim$\unit{30}{\degree} loss) places undergo severe occlusions due to a tall and large object nearby (dotted ellipses), a quarter of the entire scan overlap is lost deteriorating localization capability.}
  \label{fig:succfail}

\end{figure}

We illustrate sample cases when the proposed method succeeded and failed to localize against the map. As shown in \figref{fig:succ}, the proposed method overcome lateral and/or rotational discrepancy between map and query scans. The \ac{SCD} is successfully localized to the map even with many dynamic objects (e.g., cars). However, when a tall and large object (e.g, bus) appears very proximal to the sensor on both query and map scans, the localization might fail as in \figref{fig:fail2}. The other failure case was found when the vehicle was moving along a corridor-like place (\figref{fig:fail1}).

\subsection{Which SCD to Use?}
\label{sec:whichscd}

The final question to answer is which SCD to use and in what case. Based on the evaluation, generally, A-CC yielded the best performance even under the composite variance (i.e., \textit{Rot + Lat}) as in the case of \texttt{Oxford} (\figref{fig:exp_oxford}) and \texttt{Pangyo} (\figref{fig:exp_pangyo}). \bl{Therefore, CC and A-CC are more preferred when the target environment is an urban road. We recommend using PC or A-PC for more general environments and when semi-metric localization capability is more critical. Because classic ICP is much more sensitive to the rotational component of the initialization, PC would be a better choice despite a little sacrifice in precisions from CC (but still comparable performance). For patrolling robots and shuttles that repeat the same route with minimum variance, PC would exhibit more meaningful performance as proved in multi-session scenarios (\figref{fig:exp_multisession}).}

\subsection{Limitations and Potential Extension}
\label{sec:limitation}

\subsubsection{Invariance in One Direction}
\label{sec:limit}

The proposed method is natively invariant in one direction and we chose rotation and lateral direction to be invariance axes. This limitation was overcome by a robust search scheme and augmentation.

\subsubsection{Leveraging Deep Learning for Scan Context Descriptor}
\label{sec:deep}

The proposed descriptor itself is in ordered 2D format, and inputting this into a deep network is very straightforward. As reported in \cite{kim2019, xu2021disco}, the descriptor is learnable and provides meaningful performance by only using a small network. This type of approach would particularly be beneficial when GPU is available and the localization is almost a memorization problem.

\subsubsection{Application to Non-urban Environment}
\label{sec:indoor}

The proposed method is most powerful in an urban environment where the descriptor can encode the nearby structural variance. The proposed descriptor is 1-channel with height value but easily expandable. For example, \cite{wang2020intensity} considered LiDAR intensity value as an additional channel to successfully operate in an indoor environment. Combining deep learning with indoor application yielded meaningful performance with dense pedestrian traffic \cite{spoxel2020}. We think incorporating point cloud distribution or semantic labels as additional channels would further enhance the scan context beyond the urban environment such as indoor and natural environments.

\subsubsection{Application to Other Range Sensors}
\label{sec:radarmisc}

The proposed descriptor is not limited to LiDAR sensors but also applicable to general range sensors including radars. As we reported a potential extension to radar sensors in \cite{kim2020mulran}, the descriptor can be applied to radars.

\subsubsection{\bl{Generalizability over measurement variation}}
\label{sec:radarmisc}

\bl{Future studies examining the sensor difference between the mapping and localization phase would also be meaningful. LiDAR measurement varies depending on the hardware choice and mounting configuration. Achieving generalizability over measurement variation would be needed.}

\section{Conclusion}
\label{sec:conclusion}

In this paper, we presented a global place recognition module combining topological and metric localization. As a global localizer, the proposed method can be a solution to a kidnapped robot problem serving as a place recognizer at a \textit{wake-up} phase. We also showed the invariance of \textit{Scan Context++} in both the rotational and lateral directions. Via the evaluation, we validated that the proposed localizer achieved discriminability and real-time performance without necessitating prior knowledge.


\ifCLASSOPTIONcaptionsoff
  \newpage
\fi

\renewcommand*{\bibfont}{\footnotesize}
\bibliographystyle{IEEEtranN} 
\bibliography{string-short,references}

%

\newcommand{\bioshot}[1]{\includegraphics[width=1in,height=1.25in,clip,keepaspectratio]{#1}}

\begin{IEEEbiography}[\bioshot{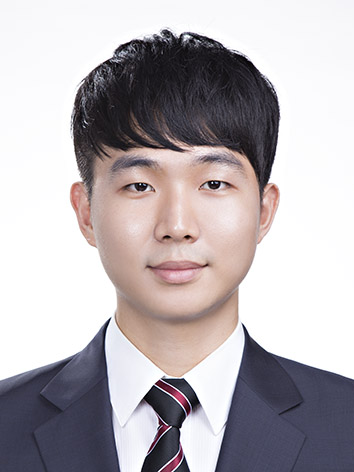}]{Giseop Kim}%
  (S'17) received the B.S. degrees in civil and environmental enginerring from KAIST, Daejeon, Korea, in 2017, and the M.S. degree in civil and environmental enginerring from KAIST, Daejeon, Korea, in 2019.
  Currently, he is the Ph.D student in the department of civil and environmental engineering, \ac{KAIST}. His research interests include LiDAR simultaneous localization and mapping and long-term map management.
\end{IEEEbiography}%
\begin{IEEEbiography}[\bioshot{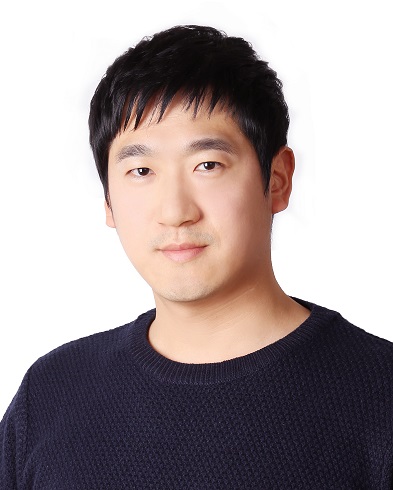}]{Sunwook Choi}%
   received the B.S. degree in electronic engineering from Inha University, Incheon, Korea, in 2007, and the M.S. degree in Information and Communication Engineering from Inha University, Incheon, Korea, in 2009. The Ph.D. degree in electronic engineering from INHA University, Incheon, Korea, in 2014. From March 2014 to January 2017, he was a research engineer at Cognitive Computing Group (Deep Learning Research Lab.), NAVER Corp., Korea. Since January 2017, he has been with the Autonomous Driving Group, NAVER LABS, Korea, where he is a senior research engineer. His research interests include simultaneous localization and mapping, perception for autonomous driving and deep learning for computer vision.
\end{IEEEbiography}%
%
%
\begin{IEEEbiography}[\bioshot{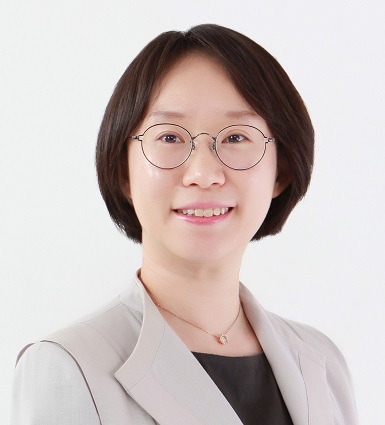}]{Ayoung Kim}%
  (S'08--M'13) received the B.S. and M.S. degrees in mechanical engineering from Seoul National University, Seoul, Korea, in 2005 and 2007, respectively, and the M.S. degree in electrical engineering and the Ph.D. degree in mechanical engineering from the University of Michigan (UM), Ann Arbor, in 2011 and 2012, respectively. She was an associate professor in the department of civil and environmental engineering with joint affiliation at KI robotics, Korea Advanced Institute of Science and Technology (KAIST) from 2014 to 2021. Currently, she is an associate professor in the department of mechanical engineering, Seoul National University (SNU). Her research interests include visual simultaneous localization and mapping, navigation.
\end{IEEEbiography}%



\vfill


\end{document}